\documentclass[]{worldbench}
\newcommand{\arxivpreamble}{}

\newif\ifarxivmode
\ifdefined\arxivpreamble \arxivmodetrue \else \arxivmodefalse \fi

\usepackage{amsmath,amssymb,amsfonts}
\usepackage{colortbl}
\usepackage{enumitem}
\usepackage{makecell}
\usepackage{mathtools}
\usepackage{multicol}
\usepackage{pifont}
\usepackage{array}
\usepackage{tabularx}
\usepackage{tikz}
\usetikzlibrary{trees,arrows.meta,positioning,shapes.geometric,calc,decorations.pathmorphing,decorations.pathreplacing,patterns,fit,backgrounds,shadows.blur}
\usepackage{wrapfig}
\usepackage{nicefrac}
\usepackage{xspace}

\usepackage{fontawesome}

\ifarxivmode
  \tcbuselibrary{breakable}
  \tcbuselibrary{skins}
  \usepackage{soul}
\else
  \usepackage{amsthm}
  \usepackage{algorithmic}
  \usepackage{booktabs}
  \usepackage{graphicx}
  \usepackage{microtype}
  \usepackage{multirow}
  \usepackage{subcaption}
  \usepackage[most]{tcolorbox}
  \tcbuselibrary{breakable}
  \tcbuselibrary{skins}
  \usepackage{balance}
\fi

\definecolor{S1color}{RGB}{58,120,48}     
\definecolor{S2color}{RGB}{42,155,132}    
\definecolor{S3color}{RGB}{119,171,86}    
\definecolor{S4color}{RGB}{138,164,38}    
\definecolor{S5color}{RGB}{79,149,195}    
\definecolor{S6color}{RGB}{152,112,182}   
\definecolor{S7color}{RGB}{82,56,112}     
\definecolor{S8color}{RGB}{232,179,65}    

\definecolor{P1color}{RGB}{119,171,86}    
\definecolor{P2color}{RGB}{79,149,195}    
\definecolor{P3color}{RGB}{114,84,141}    
\definecolor{P4color}{RGB}{232,179,65}    

\definecolor{refcolor}{RGB}{205,133,63}

\definecolor{tableheader}{HTML}{2C3E50}
\definecolor{tablelight}{HTML}{ECF0F1}
\definecolor{insightbg}{HTML}{F0F4FF}
\definecolor{linkcolor}{RGB}{52,152,219}

\ifarxivmode
  \definecolor{red}{rgb}{0.8,0,0}
  \definecolor{green}{RGB}{0, 133, 21}
  \definecolor{grey}{rgb}{0.5,0.5,0.5}
\fi

\ifarxivmode
  \hypersetup{citecolor=P2color,linkcolor=P2color}
  \RequirePackage{titletoc}
  \titlecontents{section}[0pt]
    {\addvspace{6pt}\sffamily\bfseries}
    {\contentslabel{1.5em}}{\hspace*{-1.5em}}
    {\titlerule*[0.5pc]{.}\contentspage}
  \titlecontents{subsection}[1.5em]
    {\hypersetup{linkcolor=P2color}}
    {\contentslabel{2.3em}}{\hspace*{-2.3em}}
    {\titlerule*[0.5pc]{.}\contentspage}
  \titlecontents{subsubsection}[3.8em]
    {\hypersetup{linkcolor=P2color}\small}
    {\contentslabel{3.2em}}{\hspace*{-3.2em}}
    {\titlerule*[0.5pc]{.}\contentspage}
\else
  \usepackage[pagebackref=true,colorlinks,linkcolor=linkcolor,citecolor=linkcolor,urlcolor=magenta]{hyperref}
  \usepackage{url}
  \usepackage[capitalize]{cleveref}

\fi

\ifarxivmode
  \makeatletter
  \DeclareRobustCommand\onedot{\futurelet\@let@token\@onedot}
  \def\@onedot{\ifx\@let@token.\else.\null\fi\xspace}
  \makeatother
  \newcommand*{\eg}{\textit{e.g}\onedot}

  \newcommand*{\etal}{\textit{et al}\onedot}

\else
  
  \newcommand*{\eg}{\emph{e.g.}\@\xspace}
  
  \newcommand*{\etal}{\emph{et al.}\@\xspace}
\fi

\newcommand{\cmark}{\ding{51}}
\newcommand{\pmark}{$\circ$}
\newcommand{\xmark}{---}
\newcommand{\cmarkc}{\textcolor{S3color}{\ding{51}}}
\newcommand{\pmarkc}{\textcolor{S4color}{$\circ$}}
\newcommand{\xmarkc}{\textcolor{gray!50}{---}}

\newcommand{\githubicon}[1]{\href{#1}{\raisebox{-0.1em}{\includegraphics[height=1em]{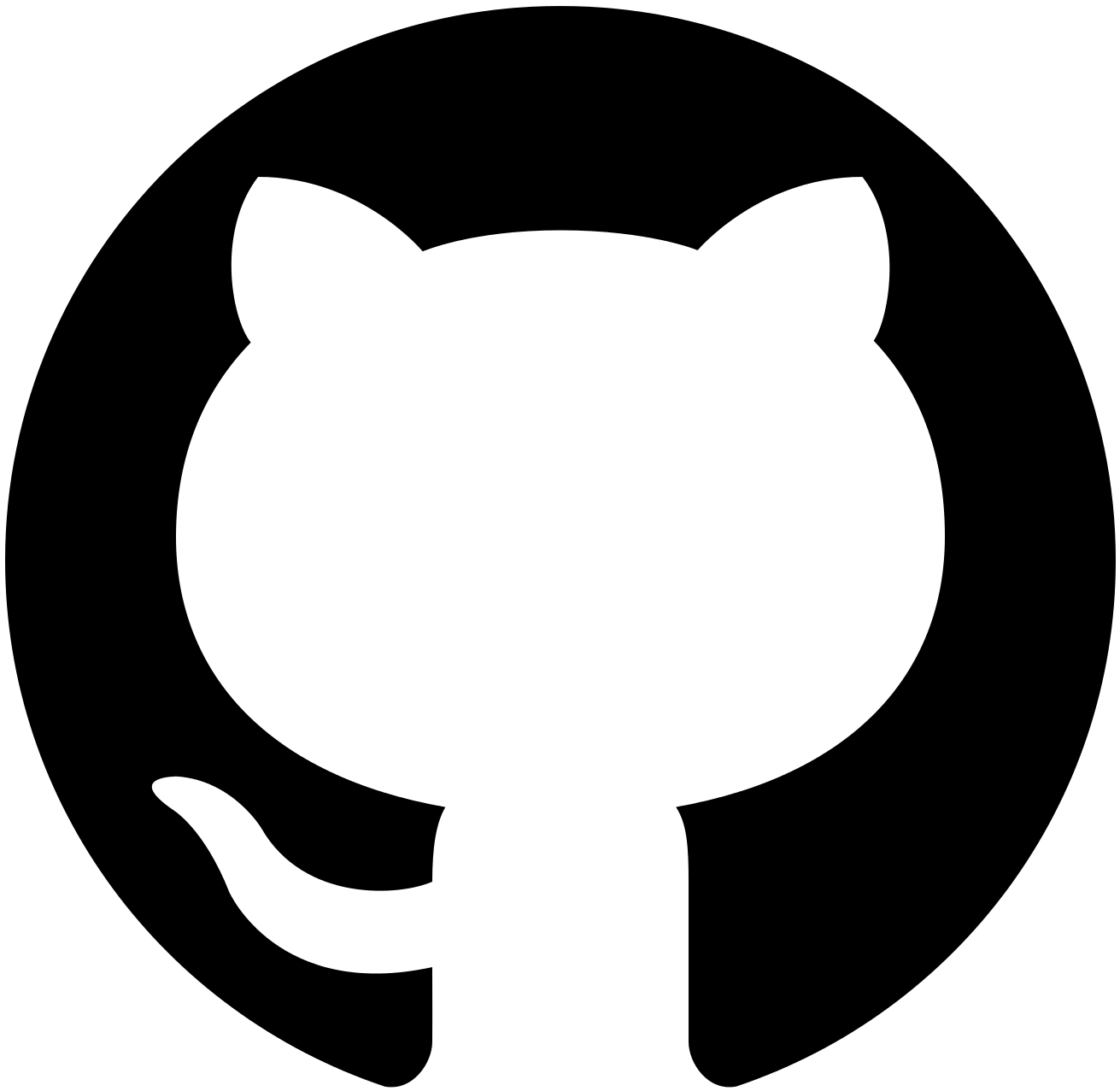}}}}
\newcommand{\hficon}[1]{\href{#1}{\raisebox{-0.1em}{\includegraphics[height=1em]{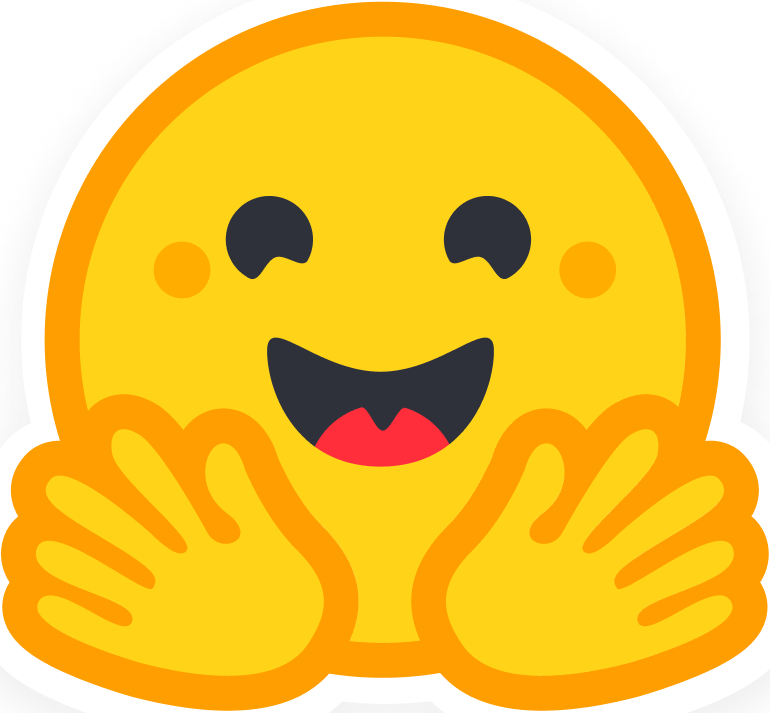}}}}

\newcommand{\pdficon}[1]{\href{#1}{\tcbox[on line, nobeforeafter, tcbox raise base, boxsep=0pt, left=1.5pt, right=1.5pt, top=0.5pt, bottom=0.5pt, arc=1pt, boxrule=0.2pt, colback=linkcolor!15, colframe=linkcolor!50]{\tiny\bfseries\sffamily\textcolor{linkcolor}{PDF}}}}

\newcommand{\levelbadge}[2]{%
  \tcbox[on line, nobeforeafter, tcbox raise base,
         boxsep=0pt, left=2pt, right=2pt, top=0.5pt, bottom=0.5pt,
         arc=1.5pt, boxrule=0.3pt,
         colback=#1!15, colframe=#1!70]%
    {\scriptsize\bfseries\sffamily\textcolor{#1!55!black}{#2}}%
}
\newcommand{\Sone}{\texorpdfstring{\levelbadge{S1color}{S1}}{S1}\xspace}
\newcommand{\Stwo}{\texorpdfstring{\levelbadge{S2color}{S2}}{S2}\xspace}
\newcommand{\Sthree}{\texorpdfstring{\levelbadge{S3color}{S3}}{S3}\xspace}
\newcommand{\Sfour}{\texorpdfstring{\levelbadge{S4color}{S4}}{S4}\xspace}
\newcommand{\Sfive}{\texorpdfstring{\levelbadge{S5color}{S5}}{S5}\xspace}
\newcommand{\Ssix}{\texorpdfstring{\levelbadge{S6color}{S6}}{S6}\xspace}
\newcommand{\Sseven}{\texorpdfstring{\levelbadge{S7color}{S7}}{S7}\xspace}
\newcommand{\Seight}{\texorpdfstring{\levelbadge{S8color}{S8}}{S8}\xspace}

\crefformat{section}{\textcolor{black}{Section~}#2#1#3}
\crefformat{subsection}{\textcolor{black}{Section~}#2#1#3}
\crefformat{subsubsection}{\textcolor{black}{Section~}#2#1#3}
\crefformat{table}{\textcolor{black}{Table~}#2#1#3}
\crefformat{figure}{\textcolor{black}{Figure~}#2#1#3}
\crefrangeformat{section}{\textcolor{black}{Sections~}#3#1#4\textcolor{black}{ to~}#5#2#6}
\Crefformat{section}{\textcolor{black}{Section~}#2#1#3}
\Crefformat{table}{\textcolor{black}{Table~}#2#1#3}
\Crefformat{figure}{\textcolor{black}{Figure~}#2#1#3}

\newtcolorbox{insightbox}[1][]{%
  enhanced, breakable,
  colback=insightbg, colframe=linkcolor!60, boxrule=0.6pt,
  left=6pt, right=6pt, top=5pt, bottom=5pt,
  fonttitle=\small\bfseries, title={#1},
  before upper={\small}, before skip=6pt, after skip=6pt,
}

\newtcolorbox{stageinsightbox}[2][]{%
  enhanced, breakable,
  colback=#2!5, colframe=#2!60, boxrule=0.6pt,
  left=6pt, right=6pt, top=5pt, bottom=5pt,
  fonttitle=\small\bfseries, coltitle=black, title={#1},
  before upper={\small}, before skip=6pt, after skip=6pt,
}

\newtcolorbox{stagebox}[2][]{%
  enhanced,
  colback=#2!5, colframe=#2!70, boxrule=0.5pt,
  left=4pt, right=4pt, top=3pt, bottom=3pt,
  fonttitle=\small\bfseries\sffamily, coltitle=white, colbacktitle=#2!85,
  title={#1}, before upper={\small}, before skip=3pt, after skip=3pt,
}

\newcommand{\anadotrule}{\par\vspace{2pt}\noindent\textcolor{black!20}{\dotfill}\par\vspace{2pt}}

\newcommand{\posbadge}[2][S1color]{\colorbox{#1!15}{\raisebox{0pt}[\height][1pt]{\scriptsize\strut\textcolor{#1!90}{\textsf{\textbf{#2}}}}}}
\newcommand{\negbadge}[1]{\colorbox{red!8}{\raisebox{0pt}[\height][1pt]{\scriptsize\strut\textcolor{red!70}{\textsf{\textbf{#1}}}}}}
\newcommand{\stageanalysis}[8]{%
\begin{tcolorbox}[
  enhanced,
  colback=white, colframe=#2!60, boxrule=0.8pt,
  left=6pt, right=6pt, top=0pt, bottom=4pt,
  fonttitle=\small\bfseries\sffamily, coltitle=white, colbacktitle=#2!80,
  title={\raisebox{-0.1em}{\faLightbulbO}~~#1},
  before skip=8pt, after skip=8pt,
  overlay={\draw[#2!80, line width=0.6pt]
    ([yshift=-2pt]frame.north) -- ([yshift=2pt]frame.south);},
]
\small
\parbox[t]{0.47\linewidth}{%
\vspace{4pt}
\noindent
\begin{minipage}[c]{0.25\linewidth}
\centering
\includegraphics[width=\linewidth]{#3}
\end{minipage}%
\hfill
\begin{minipage}[c]{0.70\linewidth}
\textbf{\textcolor{#2!80}{\faCheckCircle~~State \& Progress}}\par
\vspace{3pt}
#5
\end{minipage}
\vspace{4pt}
\anadotrule
#6
\vspace{2pt}
}%
\hfill
\parbox[t]{0.47\linewidth}{%
\vspace{4pt}
\noindent
\begin{minipage}[c]{0.25\linewidth}
\centering
\includegraphics[width=\linewidth]{#4}
\end{minipage}%
\hfill
\begin{minipage}[c]{0.70\linewidth}
\textbf{\textcolor{red!70}{\faExclamationTriangle~~Gaps \& Limitations}}\par
\vspace{3pt}
#7
\end{minipage}
\vspace{4pt}
\anadotrule
#8
\vspace{2pt}
}%
\end{tcolorbox}
}

\newcommand{\stagecard}[5]{%
  \begin{tcolorbox}[
    enhanced, boxrule=0.8pt,
    colframe=#3!70, colback=#3!4,
    left=26pt, right=14pt, top=2pt, bottom=2pt,
    before skip=5pt, after skip=5pt,
  ]
  \begin{minipage}[c]{0.12\columnwidth}
    \centering
    \includegraphics[width=\linewidth]{#4}
  \end{minipage}\hspace{12.5pt}%
  \begin{minipage}[c]{0.82\columnwidth}
    {\sffamily\bfseries #1~#2}\par\smallskip
    {\small #5}
  \end{minipage}
  \end{tcolorbox}
}

\title{AI for Auto-Research: Roadmap \& User Guide}

\author[]{Lingdong~Kong$^{*}$}
\author[]{Xian~Sun$^{*}$}
\author[]{Wei~Chow$^{*}$}
\author[]{Linfeng~Li}
\author[]{Kevin~Qinghong~Lin}
\author[]{Xuan~Billy~Zhang}
\author[]{Song~Wang}
\author[]{Rong~Li}
\author[]{Qing~Wu}
\author[]{Wei~Gao}
\author[]{Yingshuo~Wang}
\author[]{Shaoyuan~Xie}
\author[]{Jiachen~Liu}
\author[]{Leigang~Qu}
\author[]{Shijie~Li}
\author[]{Lai~Xing~Ng}
\author[]{Benoit~R.~Cottereau}
\author[]{Ziwei~Liu}
\author[]{Tat-Seng~Chua}
\author[]{Wei~Tsang~Ooi}

\affiliation[]{Awesome AI Auto-Research Team}

\abstract{AI-assisted research is crossing a threshold: fully automated systems can now generate research papers for as little as \$15, while long-horizon agents can execute experiments, draft manuscripts, and simulate critique with minimal human input. Yet this productivity frontier exposes a deeper integrity problem: under scientific pressure, even frontier LLMs still fabricate results, miss hidden errors, and fail to judge novelty reliably. Studying developments through June 2026, we present an end-to-end analysis of AI across the \emph{complete} research lifecycle, organized into four epistemological phases: $^1$\textbf{Creation} (idea generation, literature review, coding \& experiments, tables \& figures), $^2$\textbf{Writing} (paper writing), $^3$\textbf{Validation} (peer review, rebuttal \& revision), and $^4$\textbf{Dissemination} (posters, slides, videos, social media, project pages, and interactive agents). We identify a sharp, stage-dependent boundary between reliable assistance and unreliable autonomy: AI excels at structured, retrieval-grounded, and tool-mediated tasks, but remains fragile for genuinely novel ideas, research-level experiments, and scientific judgment. Generated ideas often degrade after implementation, research code lags far behind pattern-matching benchmarks, and end-to-end autonomous systems have not yet consistently reached major-venue acceptance standards. We further show that greater automation can obscure rather than eliminate failure modes, making human-governed collaboration the most credible deployment paradigm. Finally, we provide a structured taxonomy, benchmark suite, and tool inventory, cross-stage design principles, and a practitioner-oriented playbook, with resources maintained at our project page.}

\metadata[
\raisebox{-0.2em}{\includegraphics[width=0.025\linewidth]{figures/icons/hf.png}}~~Project Page]{\href{https://worldbench.github.io/awesome-ai-auto-research}{\texttt{https://worldbench.github.io/awesome-ai-auto-research}}
\\[-1.5ex]}

\metadata[
\raisebox{-0.2em}{\includegraphics[width=0.025\linewidth]{figures/icons/github.png}}~~GitHub Repo]{\href{https://github.com/worldbench/awesome-ai-auto-research}{\texttt{https://github.com/worldbench/awesome-ai-auto-research}}
\\[-1.5ex]}

\metadata[Latest Version]{July 20, 2026}

\begin{document}

\maketitle

\begin{figure}[!h]
    \centering
    \vspace{0.2cm}
    \includegraphics[width=\linewidth]{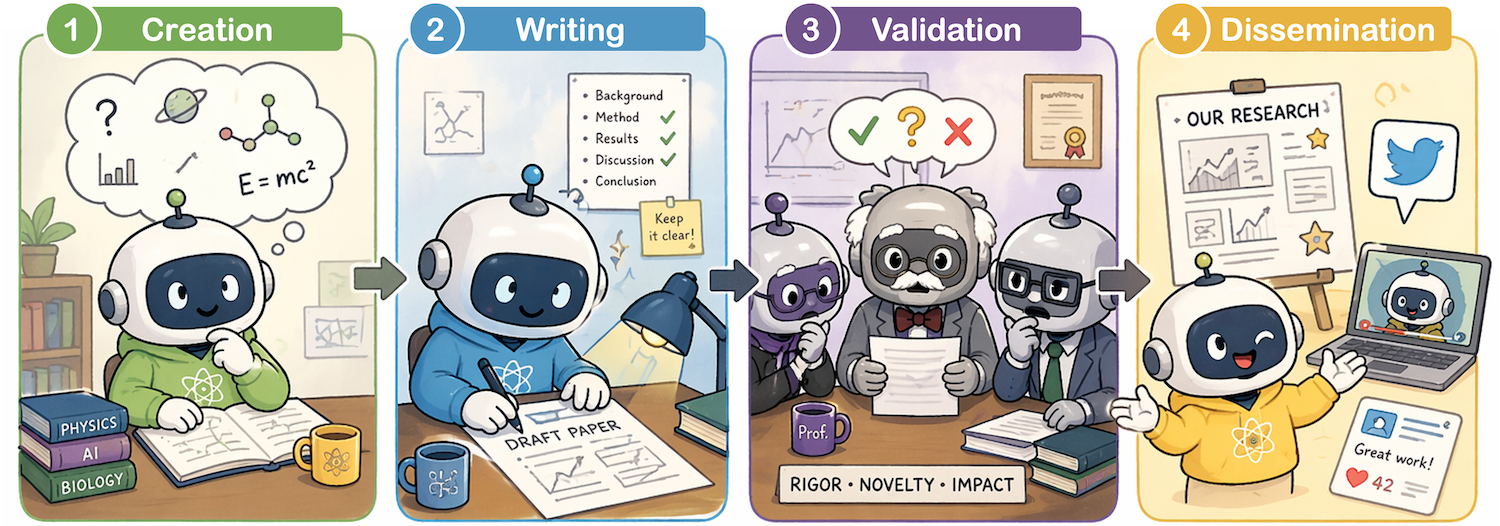}
    \vspace{-0.5cm}
    \caption{\textbf{AI auto-research across the complete lifecycle.} We organize AI assistance into four phases and eight stages: $^1$\textbf{Creation} spans idea generation, literature review, coding \& experiments, and tables \& figures; $^2$\textbf{Writing} centers on paper writing; $^3$\textbf{Validation} includes peer review and rebuttal \& revision; and $^4$\textbf{Dissemination} transforms papers into posters, slides, videos, social media, project pages, and interactive paper agents.}
    \label{fig:teaser}
    \vspace{-1cm}
\end{figure}

\clearpage
\setcounter{tocdepth}{3}
\tableofcontents
\clearpage

\section{Introduction}
\label{sec:introduction}

AI-assisted research is crossing a threshold. Large language models (LLMs) and their agentic extensions are no longer limited to local writing or coding support; they are beginning to operate across the research lifecycle itself. Recent systems illustrate the scale of this shift: The AI Scientist generated complete research papers at roughly \$15 per paper~\cite{lu2024aiscientist}; FARS ran continuously for $228$ hours, consumed $11.4$ billion tokens, and produced $100$ papers, averaging one every $2.3$ hours~\cite{fars2026_report}; and ARIS reports an overnight workflow that ran $20+$ GPU experiments, pruned unsupported claims, and improved a draft score from $5.0$ to $7.5$ through iterative review and revision~\cite{aris2025}. These systems suggest a new paradigm: AI is moving from assisting individual research tasks to orchestrating multi-stage workflows that generate ideas, search literature, execute experiments, draft manuscripts, simulate critique, and prepare dissemination materials.

This rapid progress also exposes the defining tension of the field. AI systems are increasingly capable of producing research-like artifacts, yet remain far less reliable at verifying whether those artifacts are novel, faithful, executable, and scientifically meaningful. Generated ideas can appear promising but weaken after implementation~\cite{si2025gap}; generated code can run while implementing the wrong algorithm~\cite{researchcodebench2025}; fluent manuscripts can conceal unsupported claims; automated reviews can be coherent yet lenient or vulnerable to manipulation~\cite{llmreviewer2025}; rebuttals can promise revisions that are not later fulfilled~\cite{rebuttalcommitment2026}; and dissemination materials can simplify results beyond the evidence. The core challenge is therefore no longer whether AI can produce the \emph{forms} of research, but whether it can preserve the \emph{substance} of research: evidence, judgment, provenance, and accountability.

A lifecycle view is essential for understanding this challenge. Research is not a collection of independent tasks: ideas become experiments, experiments become claims, claims become manuscripts, reviews become revisions, and papers become public-facing summaries. Errors introduced early can be amplified downstream, especially when AI systems generate plausible outputs without preserving evidence or provenance. Despite the rapid emergence of research agents, writing assistants, scientific coding tools, automated reviewers, rebuttal systems, and Paper2X applications, the field still lacks a unified analysis of AI auto-research across the complete academic lifecycle. Without such a view, it is difficult to determine where AI reliably helps, where it fails systematically, and which deployment modes are scientifically credible.

Surveying developments through June 2026, we present the first end-to-end analysis of AI auto-research across the complete academic research lifecycle. We organize the field into four epistemological phases and eight stages: $^1$\textbf{Creation}, covering idea generation, literature review, coding \& experiments, and tables \& figures; $^2$\textbf{Writing}, covering paper writing; $^3$\textbf{Validation}, covering peer review and rebuttal \& revision; and $^4$\textbf{Dissemination}, covering posters, slides, videos, social media, project pages, and interactive paper agents. This structure follows the temporal sequence of research while making explicit the distinct AI capabilities, risks, and verification requirements introduced by each phase.

Our analysis yields five central findings. First, AI capability is strongest when tasks are structured, grounded, and externally checkable, but drops sharply for open-ended research tasks requiring novelty, implicit domain knowledge, long-horizon reasoning, or scientific judgment. Second, artifact generation consistently outpaces verification: across stages, AI can often produce plausible outputs faster than it can prove that they are correct, faithful, or meaningful. Third, the most reliable deployment mode is human-governed collaboration rather than full autonomy: AI can reduce mechanical friction in retrieval, drafting, coding, visualization, review support, and dissemination, but researchers must retain responsibility for judgment, interpretation, experimental design, argumentation, and accountability. Fourth, effective systems increasingly rely on layered architectures that combine exploration, tool-based execution, and verification, suggesting that orchestration, provenance, and feedback design are as important as model scale. Fifth, AI use in research is becoming a governance problem rather than a detection problem: as AI assistance becomes routine, the key questions are disclosure, attribution, responsibility, and whether scientific integrity is preserved.

This work makes three contributions to the emerging field of AI auto-research:

\begin{itemize}

    \item We provide a unified taxonomy of AI auto-research across four phases and eight stages, covering both mature areas such as writing and coding, and underexplored areas such as rebuttal, scientific visualization, and research dissemination.

    \item We synthesize tools, benchmarks, and methodological families across the lifecycle, showing how systems have evolved from prompt-based assistance to retrieval-augmented, agentic, fine-tuned, and hybrid workflows.

    \item We identify cross-cutting capability boundaries and open challenges, including phase-boundary faithfulness, scientific judgment, reproducibility, citation provenance, governance, cross-domain generalization, and cognitive ownership.
\end{itemize}

The remainder of this paper is organized as follows. \cref{sec:preliminaries} introduces the lifecycle framework, methodological families, literature-collection scope, and development timeline. \cref{sec:creation} to \cref{sec:dissemination} build the roadmap of the four phases for AI-assisted research in temporal order. \cref{sec:cross_cutting} synthesizes end-to-end systems, evaluation paradigms, cross-cutting insights, and open challenges. \cref{sec:conclusion} concludes the paper.

\section{Preliminaries}
\label{sec:preliminaries}

As AI-assisted research tools expand from isolated single stages (such as writing or coding aids) into multi-stage assistants, the field has become increasingly difficult to compare using a single vocabulary. Existing systems differ not only in their technical designs, but also in the research stages they target, the degree of autonomy they assume, and the forms of scientific risk they introduce. 

To support a unified analysis, we first establish \textbf{four foundational elements}: (i) the high-level academic research lifecycle framework that organizes this survey (\cref{sec:lifecycle}), (ii) the methodological families that recur across each stage (\cref{sec:methods_overview}), (iii) the scope and methodology of our literature collection (\cref{sec:scope}), and (iv) a brief timeline of key developments (\cref{sec:timeline}).

\subsection{Research Lifecycle}
\label{sec:lifecycle}

We define the research lifecycle as \textbf{eight interconnected stages}, organized into \textbf{four phases}. Each phase groups stages that serve a shared function in the production, validation, and communication of scientific knowledge.

\vspace{6pt}
\noindent\underline{\textbf{Phase 1: Creation.}} This phase covers the stages through which a research contribution is materially produced, including hypothesis formation, evidence gathering, experimentation, and scientific visualization.

\stagecard{\Sone}{Idea Generation}{P1color}{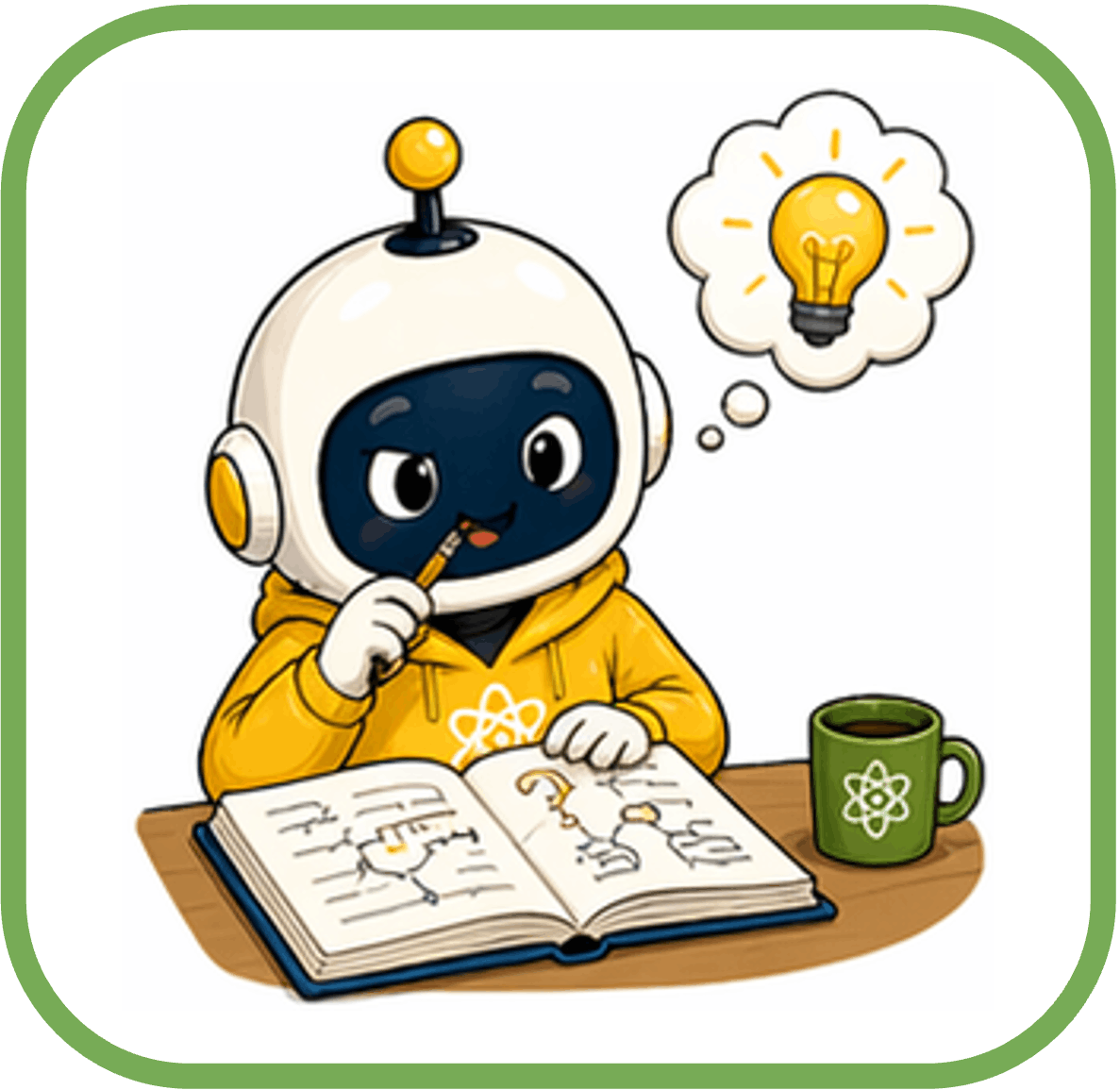}{%
Generating, refining, and evaluating research hypotheses. Techniques include direct LLM prompting, retrieval-augmented generation, knowledge-graph reasoning, and multi-agent collaboration for structured hypothesis formation.}

\stagecard{\Stwo}{Literature Review}{P1color}{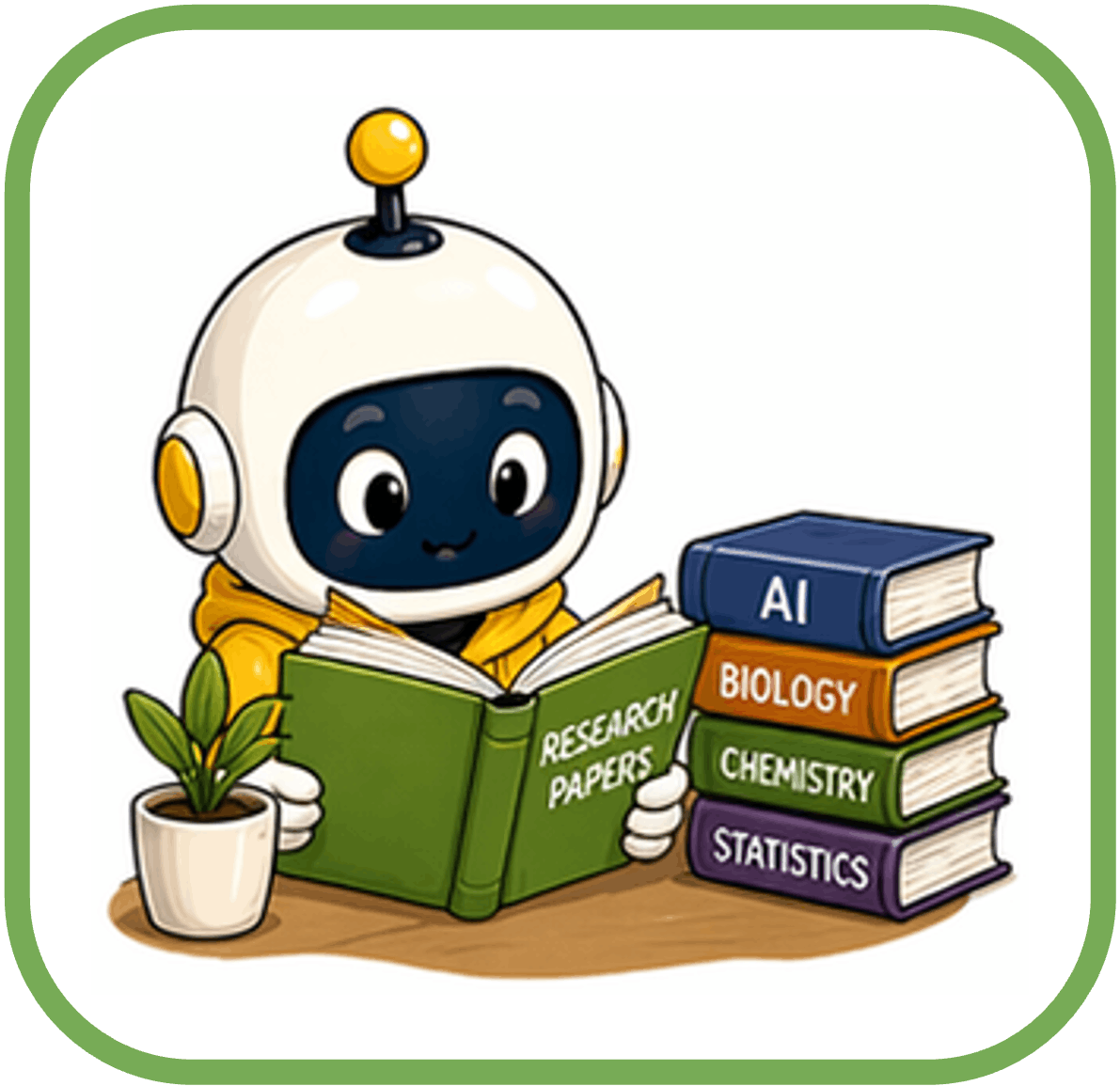}{%
Retrieving, synthesizing, and organizing prior work into coherent research contexts. Modern systems span semantic retrieval, citation-graph traversal, survey generation, and deep research agents that iteratively explore the literature.}

\stagecard{\Sthree}{Coding \& Experiments}{P1color}{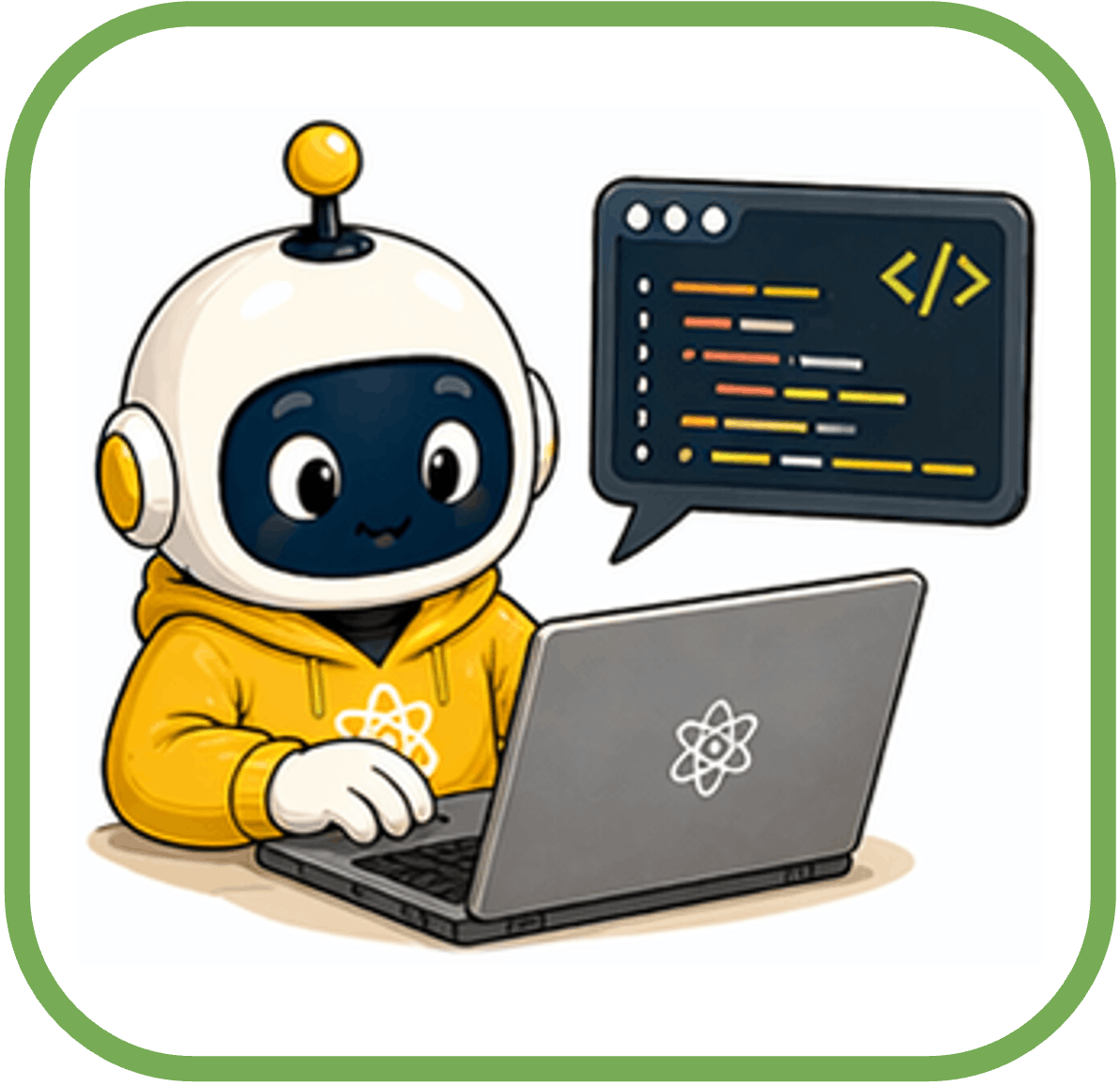}{%
Translating ideas into executable code, running experiments, and analyzing empirical results. This stage includes code generation, paper-to-code translation, autonomous experiment orchestration, and result interpretation.}

\stagecard{\Sfour}{Tables \& Figures}{P1color}{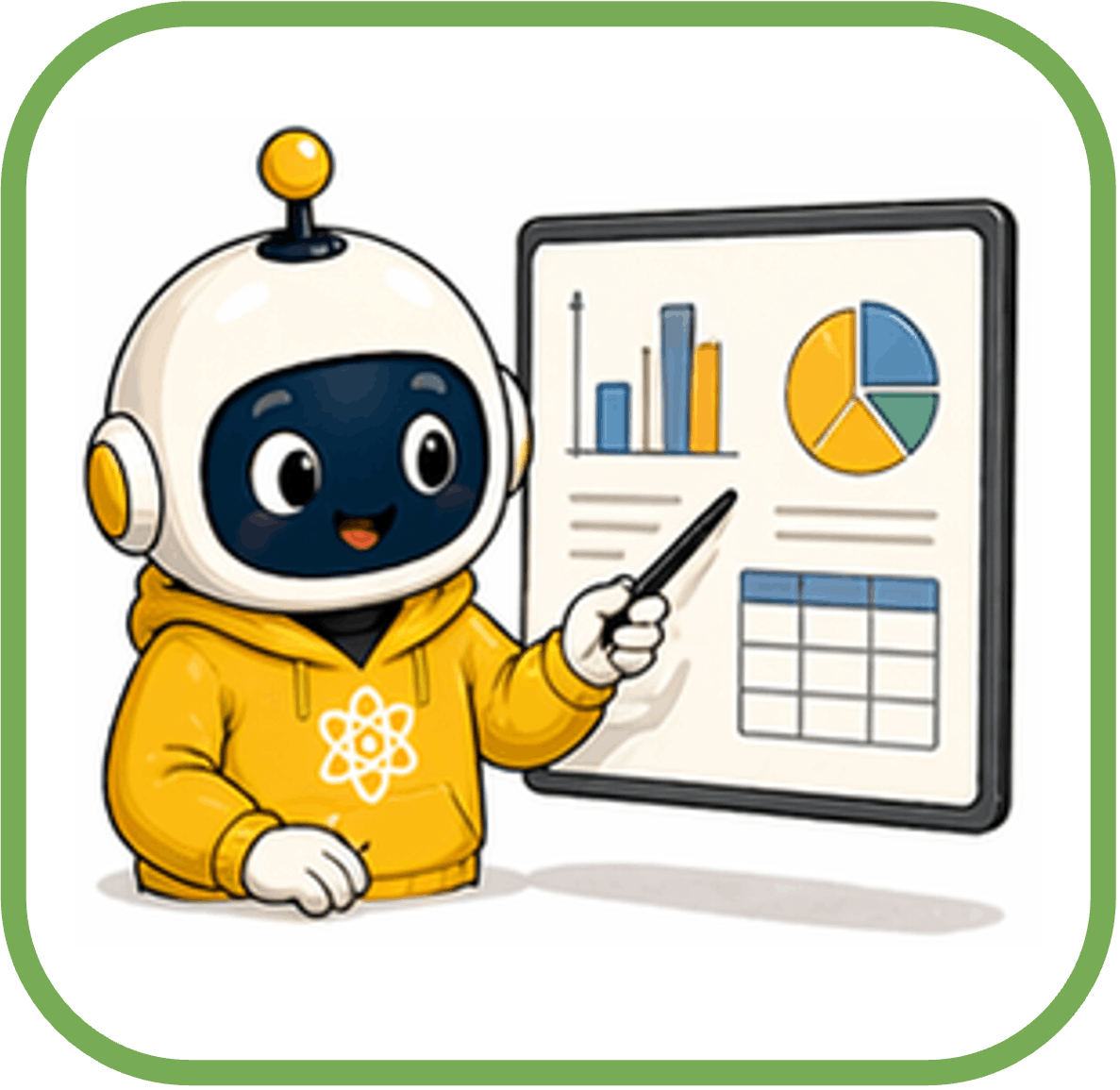}{%
Constructing method diagrams, result plots, comparison tables, mathematical formulas, and algorithmic illustrations. These artifacts transform raw outputs and conceptual designs into structured scientific representations.}

\vspace{6pt}
\noindent\underline{\textbf{Phase 2: Writing.}} This phase organizes the outputs of \emph{Creation} into a formal scholarly manuscript for communication and external scrutiny.

\stagecard{\Sfive}{Paper Writing}{P2color}{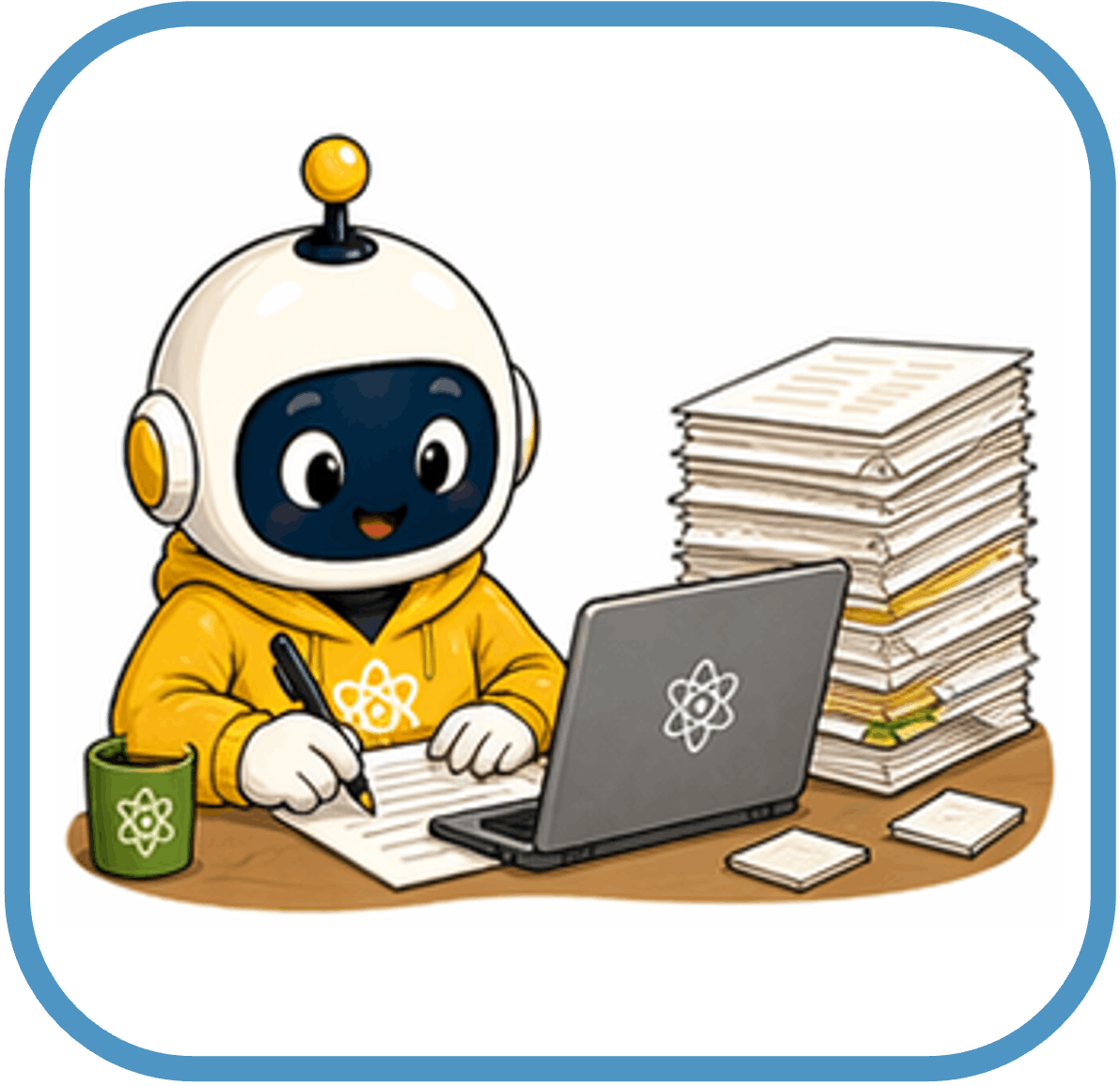}{%
Drafting, editing, polishing, and structuring academic manuscripts. AI assistance ranges from grammar correction and citation support to section-level drafting and full-paper generation.}

\vspace{6pt}
\noindent\underline{\textbf{Phase 3: Validation.}} This phase covers the stages through which the research community scrutinizes, critiques, and iteratively refines a manuscript.

\stagecard{\Ssix}{Peer Review}{P3color}{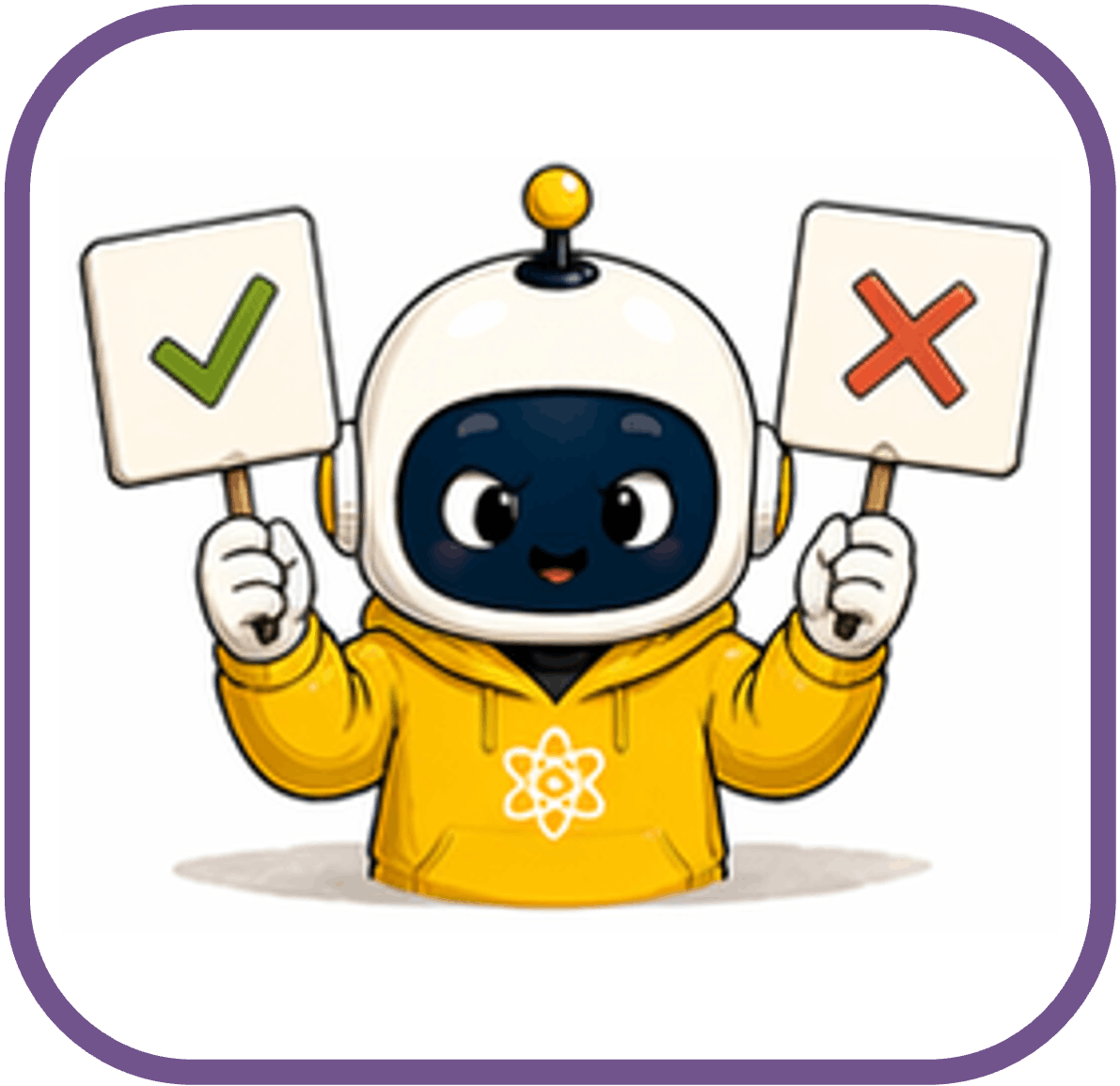}{%
Generating structured reviews, matching reviewers to manuscripts, assessing review quality, and supporting meta-review decisions. These systems aim to assist, rather than replace, the community's evaluative process.}

\stagecard{\Sseven}{Rebuttal \& Revision}{P3color}{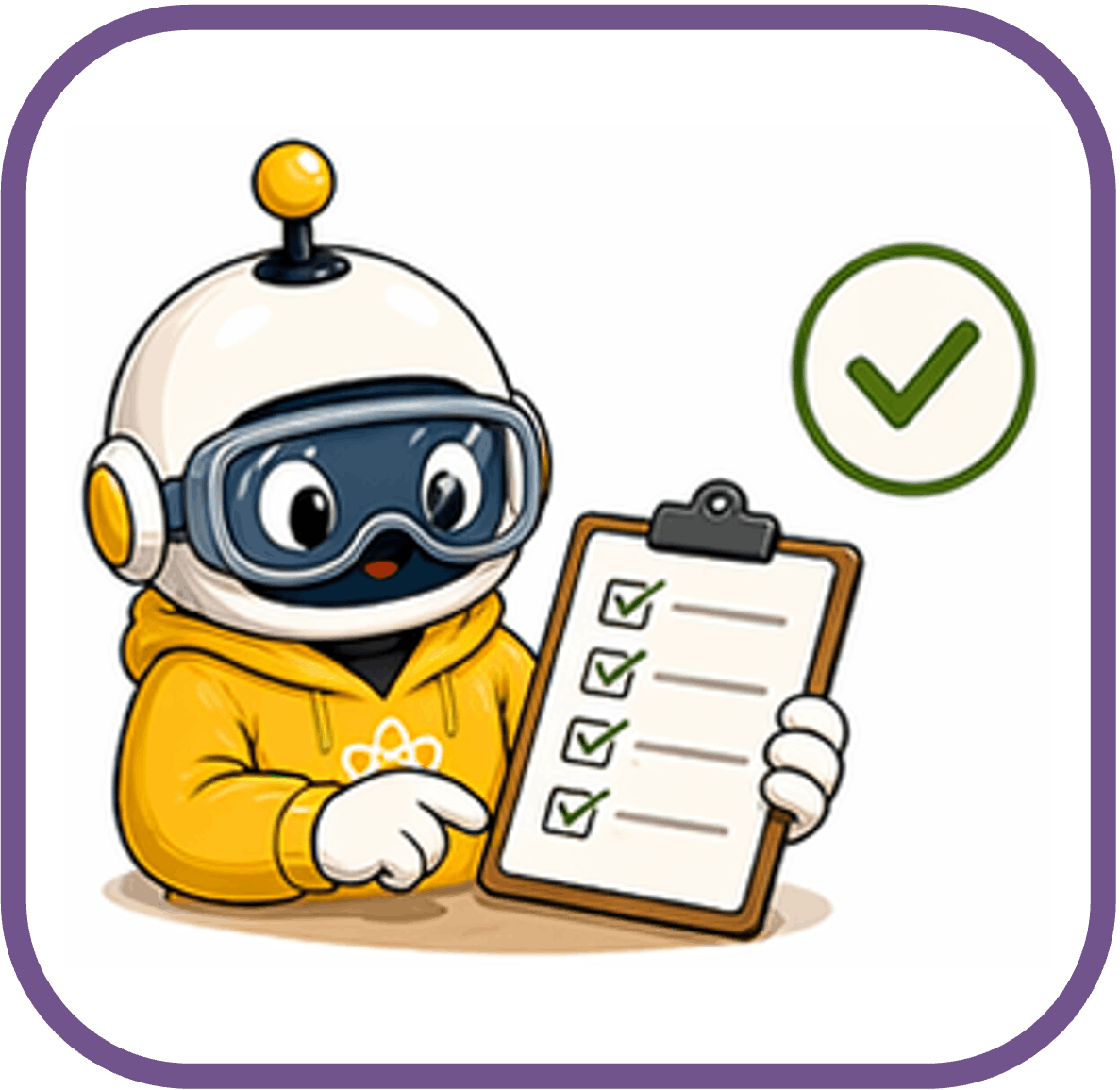}{%
Analyzing reviewer comments, identifying required evidence, drafting responses, and supporting manuscript revision. This stage connects external critique with additional analysis, clarification, and experimental follow-up.}

\vspace{6pt}
\noindent\underline{\textbf{Phase 4: Dissemination.}} This phase converts the manuscript and its supporting materials into formats accessible to broader research and public audiences.

\stagecard{\Seight}{Paper2X}{P4color}{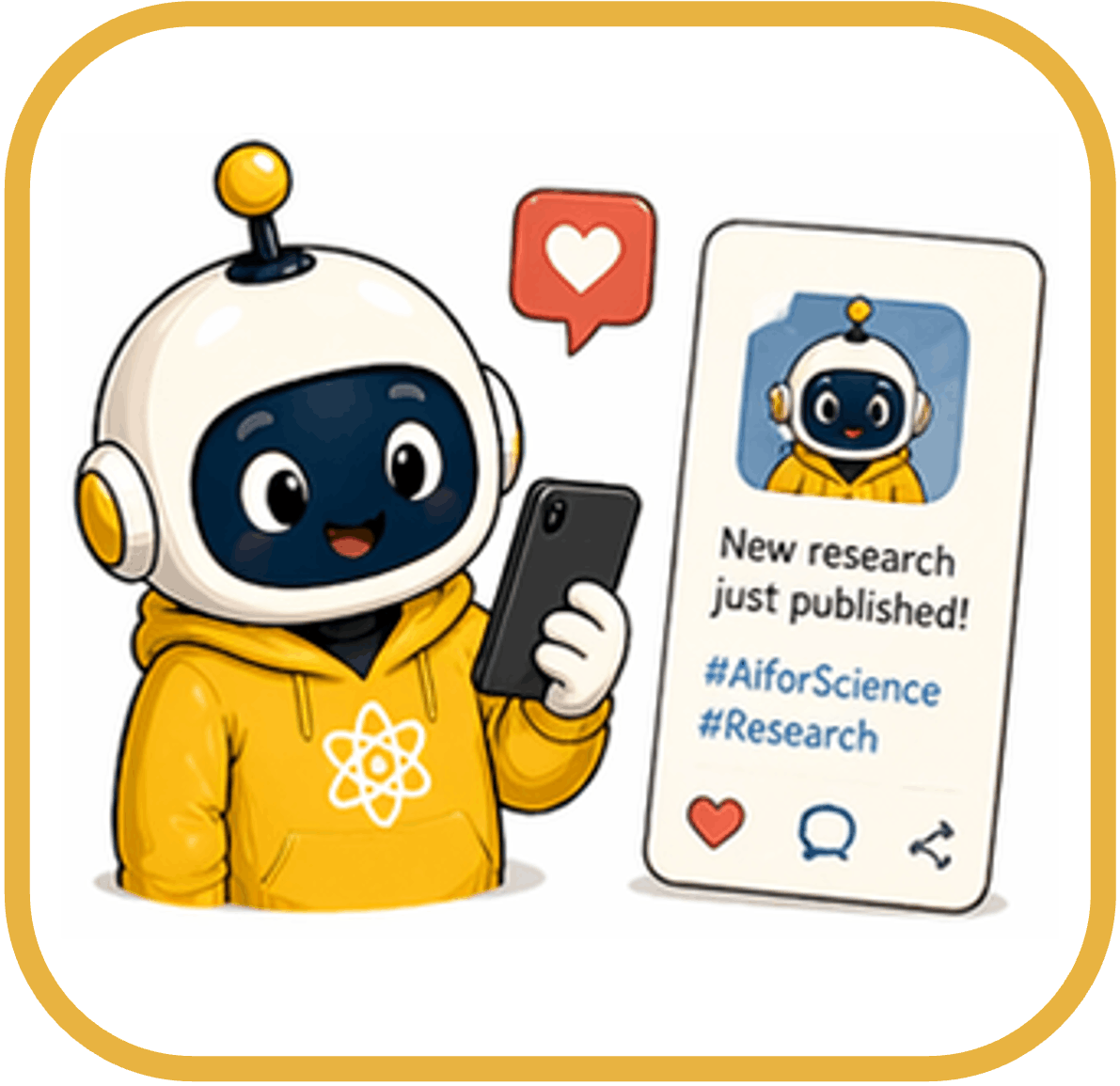}{%
Converting papers into posters, slides, videos, project pages, demos, and social media content. Each output format targets a different audience and requires distinct design choices, fidelity constraints, and communication strategies.}

\vspace{16pt}
Although presented in temporal order, the lifecycle is not strictly linear. Reviewer critiques in Phase~3 (\emph{Validation}) may require returning to Phase~1 (\emph{Creation}) for additional experiments, while dissemination outputs in Phase~4 (\emph{Dissemination}) may expose ambiguities or errors that trigger revisions in Phase~2 (\emph{Writing}). These feedback loops are central to research practice and are especially important for AI-assisted workflows, where errors can propagate across stages if not explicitly checked.

This four-phase grouping reflects the functional structure of research. Evidence and artifacts are produced in \levelbadge{P1color}{P1}~\emph{Creation}, organized into a manuscript in \levelbadge{P2color}{P2}~\emph{Writing}, externally scrutinized in \levelbadge{P3color}{P3}~\emph{Validation}, and communicated to broader audiences in \levelbadge{P4color}{P4}~\emph{Dissemination}. 

We separate \emph{Writing} from \emph{Creation} because manuscript construction is not merely a formatting step: it is a rhetorical and evidential organization process that requires different AI capabilities from those used to produce code, experiments, or figures. We group Peer Review and Rebuttal under \emph{Validation} because together they form the community-facing mechanism through which claims are challenged, defended, and revised. Finally, we treat \emph{Dissemination} as a full phase because posters, slides, videos, project pages, and social media summaries are increasingly important knowledge artifacts with their own fidelity and trust requirements.

\subsection{Methodological Families}
\label{sec:methods_overview}

Across the research lifecycle, AI-assisted research systems reuse a small set of methodological patterns. We group them into five broad families: $^1$\emph{prompt engineering}, $^2$\emph{retrieval-augmented generation (RAG)}, $^3$\emph{training-free agentic methods}, $^4$\emph{training-based methods}, and $^5$\emph{hybrid approaches}. These families are not mutually exclusive or strictly chronological; rather, they describe how current systems elicit, ground, specialize, and orchestrate LLM behavior. Many practical systems combine several of them, for example using prompts for decomposition, RAG for grounding, tools for execution, and trained modules for scoring or ranking.

\textbf{Prompt engineering} provides the simplest interface for adapting general-purpose LLMs to research tasks~\cite{wei2022chain,yao2023react}. It includes direct prompting, chain-of-thought reasoning, role assignment, structured templates, rubric-based instructions, and output constraints. Because it requires no additional training, it remains widely used for lightweight tasks such as brainstorming, editing, review drafting, rebuttal outlining, and social media generation, but it is sensitive to prompt wording and usually lacks persistent grounding.

\textbf{Retrieval-augmented generation (RAG)} grounds model outputs in external sources, including paper corpora, citation graphs, code repositories, benchmark records, and experimental logs~\cite{lewis2020retrieval}. It is especially important for literature review, citation support, evidence checking, rebuttal generation, and stages where source attribution is required. RAG reduces hallucination by exposing models to evidence at inference time, but does not ensure that selected sources are correct, version-consistent, or faithfully represented.

\textbf{Training-free agentic methods} extend LLMs with planning, tool use, memory, self-reflection, and iterative execution, enabling multi-step workflows without updating model parameters~\cite{yao2023react,schick2023toolformer,shinn2023reflexion}. These methods are central to deep literature exploration, code debugging, experiment orchestration, review-response planning, and Paper2X workflows. Their strength lies in orchestration, while their main risk is error propagation when retrieval, tool use, or self-critique fails.

\textbf{Training-based methods} specialize models for stage-specific distributions, such as peer reviews, scientific manuscripts, code repositories, citation contexts, rebuttal traces, or benchmark trajectories~\cite{ouyang2022training,wang2023selfinstruct}. They include supervised fine-tuning, instruction tuning, preference optimization, reinforcement learning, and domain-specific adaptation. They can improve consistency, format adherence, domain vocabulary, and task-specific judgment, but depend heavily on data quality and may overfit to narrow benchmark or venue distributions.

\textbf{Hybrid approaches} combine multiple families into integrated research systems, for example by coupling RAG with agentic planning, fine-tuning domain-specific submodules, or embedding prompt-based controllers inside larger workflows~\cite{lu2024aiscientist,paperqa2024github,shao2024storm,openscholar2025}. Hybrid systems are increasingly dominant because research workflows require generation and grounding, autonomy and verification, and flexible reasoning with stage-specific specialization.

\cref{tab:method_summary} maps these methodological families to the eight lifecycle stages, using primary and secondary markers to indicate common design patterns in recent systems. 

\begin{table*}[!t]
\centering
\caption{Dominant methodological families, representative systems, and research maturity across each stage of the four-phase research lifecycle. Notations: \cmark~= ``primary approach'', \pmark~= ``secondary/emerging'', \xmark~= ``not used''.}
\vspace{-0.2cm}
\label{tab:method_summary}
\renewcommand{\arraystretch}{1.12}
\setlength{\tabcolsep}{2.5pt}
\footnotesize
\begin{tabular}{l|ccccc|l|c}
\toprule
\rowcolor{tableheader!10}
\textbf{Stage} &
\rotatebox{70}{\makecell[l]{\scriptsize Prompt\\[-1pt]\scriptsize Eng.}} &
\rotatebox{70}{\makecell[l]{\scriptsize RAG}} &
\rotatebox{70}{\makecell[l]{\scriptsize Agentic}} &
\rotatebox{70}{\makecell[l]{\scriptsize Training}} &
\rotatebox{70}{\makecell[l]{\scriptsize Hybrid}} &
\textbf{Representative works} &
\textbf{Maturity} \\
\midrule
\rowcolor{P1color!8}
\multicolumn{8}{l}{\textit{\textbf{Phase~1: Creation}}} \\
\Sone: Idea Generation   & \cmark & \pmark & \cmark & \xmark & \pmark & AI Scientist~\cite{lu2024aiscientist}, VirSci~\cite{su2024virsci}, Spark~\cite{sanyal2025spark}                          & \textcolor{P1color}{\ding{72}\ding{72}\ding{72}\ding{72}}\textcolor{gray!30}{\ding{72}} \\
\Stwo: Literature Review  & \pmark & \cmark & \cmark & \xmark & \cmark & PaperQA2~\cite{skarlinski2024paperqa2}, AutoSurvey~\cite{wang2024autosurvey}, STORM~\cite{shao2024storm}                    & \textcolor{P1color}{\ding{72}\ding{72}\ding{72}}\textcolor{gray!30}{\ding{72}\ding{72}} \\
\Sthree: Coding \& Exp.     & \pmark & \pmark & \cmark & \xmark & \cmark & AIDE~\cite{jiang2025aide}, PaperCoder~\cite{papercoder2025}, R\&D-Agent~\cite{chen2025rdagent}                               & \textcolor{P1color}{\ding{72}\ding{72}\ding{72}\ding{72}}\textcolor{gray!30}{\ding{72}} \\
\Sfour: Tables \& Fig.     & \cmark & \xmark & \cmark & \pmark & \xmark & MatPlotAgent~\cite{matplotagent2024}, AutoFigure~\cite{autofigure2026}, DeTikZify~\cite{belouadi2024detikzify}               & \textcolor{P1color}{\ding{72}}\textcolor{gray!30}{\ding{72}\ding{72}\ding{72}\ding{72}} \\
\midrule
\rowcolor{P2color!8}
\multicolumn{8}{l}{\textit{\textbf{Phase~2: Writing}}} \\
\Sfive: Paper Writing      & \cmark & \cmark & \pmark & \cmark & \pmark & CycleResearcher~\cite{cycleresearcher2024}, ScholarCopilot~\cite{scholarcop2025}, XtraGPT~\cite{xtragpt2025}                 & \textcolor{P2color}{\ding{72}\ding{72}\ding{72}\ding{72}}\textcolor{gray!30}{\ding{72}} \\
\midrule
\rowcolor{P3color!8}
\multicolumn{8}{l}{\textit{\textbf{Phase~3: Validation}}} \\
\Ssix: Peer Review        & \pmark & \pmark & \cmark & \cmark & \cmark & DeepReviewer~\cite{deepreviewer2025}, MARG~\cite{darcy2024marg}, ReviewAgents~\cite{reviewagents2025}                        & \textcolor{P3color}{\ding{72}\ding{72}\ding{72}}\textcolor{gray!30}{\ding{72}\ding{72}} \\
\Sseven: Rebuttal           & \pmark & \cmark & \cmark & \xmark & \pmark & RebuttalAgent~\cite{rebuttalagent2026}, Paper2Rebuttal~\cite{paper2rebuttal2026}                                             & \textcolor{P3color}{\ding{72}}\textcolor{gray!30}{\ding{72}\ding{72}\ding{72}\ding{72}} \\
\midrule
\rowcolor{P4color!8}
\multicolumn{8}{l}{\textit{\textbf{Phase~4: Dissemination}}} \\
\Seight: Dissemination      & \cmark & \pmark & \cmark & \xmark & \xmark & Paper2Poster~\cite{paper2poster2025}, PPTAgent~\cite{pptagent2025}, SlideGen~\cite{slidegen2025}                             & \textcolor{P4color}{\ding{72}}\textcolor{gray!30}{\ding{72}\ding{72}\ding{72}\ding{72}} \\
\bottomrule
\end{tabular}
\end{table*}

\subsection{Scope \& Literature Collection}
\label{sec:scope}

This survey focuses on AI tools, methods, and benchmarks that support \emph{human-driven academic research}, with an emphasis on computer science and machine learning. We cover work published or publicly released between 2023 and early 2026, while also referencing earlier foundational methods when they define recurring technical paradigms. Cross-disciplinary systems are included when they demonstrate capabilities relevant to the research lifecycle, such as autonomous experimentation, literature synthesis, scientific coding, or evidence-grounded writing. We exclude general-purpose LLM capabilities that are not explicitly connected to research workflows, as well as closed systems for which insufficient technical or evaluative information is available.

To construct the survey corpus, we combined three complementary collection strategies:
\begin{itemize}
    \item \textbf{Systematic keyword search} across Google Scholar, Semantic Scholar, arXiv, and DBLP, using queries related to AI-assisted research, automated research agents, literature review, scientific coding, paper writing, peer review, rebuttal generation, and research dissemination.

    \item \textbf{Snowball citation tracing} from representative seed papers in each lifecycle stage, including both backward tracing to foundational work and forward tracing to recent systems and benchmarks.

    \item \textbf{Community and repository monitoring}, including open-source projects, curated reading lists, and benchmark leaderboards that document emerging tools not yet covered by formal publications.
\end{itemize}

A paper, system, or benchmark was included only if it satisfied all three criteria: (i) it targets at least one stage of the research lifecycle defined in \cref{sec:lifecycle}; (ii) it is publicly accessible through a publication, preprint, open-source repository, benchmark page, or technical report; and (iii) it provides sufficient methodological or evaluative detail to support critical analysis. When multiple versions of the same system exist, we prioritize the most recent or most technically complete version, while noting earlier versions when they mark important historical milestones.

The resulting corpus spans all four phases of the lifecycle, but the distribution is uneven. Most documented systems concentrate on \levelbadge{P1color}{P1}~(\emph{Creation}), especially literature review, coding, and experiment automation, followed by \levelbadge{P2color}{P2}~(\emph{Writing}), \levelbadge{P3color}{P3}~(\emph{Validation}), and \levelbadge{P4color}{P4}~(\emph{Dissemination}). This imbalance reflects both research maturity and publication availability: creation-stage tools are more frequently benchmarked and open-sourced, whereas dissemination-oriented tools are often commercial, workflow-specific, or evaluated through less standardized criteria. The benchmark landscape across stages is summarized in \cref{tab:benchmarks}.

\begin{table}[!ht]
\centering
\vspace{-0.2cm}
\caption{\textbf{Summary of datasets and benchmarks for AI-assisted research}, organized by phases and stages.}
\vspace{-0.2cm}
\label{tab:benchmarks}
\renewcommand{\arraystretch}{1.15}
\setlength{\tabcolsep}{3pt}
\footnotesize
\resizebox{\linewidth}{!}{\begin{tabular}{c|l|crr|c|c|l|l|c}
\toprule
\textbf{\#} & \textbf{Stage} & \textbf{Benchmark} & \textbf{Ref.} & \textbf{Year} & \textbf{GitHub} & \textbf{HF} & \textbf{Evaluation Focus} & \textbf{Scale} & \textbf{Link}
\\
\midrule\midrule
\multicolumn{10}{@{}l}{\cellcolor{P1color!8}\textbf{\textsf{~~Phase 1: Creation}}} \\
\addlinespace[1pt]
{1} & \Sone: Idea Gen. & IdeaBench & \cite{guo2025ideabench} & 2024 & - & - & Novelty, feasibility & Multiple LLMs & \href{https://arxiv.org/abs/2411.02429}{\faExternalLink} \\
\rowcolor{P1color!6}
{2} & \Sone: Idea Gen. & LiveIdeaBench & \cite{liveideabench2024} & 2024 & - & - & Real-time model comparison & 40+ models & \href{https://arxiv.org/abs/2412.17596}{\faExternalLink} \\
{3} & \Sone: Idea Gen. & AI Idea Bench 2025 & \cite{aiideabench2025} & 2025 & \githubicon{https://github.com/yansheng-qiu/AI_Idea_Bench_2025} & - & Multi-dimensional assessment & 3,495 papers & \href{https://arxiv.org/abs/2504.14191}{\faExternalLink} \\
\rowcolor{P1color!6}
{4} & \Sone: Idea Gen. & ResearchBench & \cite{researchbench2025} & 2025 & - & - & Inspiration-based task decomp. & - & \href{https://arxiv.org/abs/2503.21248}{\faExternalLink} \\
{5} & \Sone: Idea Gen. & Scientist-Bench & \cite{airesearcher2025} & 2025 & - & - & Guided \& open-ended AI research & Multi-domain & \href{https://arxiv.org/abs/2505.18705}{\faExternalLink} \\
\rowcolor{P1color!6}
{6} & \Sone: Idea Gen. & HindSight & \cite{hindsight2026} & 2026 & - & - & Impact-based idea evaluation & - & \href{https://arxiv.org/abs/2603.15164}{\faExternalLink} \\
{7} & \Sone: Idea Gen. & HeurekaBench & \cite{heurekabench2026} & 2026 & \githubicon{https://github.com/mlbio-epfl/HeurekaBench} & - & Open-ended data-driven science & Multi-domain & \href{https://arxiv.org/abs/2601.01678}{\faExternalLink}
\\\midrule
\addlinespace[2pt]
\rowcolor{P1color!6}
{8} & \Stwo: Lit.\ Rev. & LitSearch & \cite{litsearch2024} & 2024 & \githubicon{https://github.com/princeton-nlp/LitSearch} & \hficon{https://huggingface.co/datasets/princeton-nlp/LitSearch} & Literature retrieval & - & \href{https://arxiv.org/abs/2407.18940}{\faExternalLink} \\
{9} & \Stwo: Lit.\ Rev. & DeepScholar-Bench & \cite{deepscholar2025} & 2025 & \githubicon{https://github.com/guestrin-lab/deepscholar-bench} & - & Research synthesis quality & - & \href{https://arxiv.org/abs/2508.20033}{\faExternalLink} \\
\rowcolor{P1color!6}
{10} & \Stwo: Lit.\ Rev. & ReportBench & \cite{reportbench2025} & 2025 & \githubicon{https://github.com/ByteDance-BandAI/ReportBench} & - & Deep research report quality & 100 prompts & \href{https://arxiv.org/abs/2508.15804}{\faExternalLink} \\
{11} & \Stwo: Lit.\ Rev. & ScholarGym & \cite{scholargym2026} & 2026 & - & - & Information-gathering evaluation & 2,536 queries & \href{https://arxiv.org/abs/2601.21654}{\faExternalLink} \\
\rowcolor{P1color!6}
{12} & \Stwo: Lit.\ Rev. & SciNetBench & \cite{scinetbench2026} & 2026 & - & - & Relation-aware retrieval & 18M papers & \href{https://arxiv.org/abs/2601.03260}{\faExternalLink} \\
{13} & \Stwo: Lit.\ Rev. & IDRBench & \cite{idrbench2026} & 2026 & - & - & Interactive deep research & 100 tasks & \href{https://arxiv.org/abs/2601.06676}{\faExternalLink}
\\\midrule
\addlinespace[2pt]
\rowcolor{P1color!6}
{14} & \Sthree: Coding & SWE-bench & \cite{swebench2024} & 2024 & \githubicon{https://github.com/princeton-nlp/SWE-bench} & \hficon{https://huggingface.co/datasets/princeton-nlp/SWE-bench_Verified} & GitHub issue resolution & 500 problems & \href{https://arxiv.org/abs/2310.06770}{\faExternalLink} \\
{15} & \Sthree: Coding & MLAgentBench & \cite{mlagentbench2024} & 2024 & \githubicon{https://github.com/snap-stanford/MLAgentBench} & - & ML experimentation & 13 tasks & \href{https://arxiv.org/abs/2310.03302}{\faExternalLink} \\
\rowcolor{P1color!6}
{16} & \Sthree: Coding & LAB-Bench & \cite{labbench2024} & 2024 & \githubicon{https://github.com/Future-House/LAB-Bench} & \hficon{https://huggingface.co/datasets/futurehouse/lab-bench} & Biology research tasks & Multi-domain & \href{https://arxiv.org/abs/2407.10362}{\faExternalLink} \\
{17} & \Sthree: Coding & DiscoveryBench & \cite{majumder2024discoverybench} & 2024 & \githubicon{https://github.com/allenai/discoverybench} & \hficon{https://huggingface.co/datasets/allenai/discoverybench} & Data-driven discovery & - & \href{https://arxiv.org/abs/2407.01725}{\faExternalLink} \\
\rowcolor{P1color!6}
{18} & \Sthree: Coding & DiscoveryWorld & \cite{discoveryworld2024} & 2024 & \githubicon{https://github.com/allenai/discoveryworld} & - & Virtual discovery environment & 120 tasks & \href{https://arxiv.org/abs/2406.06769}{\faExternalLink} \\
{19} & \Sthree: Coding & MLE-Bench & \cite{chan2024mlebench} & 2024 & \githubicon{https://github.com/openai/mle-bench} & \hficon{https://huggingface.co/datasets/TIGER-Lab/mle-bench} & Kaggle ML competitions & 75 tasks & \href{https://arxiv.org/abs/2410.07095}{\faExternalLink} \\
\rowcolor{P1color!6}
{20} & \Sthree: Coding & ScienceAgentBench & \cite{scienceagentbench2024} & 2024 & \githubicon{https://github.com/OSU-NLP-Group/ScienceAgentBench} & \hficon{https://huggingface.co/datasets/osunlp/ScienceAgentBench} & Scientific data analysis & - & \href{https://arxiv.org/abs/2410.05080}{\faExternalLink} \\
{21} & \Sthree: Coding & KernelBench & \cite{kernelbench2025} & 2025 & \githubicon{https://github.com/ScalingIntelligence/KernelBench} & \hficon{https://huggingface.co/datasets/ScalingIntelligence/KernelBench} & GPU kernel generation & - & \href{https://arxiv.org/abs/2502.10517}{\faExternalLink} \\
\rowcolor{P1color!6}
{22} & \Sthree: Coding & TritonBench & \cite{tritonbench2025} & 2025 & \githubicon{https://github.com/thunlp/TritonBench} & - & Triton operator generation & - & \href{https://arxiv.org/abs/2502.14752}{\faExternalLink} \\
{23} & \Sthree: Coding & ResearchCodeBench & \cite{researchcodebench2025} & 2025 & - & - & Novel ML code implementation & 212 tasks & \href{https://arxiv.org/abs/2506.02314}{\faExternalLink} \\
\rowcolor{P1color!6}
{24} & \Sthree: Coding & SciReplicate-Bench & \cite{scireplicatebench2025} & 2025 & \githubicon{https://github.com/xyzCS/SciReplicate-Bench} & - & Algorithm reproduction & 100 tasks & \href{https://arxiv.org/abs/2504.00255}{\faExternalLink} \\
{25} & \Sthree: Coding & MLR-Bench & \cite{mlrbench2025} & 2025 & - & \hficon{https://huggingface.co/datasets/chchenhui/mlrbench-tasks} & Open-ended ML research & 201 tasks & \href{https://arxiv.org/abs/2505.19955}{\faExternalLink} \\
\rowcolor{P1color!6}
{26} & \Sthree: Coding & MLGym & \cite{mlgym2025} & 2025 & - & - & AI research agent framework & - & \href{https://arxiv.org/abs/2502.14499}{\faExternalLink} \\
{27} & \Sthree: Coding & CURIE & \cite{curie2025} & 2025 & \githubicon{https://github.com/Just-Curieous/Curie} & - & Rigorous experimentation & - & \href{https://arxiv.org/abs/2502.16069}{\faExternalLink} \\
\rowcolor{P1color!6}
{28} & \Sthree: Coding & PaperBench & \cite{paperbench2025} & 2025 & \githubicon{https://github.com/openai/preparedness} & \hficon{https://huggingface.co/datasets/josancamon/paperbench} & Paper replication & 20 ICML papers & \href{https://arxiv.org/abs/2504.01848}{\faExternalLink} \\
{29} & \Sthree: Coding & AstaBench & \cite{astabench2025} & 2025 & \githubicon{https://github.com/allenai/asta-bench} & \hficon{https://huggingface.co/datasets/allenai/asta-bench} & Scientific research suite & 2,400+ problems & \href{https://arxiv.org/abs/2510.21652}{\faExternalLink} \\
\rowcolor{P1color!6}
{30} & \Sthree: Coding & ResearchClawBench & \cite{researchclawbench2025} & 2025 & \githubicon{https://github.com/InternScience/ResearchClawBench} & - & Scientist-aligned workflows & Multi-domain & \href{https://arxiv.org/abs/2512.16969}{\faExternalLink} \\
{31} & \Sthree: Coding & EXP-Bench & \cite{expbench2025} & 2026 & \githubicon{https://github.com/Just-Curieous/Curie/tree/main/benchmark/exp_bench} & \hficon{https://huggingface.co/datasets/Just-Curieous/EXP-Bench} & AI conducting experiments & 461 tasks/51 papers & \href{https://openreview.net/forum?id=KjgyAm383Z}{\faExternalLink} \\
\rowcolor{P1color!6}
{32} & \Sthree: Coding & FrontierScience & \cite{wang2026frontierscience} & 2026 & - & - & Expert-level scientific tasks & Olympiad + PhD & \href{https://arxiv.org/abs/2601.21165}{\faExternalLink} \\
{33} & \Sthree: Coding & PostTrainBench & \cite{posttrainbench2026} & 2026 & \githubicon{https://github.com/aisa-group/PostTrainBench} & \hficon{https://huggingface.co/datasets/aisa-group/PostTrainBench-Trajectories} & LLM post-training automation & - & \href{https://arxiv.org/abs/2603.08640}{\faExternalLink}
\\\midrule
\addlinespace[2pt]
\rowcolor{P1color!6}
{34} & \Sfour: Tab.~\&~Fig. & MatPlotBench & \cite{matplotagent2024} & 2024 & - & - & Data visualization & - & \href{https://arxiv.org/abs/2402.11453}{\faExternalLink} \\
{35} & \Sfour: Tab.~\&~Fig. & PlotCraft & \cite{plotcraft2025} & 2025 & - & - & Complex visualization & 1K tasks & \href{https://arxiv.org/abs/2511.00010}{\faExternalLink} \\
\rowcolor{P1color!6}
{36} & \Sfour: Tab.~\&~Fig. & TeXpert & \cite{texpert2025} & 2025 & - & - & LaTeX code generation & 3 difficulty levels & \href{https://arxiv.org/abs/2506.16990}{\faExternalLink} \\
{37} & \Sfour: Tab.~\&~Fig. & PaperBananaBench & \cite{paperbanana2026} & 2026 & - & - & Scientific illustration quality & 292 test cases & \href{https://arxiv.org/abs/2601.23265}{\faExternalLink} \\
\rowcolor{P1color!6}
{38} & \Sfour: Tab.~\&~Fig. & SciFlow-Bench & \cite{scifigbench2026} & 2026 & - & - & Framework figure evaluation & 500 figures & \href{https://arxiv.org/abs/2602.09809}{\faExternalLink} \\
{39} & \Sfour: Tab.~\&~Fig. & Figure-Bench & \cite{autofigure2026iclr} & 2026 & \githubicon{https://github.com/ResearAI/AutoFigure} & \hficon{https://huggingface.co/datasets/WestlakeNLP/FigureBench} & Text-to-illustration generation & 3,300 pairs & \href{https://arxiv.org/abs/2602.03828}{\faExternalLink}
\\\midrule
\addlinespace[3pt]
\multicolumn{10}{@{}l}{\cellcolor{P2color!8}\textbf{\textsf{~~Phase 2: Writing}}} \\
\addlinespace[1pt]
{40} & \Sfive: Writing & ScholarCopilot & \cite{scholarcop2025} & 2025 & - & - & Citation accuracy & 40.1\% top-1 acc. & \href{https://arxiv.org/abs/2504.00824}{\faExternalLink} \\
\rowcolor{P2color!6}
{41} & \Sfive: Writing & SciIG & \cite{sciig2025} & 2025 & - & - & Introduction writing quality & NAACL/ICLR papers & \href{https://arxiv.org/abs/2508.14273}{\faExternalLink} \\
{42} & \Sfive: Writing & PaperWritingBench & \cite{paperwritingbench2026} & 2026 & - & - & AI paper writing quality & 200 papers & \href{https://arxiv.org/abs/2604.05018}{\faExternalLink}
\\\midrule
\addlinespace[3pt]
\multicolumn{10}{@{}l}{\cellcolor{P3color!8}\textbf{\textsf{~~Phase 3: Validation}}}
\\
\addlinespace[1pt]
{43} & \Ssix: Peer Rev. & ClaimCheck & \cite{claimcheck2025} & 2025 & - & - & Grounded LLM critiques & - & \href{https://arxiv.org/abs/2503.21717}{\faExternalLink} \\
\rowcolor{P3color!6}
{44} & \Ssix: Peer Rev. & Review-CoT & \cite{reviewagents2025} & 2025 & - & - & Review reasoning chains & 142K reviews & \href{https://arxiv.org/abs/2503.08506}{\faExternalLink} \\
{45} & \Ssix: Peer Rev. & AI Detection Bench & \cite{aidetectionreview2025} & 2025 & - & - & AI review detection & 788K reviews & \href{https://arxiv.org/abs/2502.19614}{\faExternalLink}
\\\midrule
\addlinespace[2pt]
\rowcolor{P3color!6}
{46} & \Sseven: Rebuttal & ReviewMT & \cite{reviewmt2024} & 2024 & - & - & Multi-turn review dialogue & 26,841 papers & \href{https://arxiv.org/abs/2406.05688}{\faExternalLink} \\
{47} & \Sseven: Rebuttal & Re$^2$ & \cite{re2dataset2025} & 2025 & - & - & Full-stage review + rebuttal & 19,926 papers & \href{https://arxiv.org/abs/2505.07920}{\faExternalLink} \\
\rowcolor{P3color!6}
{48} & \Sseven: Rebuttal & Commitment Checklist & \cite{rebuttalcommitment2026} & 2026 & - & - & Unfulfilled rebuttal commitments & ICLR 2025 & \href{https://arxiv.org/abs/2603.00003}{\faExternalLink}
\\\midrule
\addlinespace[3pt]
\multicolumn{10}{@{}l}{\cellcolor{P4color!8}\textbf{\textsf{~~Phase 4: Dissemination}}} \\
\addlinespace[1pt]
{49} & \Seight: P2Slides & PPTEval & \cite{pptagent2025} & 2025 & \githubicon{https://github.com/icip-cas/PPTAgent} & - & Slide content, design, coherence & 10K+ presentations & \href{https://arxiv.org/abs/2501.03936}{\faExternalLink} \\
\rowcolor{P4color!6}
{50} & \Seight: P2Video & PresentQuiz & \cite{paper2video2025} & 2025 & \githubicon{https://github.com/showlab/Paper2Video} & - & Video faithfulness & 101 paper-video pairs & \href{https://arxiv.org/abs/2510.05096}{\faExternalLink}
\\\midrule
\addlinespace[3pt]
\multicolumn{10}{@{}l}{\cellcolor{tableheader!5}\textbf{\textsf{~~Cross-Phase}}} \\
\addlinespace[1pt]
{51} & Cross-Phase & RE-Bench & \cite{rebench2024} & 2024 & \githubicon{https://github.com/METR/RE-Bench} & - & Open-ended ML R\&D & 7 environments & \href{https://arxiv.org/abs/2411.15114}{\faExternalLink} \\
\rowcolor{tableheader!6}
{52} & Cross-Phase & PaperBench & \cite{paperbench2025} & 2025 & \githubicon{https://github.com/openai/preparedness} & \hficon{https://huggingface.co/datasets/josancamon/paperbench} & End-to-end paper replication & 20 ICML papers & \href{https://arxiv.org/abs/2504.01848}{\faExternalLink} \\
\bottomrule
\end{tabular}}
\vspace{-0.7cm}
\end{table}

\subsection{Development Timeline}
\label{sec:timeline}

The development of AI-assisted research can be understood as a shift from \emph{stage-specific assistance} toward \emph{multi-stage research automation}. Before 2024, most systems targeted isolated research tasks, such as literature search, scientific question answering, code generation, or domain-specific experiment planning. Early demonstrations, including Coscientist~\cite{boiko2023coscientist}, showed that LLM-based agents could plan and execute scientific workflows in constrained laboratory settings, while domain foundation models such as AlphaFold~3~\cite{abramson2024alphafold3} illustrated the broader potential of AI systems to transform specialized scientific discovery.

In 2024, the field began moving from isolated tools toward end-to-end research agents. The AI Scientist~\cite{lu2024aiscientist} provided an early demonstration of an automated pipeline spanning idea generation, experiment execution, paper writing, and review-style evaluation. Around the same period, general coding agents, retrieval-augmented literature systems, and scientific reasoning benchmarks matured rapidly, making it possible to evaluate individual components of the research lifecycle more systematically. This transition marked an important change in emphasis: AI systems were no longer viewed only as assistants for local tasks, but increasingly as orchestrators of multi-step research workflows.

By 2025 and early 2026, the field entered a stage of rapid specialization and benchmarking. Dedicated systems emerged for nearly every lifecycle stage, including literature synthesis, paper-to-code translation, autonomous experiment orchestration, manuscript writing, peer review, rebuttal support, figure generation, and research dissemination. For example, OpenScholar~\cite{openscholar2025} advanced retrieval-augmented scientific synthesis, AI Scientist v2~\cite{yamada2025aiscientistv2} explored stronger forms of end-to-end automated research, and FARS~\cite{fars2026_report} demonstrated large-scale autonomous paper generation. At the same time, previously underexplored stages began receiving dedicated attention, including rebuttal writing (\eg, RebuttalAgent~\cite{rebuttalagent2026}) and scientific visualization (\eg, AutoFigure-Edit~\cite{autofigure2026}). 

These developments suggest that the field is no longer bottlenecked by model capability alone, but also by orchestration, evaluation, reliability, and governance across the full research lifecycle.

\clearpage\clearpage
\section{Phase 1: Creation}
\label{sec:creation}

This phase covers the stages through which a research contribution is materially produced: generating an idea (\Sone), situating it within prior work (\Stwo), producing empirical or analytical evidence (\Sthree), and constructing visual representations of methods and results (\Sfour). Together, these stages address two foundational questions: \emph{what is the contribution, and what evidence supports it?}

Among the four phases, \emph{Creation} currently has the richest tool ecosystem and broadest benchmark coverage, but its maturity remains uneven. \Sone (\emph{Idea Generation}) has attracted extensive tooling, yet suffers from an ideation--execution gap in which seemingly novel ideas often weaken after implementation. \Stwo (\emph{Literature Review}) is rapidly improving through retrieval-augmented and agentic synthesis, but citation fidelity, coverage completeness, and multi-paper relational reasoning remain difficult. \Sthree (\emph{Coding and Experiments}) has progressed through code generation, paper-to-code translation, and autonomous experiment orchestration, but performance still drops sharply on genuinely novel research code. \Sfour (\emph{Tables and Figures}) remains comparatively underdeveloped despite its importance in daily research practice. We discuss these four stages in order below.

\subsection{Idea Generation}
\label{sec:ideation}

Idea generation is the entry point of the research lifecycle, where candidate hypotheses, research questions, and experimental directions are proposed and refined. Existing approaches range from direct LLM prompting to externally grounded generation, multi-agent collaboration, and dedicated evaluation of novelty, feasibility, diversity, and downstream impact. Across these directions, the central challenge is that LLMs can produce ideas that appear novel and well-motivated, yet often struggle to generate ideas that remain feasible, distinctive, and impactful after execution. What makes this stage distinctive is that, unlike later stages whose outputs can be checked against code, data, or reviewer feedback, the value of an idea is only weakly observable at the moment it is proposed: it depends on experiments not yet run and on a research context that no single prompt fully captures. Idea generation is therefore simultaneously the most heavily tooled and the least verifiable stage of \emph{Creation}, which is why much of the recent progress lies less in producing more ideas than in grounding and assessing them.

We organize methods into four groups that trace this progression: generation from the model's internal parametric knowledge (\cref{sec:idea_internal}), generation anchored in external signals such as knowledge graphs, retrieved literature, and research trends (\cref{sec:idea_external}), multi-agent collaborative generation that simulates community-style critique and debate (\cref{sec:idea_multiagent}), and the dedicated assessment of idea quality along novelty and feasibility (\cref{sec:idea_eval}). A comprehensive inventory of ideation systems is provided in \cref{tab:appendix_s1} (Appendix).

\subsubsection{LLM Internal Knowledge-Based Generation}
\label{sec:idea_internal}

The simplest form of AI-assisted ideation prompts an LLM directly with a research domain, problem description, or literature context. Si~\etal~\cite{si2024ideas} established an influential baseline through a large-scale human study involving $100+$ NLP researchers, finding that LLM-generated ideas were rated significantly higher in novelty than human ideas ($p<0.05$). This result demonstrates the surface-level generative capacity of LLMs, but it also raises a central question for this stage: whether apparent novelty corresponds to executable and impactful research.
Subsequent work has explored three ways to strengthen direct generation. First, \emph{iterative refinement} uses feedback loops to improve idea specificity and reduce shallow novelty. ResearchAgent~\cite{baek2024researchagent} incorporates academic graph feedback to refine generated ideas, SciMON~\cite{wang2024scimon} iteratively compares candidate ideas against prior work to mitigate the tendency of direct LLM prompting toward shallow contributions, and Chain of Ideas~\cite{li2024chainofideas} organizes literature into progressive reasoning chains that outperform simple prompting baselines. 

Second, \emph{learned quality signals} introduce explicit scoring or optimization objectives. Spark~\cite{sanyal2025spark} combines retrieval-augmented generation with a judge model trained on $600$K OpenReview reviews to estimate creativity, DeepInnovator~\cite{deepinnovator2026} trains a $14$B model under a ``Next Idea Prediction'' paradigm and reports $80$--$94\%$ win rates against frontier models on ideation tasks, and Goel~\etal~\cite{rubricrewards2025} optimize AI co-scientist plans using rubric rewards extracted from existing papers, with RL-optimized plans preferred by human experts $70\%$ of the time. 

Third, \emph{adaptive test-time compute} treats reasoning effort as a controllable resource. IRIS~\cite{iris2025} uses MCTS in a human-in-the-loop ideation platform to allocate search as ideas converge, while FlowPIE~\cite{flowpie2026} evolves scientific ideas at test time through flow-guided literature exploration. 

A recent creativity-centered survey~\cite{shahhosseini2025ideationsurvey} further categorizes these methods into knowledge augmentation, prompt steering, inference-time scaling, multi-agent collaboration, and parameter adaptation.

\subsubsection{External Signal-Driven Generation}
\label{sec:idea_external}

Direct LLM generation is limited by the model's parametric knowledge and by its tendency to produce plausible but weakly grounded ideas. External signal-driven methods address this limitation by anchoring ideation in structured knowledge, retrieved literature, or temporal research trends. Three signal sources are especially common, each grounding ideas from a different perspective: relational structure, textual evidence, and temporal opportunity.

\emph{Knowledge graphs} provide relational structure for hypothesis formation. SciAgents~\cite{ghafarollahi2024sciagents} performs multi-agent reasoning over scientific knowledge graphs, while MOOSE-Chem~\cite{yang2024moosechem} decomposes chemistry hypothesis generation into inspiration retrieval, hypothesis composition, and ranking, rediscovering hypotheses from $51$ high-impact papers. MOOSE-Chem2~\cite{yang2025moosechem2} extends this direction toward fine-grained, experimentally actionable hypotheses. \emph{Paper retrieval} grounds ideas in unstructured literature. SciPIP~\cite{wang2024scipip} proposes ideas anchored to retrieved papers, and IdeaSynth~\cite{pu2025ideasynth} represents idea facets as nodes on an interactive canvas for literature-grounded refinement; in a user study with $20$ participants, IdeaSynth encouraged users to explore more alternatives than LLM-only baselines. \emph{Trend analysis} targets the temporal dimension of research opportunity. Nova~\cite{hu2024nova} uses iterative planning and search to identify emerging research directions with improved diversity. Together, these methods suggest that external grounding is not merely an auxiliary feature, but a key mechanism for connecting generated ideas to the research frontier.

\subsubsection{Multi-Agent Collaborative Generation}
\label{sec:idea_multiagent}

Multi-agent ideation systems attempt to improve idea quality by simulating aspects of research-community interaction, such as role specialization, critique, revision, and debate. VirSci~\cite{su2024virsci} constructs a virtual scientific community in which multiple LLM agents participate in structured discussions, reporting higher novelty scores than a single-agent AI Scientist baseline ($5.24$ vs.\ $4.94$). Its analysis suggests that agent diversity and discussion structure matter, with the best configuration using $8$ members over $5$ rounds with $50\%$ diversity.

However, multi-agent scaling is not uniformly beneficial. A SIGDIAL 2025 study~\cite{sigdial2025multiagent} finds that three critique--revision rounds are often sufficient, while additional rounds produce diminishing returns. Other systems explore richer collaboration mechanisms beyond discussion alone: Gu~\etal~\cite{gu2024combinatorial} study \emph{combinatorial creativity} by composing ideas across domains, and Deep Ideation~\cite{zhao2025deepideation} designs agents that navigate scientific concept networks through structured graph exploration. 

Yet recent evidence also points to a deeper limitation: the ``Artificial Hivemind'' study~\cite{jiang2025artificialhivemind} reports that LLM-generated ideas tend to cluster in narrow regions of the idea space, suggesting that diversity collapse may be a structural property of current models rather than a problem solved simply by adding more agents. In response, recent methods target diversity explicitly: \emph{Towards Diverse Scientific Hypothesis Search}~\cite{diversehypothesis2026} casts hypothesis generation as a diversity-aware search problem, using population-based sampling to escape these narrow modes across molecular, equation, and algorithm discovery. Complementing diversity-oriented search, auditable hypothesis-evolution protocols~\cite{takahara2026auditable} make the iterative proposal and refinement of hypotheses by LLM agents transparent and inspectable.

\subsubsection{Assessment: Novelty and Feasibility}
\label{sec:idea_eval}

Evaluating generated ideas is difficult because strong research ideas must satisfy multiple criteria simultaneously: novelty, feasibility, clarity, significance, and eventual impact. Early benchmarks quantify parts of this space, but the central question is whether an idea remains valuable after it is implemented, tested, and situated against prior work.

IdeaBench~\cite{guo2025ideabench} evaluates idea generation against $2{,}374$ influential papers across eight research domains, while LiveIdeaBench~\cite{liveideabench2024} probes scientific creativity using $1{,}180$ keyword prompts across $22$ domains. Both suggest that scientific creativity is not well predicted by general-purpose benchmarks, with reasoning-focused models often performing better. ResearchBench~\cite{researchbench2025} extends evaluation through inspiration-based task decomposition, and AI Idea Bench 2025~\cite{aiideabench2025} scales assessment to $3{,}495$ papers across two evaluation axes.

A recurring pattern across these benchmarks is the gap between apparent novelty and practical feasibility. IdeaBench reports that many LLMs score above $0.6$ on novelty but below $0.5$ on feasibility~\cite{guo2025ideabench}, indicating that generating plausible-sounding ideas remains easier than generating ideas that can be executed and validated. HindSight~\cite{hindsight2026} sharpens this concern by introducing a time-split, impact-based evaluation, showing that LLM-as-Judge can overvalue novel-sounding ideas that do not later materialize into impactful work. This finding suggests that current evaluation protocols may reward apparent novelty rather than genuine research potential, reinforcing the need for execution-grounded and temporally robust assessment. Two recent benchmarks question whether models can perform this assessment at all: SoundnessBench~\cite{soundnessbench2026} tests whether AI-scientist systems can distinguish methodologically sound proposals from flawed ones, while a complementary analysis~\cite{llmjudgenovelty2026} characterizes the limits of LLM-as-Judge for novelty assessment; both report a pervasive optimism bias that undermines the use of LLMs as first-pass idea evaluators.

\subsubsection{Findings and Observations}
\label{sec:s1_findings}

\stageanalysis{~Stage 1: Idea Generation}{S1color}{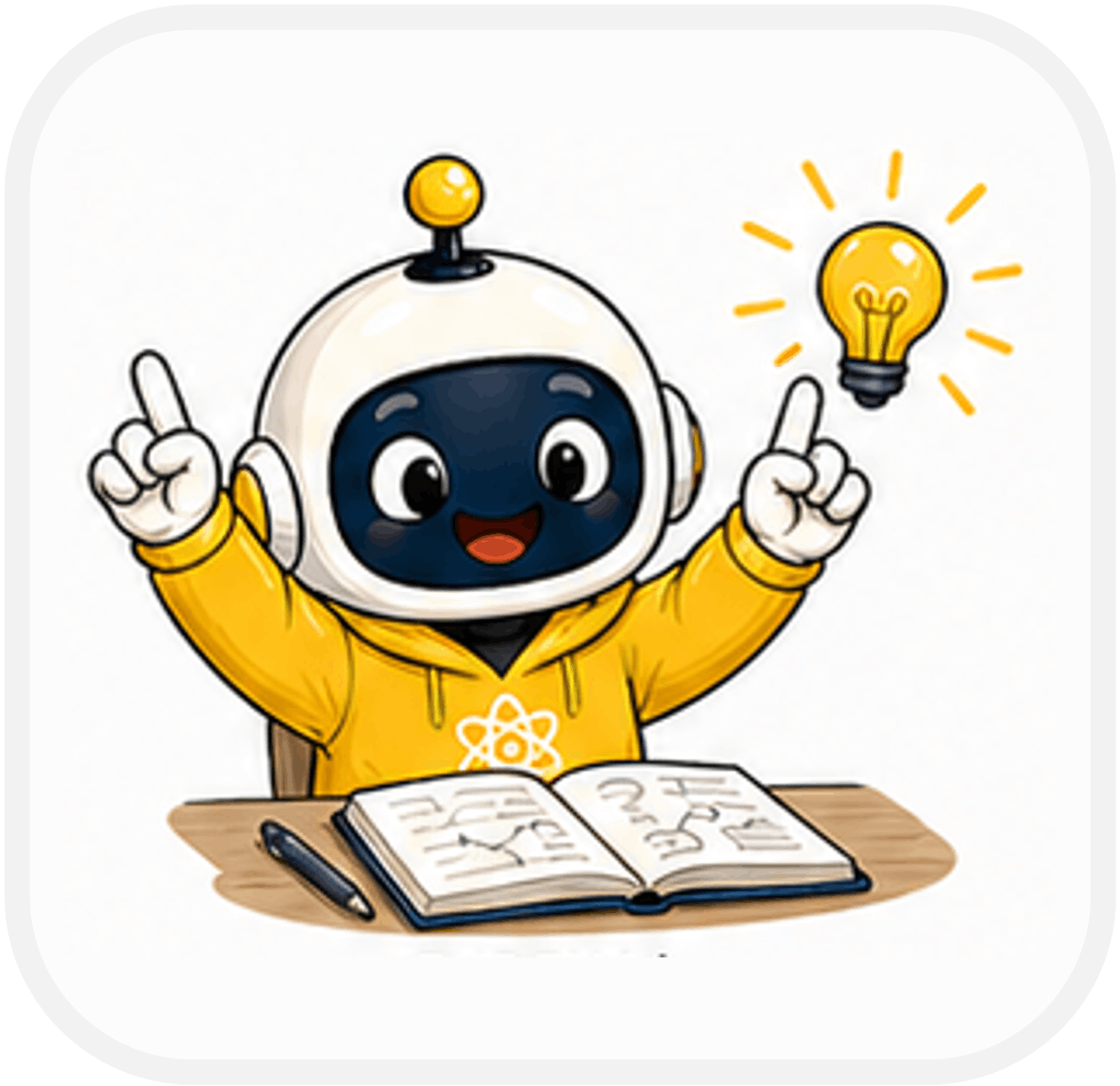}{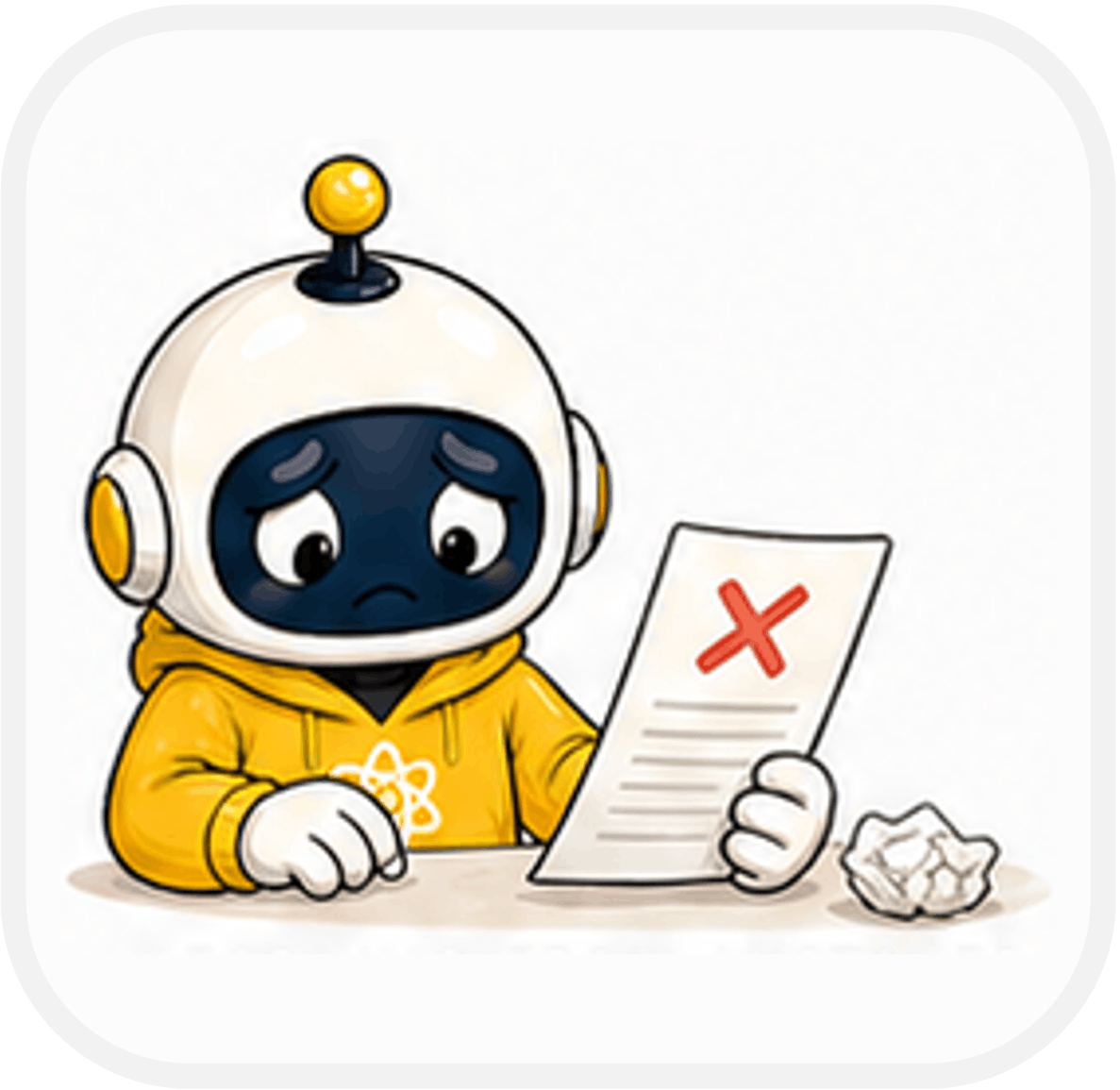}{%
\posbadge{Maturity} \posbadge{Progression} \posbadge{Grounding}\par\vspace{2pt}
\posbadge{Collaboration} \posbadge{Training} \posbadge{Benchmarks}
}{%
\begin{itemize}[leftmargin=8pt, itemsep=0pt, topsep=0pt, parsep=0pt]
\item Idea generation is one of the most tool-rich stages in Phase~1 (\emph{Creation}), with systems spanning prompting, retrieval, multi-agent collaboration, learned scoring, and test-time search.
\end{itemize}
\anadotrule
\begin{itemize}[leftmargin=8pt, itemsep=0pt, topsep=0pt, parsep=0pt]
\item Clear capability progression: prompting $\to$ RAG $\to$ multi-agent $\to$ RL-trained, each generation addressing the weaknesses of its predecessor.
\end{itemize}
\anadotrule
\begin{itemize}[leftmargin=8pt, itemsep=0pt, topsep=0pt, parsep=0pt]
\item External grounding is increasingly central: retrieval- and knowledge-graph-based methods better connect generated ideas to the research frontier than LLM-only prompting~\cite{yang2024moosechem,wang2024scipip}.
\end{itemize}
}{%
\negbadge{Execution} \negbadge{Feasibility} \negbadge{Diversity}\par\vspace{2pt}
\negbadge{Mis-Evaluation} \negbadge{Shallow} \negbadge{Closed-Loop}
}{%
\begin{itemize}[leftmargin=8pt, itemsep=0pt, topsep=0pt, parsep=0pt]
\item Ideas that score well before implementation can degrade substantially after execution ($\Delta=-1.98$ vs.\ $-0.63$ for human ideas~\cite{si2025gap}), exposing a gap between surface novelty and executable substance.
\end{itemize}
\anadotrule
\begin{itemize}[leftmargin=8pt, itemsep=0pt, topsep=0pt, parsep=0pt]
\item Persistent novelty-feasibility tradeoff ($>0.6$ \emph{vs.} $<0.5$~\cite{guo2025ideabench}) remains unresolved, and diversity collapse is structural, not solvable by scaling~\cite{jiang2025artificialhivemind}.
\end{itemize}
\anadotrule
\begin{itemize}[leftmargin=8pt, itemsep=0pt, topsep=0pt, parsep=0pt]
\item LLM-as-Judge evaluation can reward apparent rather than genuine innovation, with reported novelty judgments negatively correlating with later real-world impact ($\rho=-0.29$~\cite{hindsight2026}).
\end{itemize}
}
\vspace{8pt}

\subsection{Literature Review}
\label{sec:literature_review}

Literature review anchors research in prior knowledge by retrieving relevant work, synthesizing evidence, and organizing existing findings into a coherent intellectual context. Compared with idea generation, this stage is more grounded and externally verifiable, making it one of the fastest-maturing areas in AI-assisted research. Existing systems have moved from semantic paper retrieval to citation-aware synthesis and long-horizon deep research agents. Yet two limitations remain central: systems can retrieve and summarize individual papers increasingly well, but still struggle with faithful citation, coverage completeness, and multi-paper relational reasoning. These limitations carry outsized consequences because literature review sits upstream of the rest of the lifecycle: a missed competing method, a misattributed result, or an incomplete account of prior work propagates silently into ideation, writing, and even the author's later response to reviewers. For this stage, then, the binding constraint is not fluency but \emph{faithfulness}---whether each surfaced claim is correctly attributed to a source that genuinely supports it.

We follow the natural pipeline of AI-assisted reviewing: retrieving candidate papers from large scientific corpora (\cref{sec:lit_retrieval}), synthesizing them into survey or related-work narratives (\cref{sec:related_work}), and the long-horizon deep research agents (\cref{sec:deep_research}) that increasingly fold retrieval and synthesis into a single iterative loop; we then examine how retrieval and synthesis quality are assessed (\cref{sec:lit_eval}). A comprehensive inventory of literature review systems is provided in \cref{tab:appendix_s2} (Appendix).

\subsubsection{Literature Retrieval}
\label{sec:lit_retrieval}

Retrieval is the foundation of AI-assisted literature review: every downstream synthesis depends on whether the system can surface the right papers from scientific corpora that now contain tens of millions of entries. Existing methods can be grouped into three modes. \emph{Semantic retrieval} forms the baseline, using dense representations and LLM-based query understanding to move beyond keyword matching. LitLLM~\cite{agarwal2024litllm} integrates LLMs with academic databases for dense retrieval, while PaperQA2~\cite{skarlinski2024paperqa2} extends this direction with citation verification and reports strong performance on scientific literature search.

\emph{Citation-graph-augmented retrieval} adds structural signals beyond embeddings. Instead of treating papers as isolated documents, these methods use citation links, paper relations, and graph traversal to improve contextual coverage. OpenResearcher~\cite{li2024openresearcher} combines RAG with graph traversal for accelerated literature exploration. \emph{Agentic multi-step retrieval} further shifts retrieval from a one-shot ranking problem to an iterative search process. PaSa~\cite{pasa2025} deploys an LLM agent that issues follow-up queries and refines candidate sets, approximating how human researchers probe an unfamiliar topic. Pushing this agentic view further, self-evolving retrieval agents~\cite{selfevolveretrieval2026} continually adapt their own search policy rather than relying on a fixed retrieval strategy. Alongside these methods, dedicated benchmarks have emerged to audit retrieval quality: LitSearch~\cite{litsearch2024} targets retrieval precision, CiteME~\cite{citeme2024} focuses on citation fidelity, and MasterSet~\cite{masterset2026} provides a large-scale benchmark for must-cite citation recommendation in the AI/ML literature. Together, these efforts show that finding relevant papers is becoming easier, but ensuring that retrieved papers are used faithfully remains difficult.

\subsubsection{Survey and Related Work Generation}
\label{sec:related_work}

Synthesis transforms retrieved papers into structured narratives. This marks a shift from retrieval-oriented systems, which optimize paper ranking and coverage, to generation-oriented systems, which must identify themes, compare methods, expose contradictions, and articulate research gaps. The subfield has developed through several increasingly structured designs.

\emph{Single-pass systems} established the feasibility of automated survey drafting. AutoSurvey~\cite{wang2024autosurvey} demonstrated that LLMs can generate surveys of reasonable quality end-to-end, while SurveyX~\cite{liang2025surveyx} improved content quality and approached human-expert performance on selected dimensions. \emph{Structure-aware systems} then elevated outline planning from a formatting step to a core synthesis artifact. STORM~\cite{shao2024storm} introduces multi-perspective question-asking to build comprehensive topic outlines, and SurveyForge~\cite{gao2025surveyforge} learns outline heuristics from human-written surveys together with memory-driven content generation, outperforming AutoSurvey on outline quality.

\emph{Multi-agent decomposition} separates retrieval, verification, organization, and narrative writing into specialized subtasks. LiRA~\cite{lira2025} and Agentic AutoSurvey~\cite{agenticautosurvey2025} employ dedicated agents for different roles, while IterSurvey~\cite{itersurvey2025} treats outline generation as an iterative planning problem with stability checks. InteractiveSurvey~\cite{interactivesurvey2025} further introduces user customization, allowing researchers to refine reference categorization and outline structure through an interactive interface. STRUCTSURVEY~\cite{pedinotti2026structsurvey} pushes the agentic direction further, coupling structured retrieval to outline planning and evidence gathering for automated survey-paper generation.

\emph{Citation- and editor-aware systems} close the loop between synthesis and the writing environment. SurveyG~\cite{surveyg2025} constructs a three-layer citation graph (Foundation/Development/Frontier) with hierarchical traversal, Citegeist~\cite{beger2025citegeist} builds a dynamic RAG pipeline on the arXiv corpus, and CiteLLM~\cite{citellm2026} embeds hallucination-free reference discovery directly inside a LaTeX editor. DeepSurvey~\cite{deepsurvey2026} pushes this direction further, coupling full-text key-point extraction with evidence-constrained citation assignment to improve both analytical depth and citation reliability. Open-source systems such as GPT Researcher~\cite{gptresearcher2024}, PaperQA~\cite{paperqa2024github}, and ChatPaper~\cite{chatpaper2023} further illustrate the growing practical adoption of literature synthesis tools beyond controlled research prototypes. However, citation fidelity remains a bottleneck: ScholarCopilot~\cite{scholarcop2025} reports only $40.1\%$ top-$1$ citation accuracy, suggesting that generating plausible related-work text is still easier than grounding each claim in the correct source.

\subsubsection{Deep Research Agents}
\label{sec:deep_research}

Deep research agents differ from single-pass retrieval or survey-generation systems by treating literature exploration as an \emph{iterative, agentic} process. Given an open-ended query, they plan sub-queries, retrieve and read sources, update their internal state, and continue until a report can be synthesized with sufficient confidence. This loop makes deep research agents closer to a workflow for long-horizon information seeking than a single retrieval model.

\emph{Commercial systems} have popularized this paradigm for broad information synthesis. OpenAI Deep Research, Google Deep Research, Perplexity, and Elicit all support multi-source retrieval and report generation, though they differ in latency, citation style, interactivity, and target use cases. \emph{Open-source literature-specific systems} adapt this paradigm to scientific research. OpenScholar~\cite{openscholar2025}, published in Nature, is a retrieval-augmented LM that searches large-scale open-access scientific corpora and outperforms PaperQA2 and Perplexity Pro on scientific literature benchmarks. Tongyi DeepResearch~\cite{tongyi2025deepresearch} from Alibaba is an agentic LLM specialized for long-horizon deep information seeking, achieving strong results on deep research benchmarks.

\emph{Training-era approaches} target the data and optimization bottlenecks that limit long-horizon research agents. O-Researcher~\cite{oresearcher2026} combines multi-agent distillation with agentic reinforcement learning to improve benchmark performance, while OpenResearcher~\cite{li2026openresearcher} addresses the trajectory-data bottleneck by constructing an offline trajectory synthesis pipeline over large document collections. These synthesized trajectories provide long-horizon tool-use supervision for training research agents. \emph{Domain-focused variants} remain important for specialized synthesis tasks: CHIME~\cite{kang2024chime} provides LLM-assisted hierarchical organization of scientific studies, and ASReview~\cite{asreview2020}, published in Nature Machine Intelligence, uses active-learning-based screening to reduce manual effort in systematic reviews while maintaining recall. Collectively, deep research agents span a spectrum from lightweight factual lookup to long-horizon autonomous synthesis, but increasingly converge on the same iterative architecture: plan, retrieve, read, update, and synthesize.

\subsubsection{Assessment: Retrieval and Synthesis Quality}
\label{sec:lit_eval}

Evaluation has shifted from \emph{retrieval accuracy} alone (``did the system find the right papers?'') toward broader \emph{synthesis quality} (``did it produce a useful, accurate, and well-organized review?''). 

At the output level, DeepScholar-Bench~\cite{deepscholar2025} establishes a dedicated benchmark for research synthesis across coverage, coherence, and factual accuracy. ReportBench~\cite{reportbench2025} scales this direction to deep research reports derived from survey-style prompts.
At the process level, ScholarGym~\cite{scholargym2026} isolates the information-gathering stage of deep research by decomposing it into query planning, tool invocation, and relevance assessment. This is an early step toward evaluating \emph{how} a system reaches its answer, rather than judging only the final output. Benchmarks have also begun probing structural and interactive dimensions of literature competence. SciNetBench~\cite{scinetbench2026} introduces a relation-aware benchmark for literature retrieval agents over large-scale AI literature, revealing that relation-aware retrieval accuracy often remains low. IDRBench~\cite{idrbench2026} addresses the human-in-the-loop dimension through interactive deep research tasks with on-demand user interaction. AutoResearchBench~\cite{autoresearchbench2026} complements these by benchmarking agents on complex, multi-hop scientific literature discovery.

Across these efforts, four evaluation dimensions have crystallized: 
\begin{itemize}
    \item \emph{Citation Accuracy}, whether the references in the paper are correctly attributed and faithfully support the associated claims.

    \item \emph{Coverage Completeness}, whether the review captures the relevant landscape without major omissions.

    \item \emph{Narrative Coherence}, whether the synthesis has logical flow, thematic organization, and readability.

    \item \emph{Factual Grounding}, whether claims are supported by cited evidence rather than hallucinated. 
\end{itemize}

SurveyX~\cite{liang2025surveyx} exemplifies this multi-dimensional view by evaluating content quality, structure quality, and citation accuracy as separate axes. The main open challenge is to develop automated metrics that correlate with expert judgment on synthesis quality while remaining robust across domains, venues, and writing styles.

\subsubsection{Findings and Observations}
\label{sec:s2_findings}

\stageanalysis{~Stage 2: Literature Review}{S2color}{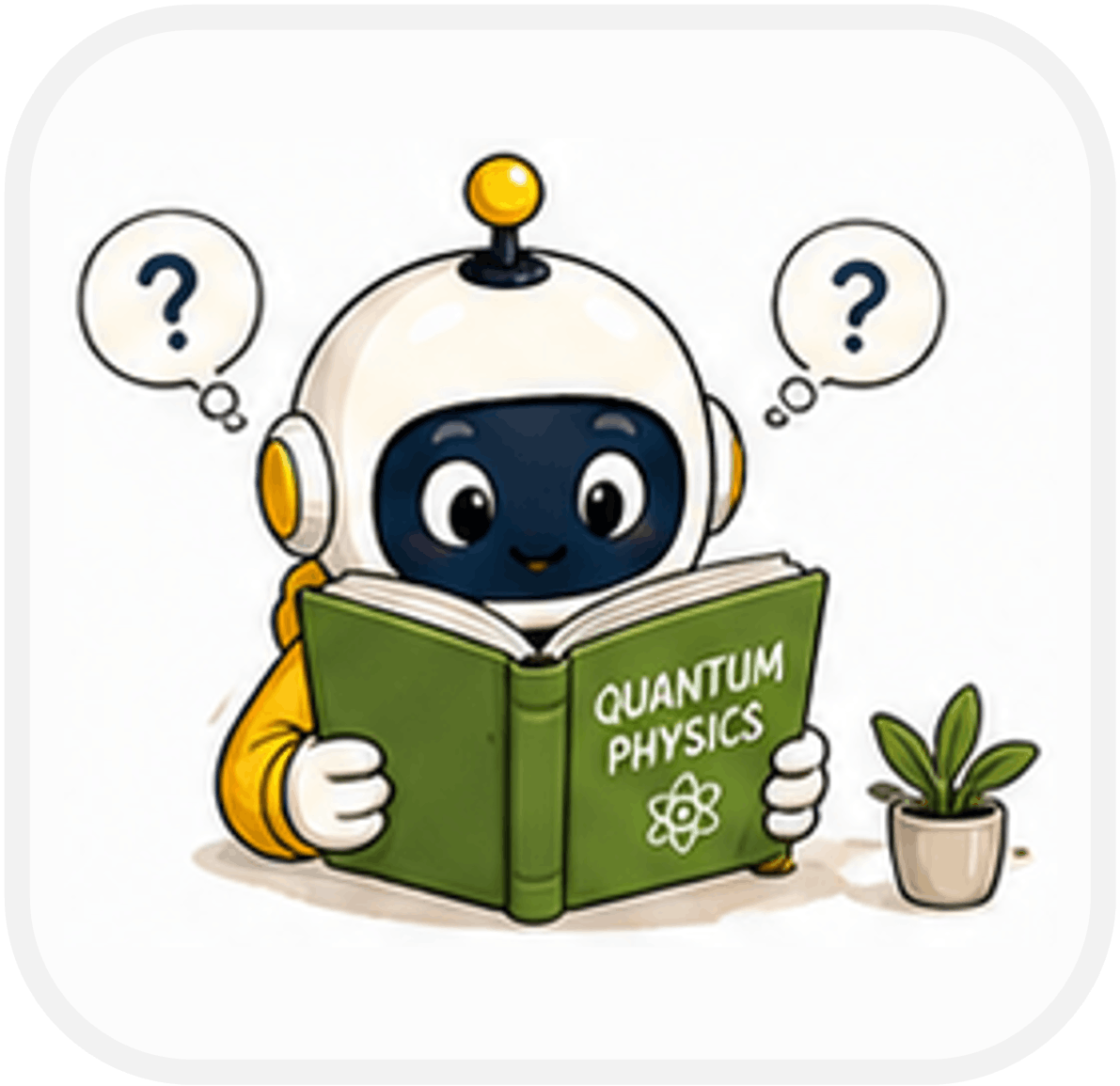}{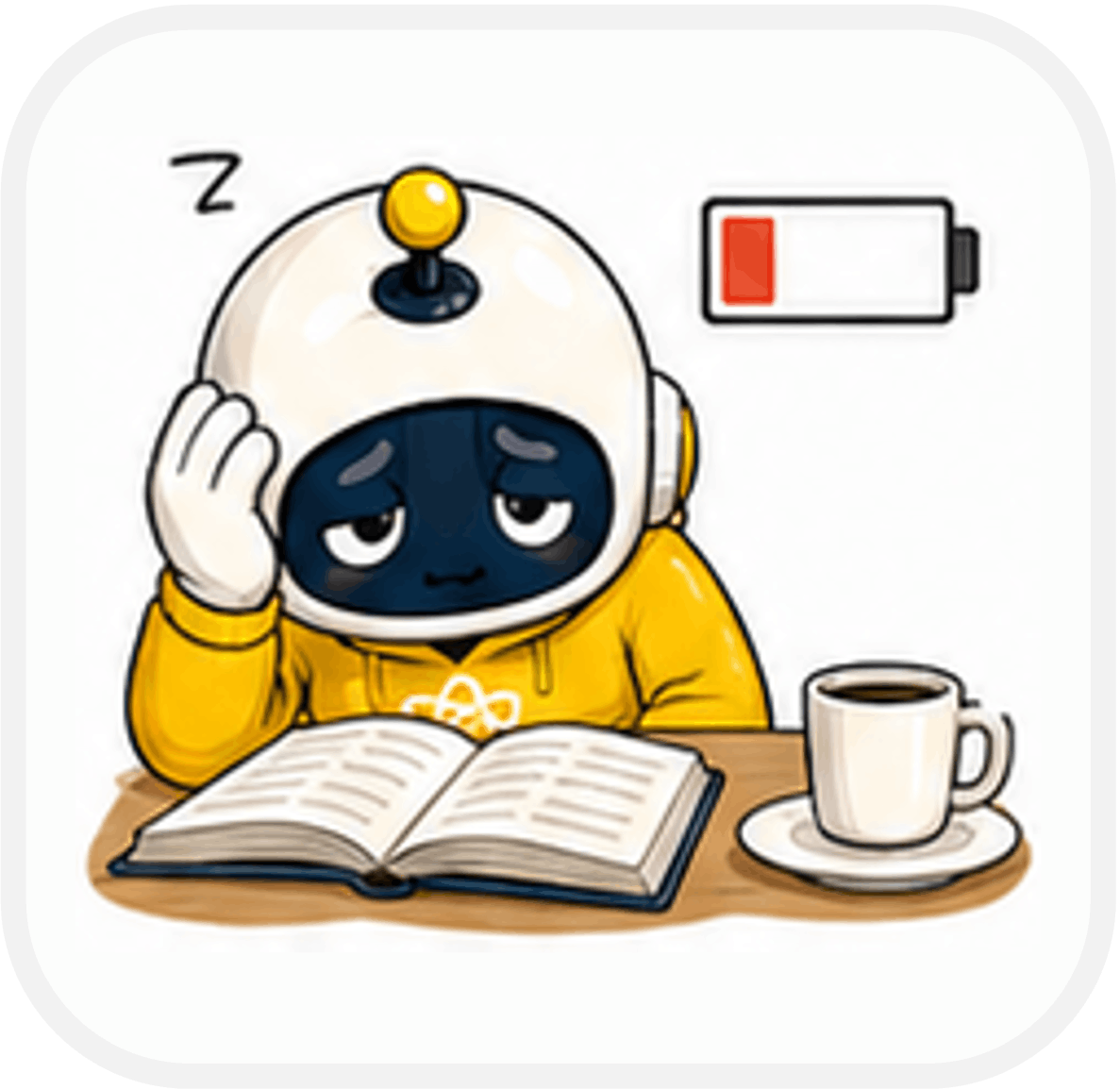}{%
\posbadge[S2color]{Fastest} \posbadge[S2color]{Convergence} \posbadge[S2color]{Open-Source}\par\vspace{2pt}
\posbadge[S2color]{Retrieval} \posbadge[S2color]{Synthesis} \posbadge[S2color]{DeepResearch}
}{%
\begin{itemize}[leftmargin=8pt, itemsep=0pt, topsep=0pt, parsep=0pt]
\item Fastest-maturing stage: four generations in two years (single-pass $\to$ structure-aware $\to$ multi-agent $\to$ editor-aware), with $35$ systems spanning retrieval, synthesis, and deep research.
\end{itemize}
\anadotrule
\begin{itemize}[leftmargin=8pt, itemsep=0pt, topsep=0pt, parsep=0pt]
\item Commercial and open-source systems increasingly converge on an iterative architecture: plan $\to$ retrieve $\to$ read $\to$ update $\to$ synthesize.
\end{itemize}
\anadotrule
\begin{itemize}[leftmargin=8pt, itemsep=0pt, topsep=0pt, parsep=0pt]
\item Recent evidence suggests that trajectory data and long-horizon tool-use supervision can be as important as model scale for improving the performance of deep research systems \cite{li2026openresearcher}.
\end{itemize}
}{%
\negbadge{Relations} \negbadge{Hallucination} \negbadge{Citation}\par\vspace{2pt}
\negbadge{Cross-Domain} \negbadge{Coherence} \negbadge{Scalability}
}{%
\begin{itemize}[leftmargin=8pt, itemsep=0pt, topsep=0pt, parsep=0pt]
\item Multi-paper relational reasoning remains a core bottleneck: the citation top-$1$ accuracy metric remains largely limited~\cite{scholarcop2025}, and relation-aware retrieval often remains weak~\cite{scinetbench2026}.
\end{itemize}
\anadotrule
\begin{itemize}[leftmargin=8pt, itemsep=0pt, topsep=0pt, parsep=0pt]
\item Hallucination has shifted from obvious fabrication to subtle misgrounding: generated claims may appear well-cited while not being faithfully supported.
\end{itemize}
\anadotrule
\begin{itemize}[leftmargin=8pt, itemsep=0pt, topsep=0pt, parsep=0pt]
\item Nearly all benchmarks and systems target ML/NLP literature; cross-domain synthesis (chemistry, biology, physics) remains largely untested and likely requires domain-specific retrieval infrastructure.
\end{itemize}
}
\vspace{8pt}

\subsection{Coding and Experiments}
\label{sec:experiment}

This stage translates research ideas into executable implementations, runs experiments, and analyzes the resulting evidence. Compared with literature review, coding and experimentation require AI systems to interact with external environments: repositories, dependencies, datasets, compute resources, test suites, and evaluation scripts. Existing work spans general code generation, paper-to-code translation, experiment orchestration, and result analysis. Across these directions, the central challenge is not whether LLMs can write plausible code, but whether they can produce semantically correct research implementations, execute meaningful experiments, and interpret results reliably. 

A comprehensive inventory of coding and experiment systems is provided in \cref{tab:appendix_s3} (Appendix).

\subsubsection{Code Generation}
\label{sec:code_gen}

General-purpose code generation has become one of the most mature capabilities of current LLMs. On SWE-bench Verified~\cite{swebench2024_leaderboard}, which evaluates real-world GitHub issue resolution, frontier systems now exceed $76\%$. Agent frameworks have played a central role in this progress. SWE-agent~\cite{yang2024sweagent} established the \emph{agent--computer interface} paradigm, giving LLMs structured access to files, tests, and tool calls rather than relying on unstructured shell interaction. OpenHands~\cite{wang2024openhands} extends this direction into a general open platform for software engineering agents and has become a common backbone for coding-oriented workflows.

However, high performance on standard software benchmarks does not directly imply readiness for research coding. SWE-bench Verified has been questioned for potential contamination, and more challenging variants expose a sharper limitation: performance drops to $23\%$ on SWE-bench Pro~\cite{deng2025swebenchpro} and $25\%$ on SWE-EVO~\cite{thai2025sweevo}. These results suggest that standard benchmarks may overestimate robustness when tasks are familiar, well-scaffolded, or pattern-matchable. This distinction becomes more pronounced in research settings, where the target is not only to fix existing software but to implement underspecified algorithms, reproduce implicit design choices, and validate scientific claims.

\subsubsection{Paper-to-Code}
\label{sec:paper2code}

Paper-to-code translation is a research-specific form of code generation. It is harder than conventional software engineering because research papers often mix natural-language descriptions, equations, pseudocode, ablation details, and domain conventions, while leaving key implementation choices implicit. PaperCoder~\cite{papercoder2025} addresses this setting with a three-stage multi-agent framework for planning, analysis, and code generation, transforming ML papers into executable repositories.

Dedicated benchmarks quantify how difficult this setting remains. ResearchCodeBench~\cite{researchcodebench2025} evaluates LLMs on $212$ novel ML implementation tasks, where the best reported model achieves only $37.3\%$ accuracy; notably, $58.6\%$ of errors are semantic, meaning that the generated code runs but implements the wrong algorithm or behavior. SciReplicate-Bench~\cite{scireplicatebench2025} reports a similar ceiling of $39\%$ across $100$ tasks from $36$ NLP papers. SciCode~\cite{scicode2024} extends research-level coding evaluation to mathematics, physics, and chemistry, while PaperBench~\cite{paperbench2025} decomposes 20 ICML 2024 papers into individually gradable subtasks covering environment setup, experiment execution, and result reproduction. Together, these benchmarks reveal a substantial gap between general software issue resolution and faithful research implementation.

At the high end, FunSearch~\cite{funsearch2024} demonstrates that LLM-generated programs can contribute to genuine mathematical discovery when embedded inside an evolutionary search loop. This result is important, but it also clarifies the boundary of current capability: success comes not from raw one-shot code generation alone, but from coupling generation with aggressive search, evaluation, and selection. The resulting contrast between strong performance on familiar software benchmarks and much lower performance on novel research code defines the capability cliff of this stage.

\subsubsection{Experiment Execution and Orchestration}
\label{sec:exp_execution}

Once the code is available, the next challenge is to run experiments systematically and efficiently. Experiment orchestration systems provide infrastructure for planning runs, modifying code, launching jobs, monitoring results, and iterating over failures. MLAgentBench~\cite{mlagentbench2024} evaluates language agents on ML experimentation; MLR-Copilot~\cite{mlrcopilot2024} separates autonomous research into idea and experiment agents; DS-Agent~\cite{dsagent2024} targets end-to-end data-science workflows; EvoDS~\cite{evods2026} extends data-science agents with self-evolving skill learning and explicit context management; and AIDE~\cite{jiang2025aide} frames ML engineering as tree search in code space. Broader evaluation environments, including MLR-Bench~\cite{mlrbench2025}, MLE-Bench~\cite{chan2024mlebench}, MLGym~\cite{mlgym2025}, and CURIE~\cite{curie2025}, provide increasingly standardized testbeds for measuring autonomous experimentation.

Recent systems push this infrastructure toward higher-throughput and closed-loop research workflows. R\&D-Agent~\cite{chen2025rdagent} uses a Researcher-Developer dual-agent design for ML experimentation, while Karpathy's autoresearch~\cite{karpathy2025autoresearch} demonstrates high-throughput experiment iteration. Closed-loop systems such as CodeScientist~\cite{codescientist2025}, Dolphin~\cite{dolphin2025}, and NovelSeek~\cite{novelseek2025} attempt to connect hypothesis generation, implementation, execution, and verification. EvoScientist~\cite{evoscientist2026} further illustrates the ambition of this direction by reporting accepted papers generated through a self-evolving research pipeline. AutoScientists~\cite{gao2026autoscientists} organizes self-organizing agent teams for long-running experimentation, while EurekAgent~\cite{xin2026eurekagent} argues that engineering the agent's environment is central to autonomous scientific discovery. These systems show that experimental throughput and workflow automation are improving rapidly, but their reliability still depends heavily on task scaffolding, benchmark design, and verification quality.

A complementary line of work couples execution with search and learning signals. AlphaEvolve~\cite{alphaevolve2025} improves algorithms through LLM-generated mutations and automated evaluation. Si~\etal~\cite{si2026executiongrounded} use execution-grounded search with large-scale parallel GPU experiments, outperforming GRPO baselines. SciNav~\cite{scinav2026} uses pairwise tree-search judgments to select promising branches, while Yuksekgonul~\etal~\cite{yuksekgonul2026learntodiscover} combine test-time training and reinforcement learning for continuous improvement across mathematics, GPU kernel optimization, and computational biology. AutoReproduce~\cite{autoreproduce2025} addresses a different but related problem: reproducing cited experiments by extracting implicit knowledge from paper lineages.

Domain-specific systems illustrate how orchestration changes when the environment is scientific rather than purely computational. In chemistry, Coscientist~\cite{boiko2023coscientist} and ChemCrow~\cite{bran2024chemcrow} use LLM-driven tools to support autonomous research workflows. In biology, AlphaFold~3~\cite{abramson2024alphafold3} extends protein structure prediction to biomolecular complexes, while CRISPR-GPT~\cite{crisprgpt2024}, BioPlanner~\cite{bioplanner2024}, and LAB-Bench~\cite{labbench2024} target gene-editing design, protocol planning, and biology research evaluation. For systems-level optimization, KernelBench~\cite{kernelbench2025} and TritonBench~\cite{tritonbench2025} evaluate whether LLMs can generate efficient GPU kernels and Triton operators. Cross-domain suites such as AstaBench~\cite{astabench2025} and EXP-Bench~\cite{expbench2025} broaden evaluation to multi-domain scientific tasks and autonomous experiment execution.

Overall, the execution layer has advanced quickly, especially when tasks are well specified and feedback is automated. The harder problem is experiment \emph{planning}: deciding which experiments are worth running, in what order, and how to interpret failures. Many current systems perform well on prescribed task pools, but remain less reliable when asked to choose genuinely novel research directions. In this sense, coding and experimentation expose the same broader pattern as idea generation: execution capability is improving faster than the scientific judgment needed to decide what should be executed.

\subsubsection{Assessment: Code Correctness and Reproducibility}
\label{sec:exp_analysis}

Assessing coding and experiment systems requires more than checking whether the generated code runs. Research code must implement the intended algorithm, reproduce reported results, support meaningful ablations, and generate evidence that can be interpreted correctly. This makes semantic correctness and reproducibility central evaluation criteria.
Several benchmarks expose the difficulty of this interpretive layer. DiscoveryBench~\cite{majumder2024discoverybench} and ScienceAgentBench~\cite{scienceagentbench2024} evaluate scientific reasoning over experimental data, showing that LLMs still struggle with multi-step analysis over complex result sets. DiscoveryWorld~\cite{discoveryworld2024} provides a virtual environment with $120$ challenge tasks for automated scientific discovery agents. InfiAgent-DABench~\cite{infidabench2024} benchmarks end-to-end data-analysis workflows, including data cleaning, statistical testing, and visualization generation across diverse domains. At the whole-system level, MLReplicate~\cite{mlreplicate2026} benchmarks autonomous research systems on end-to-end machine-learning reproducibility, directly probing whether agents can regenerate the results reported in a paper.

The core bottleneck is moving from raw outputs to trustworthy claims. Current systems can often produce plots, summary statistics, and local interpretations, but they are less reliable at identifying statistically meaningful trends, diagnosing failure modes, designing decisive ablations, and synthesizing results into a coherent empirical argument. This limitation is particularly consequential because coding errors and experimental misinterpretations can propagate into later writing and review stages, where polished narratives may obscure weak or incorrect evidence.

\subsubsection{Findings and Observations}
\label{sec:s3_findings}

\stageanalysis{~Stage 3: Coding \& Experiments}{S3color}{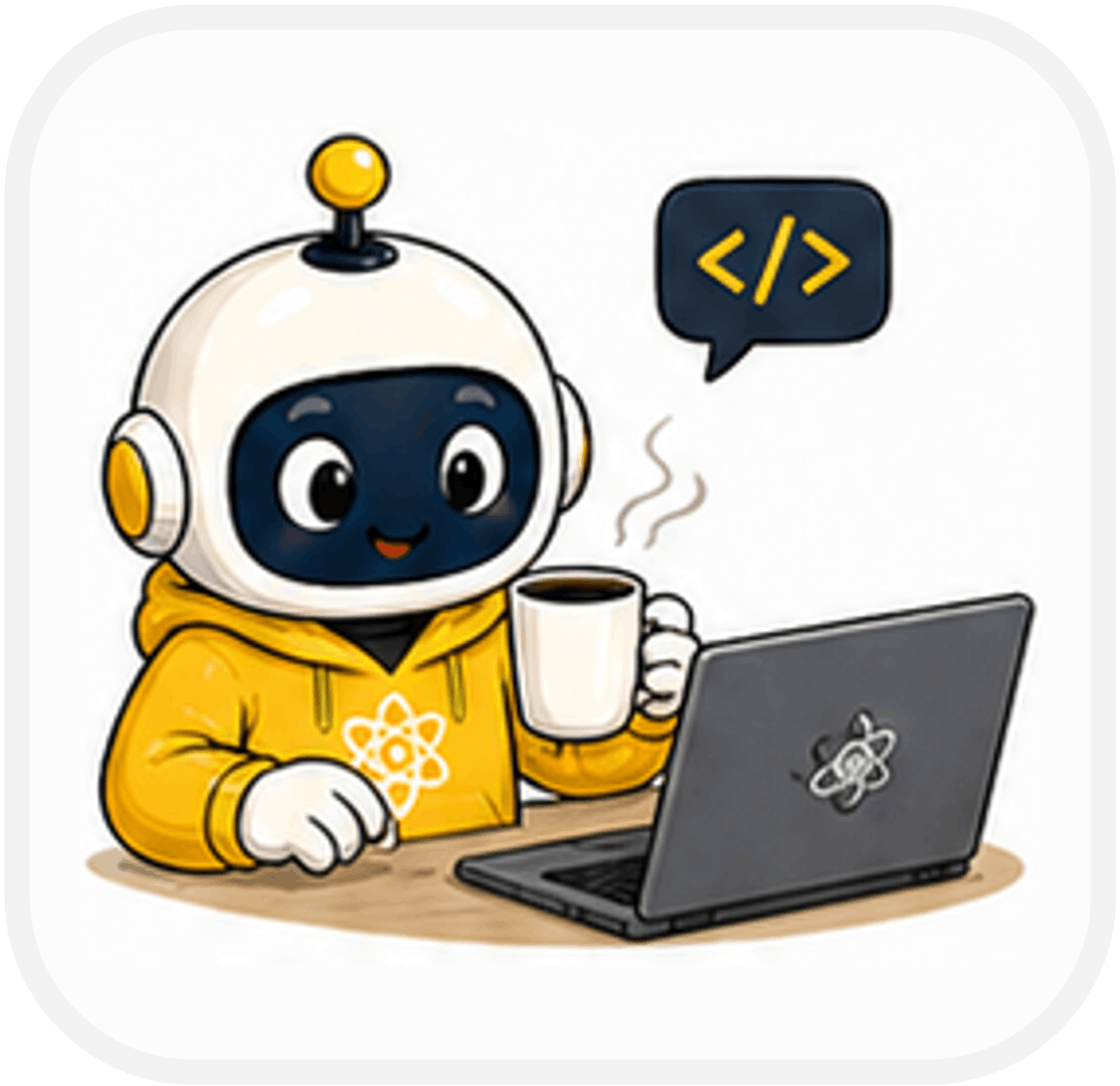}{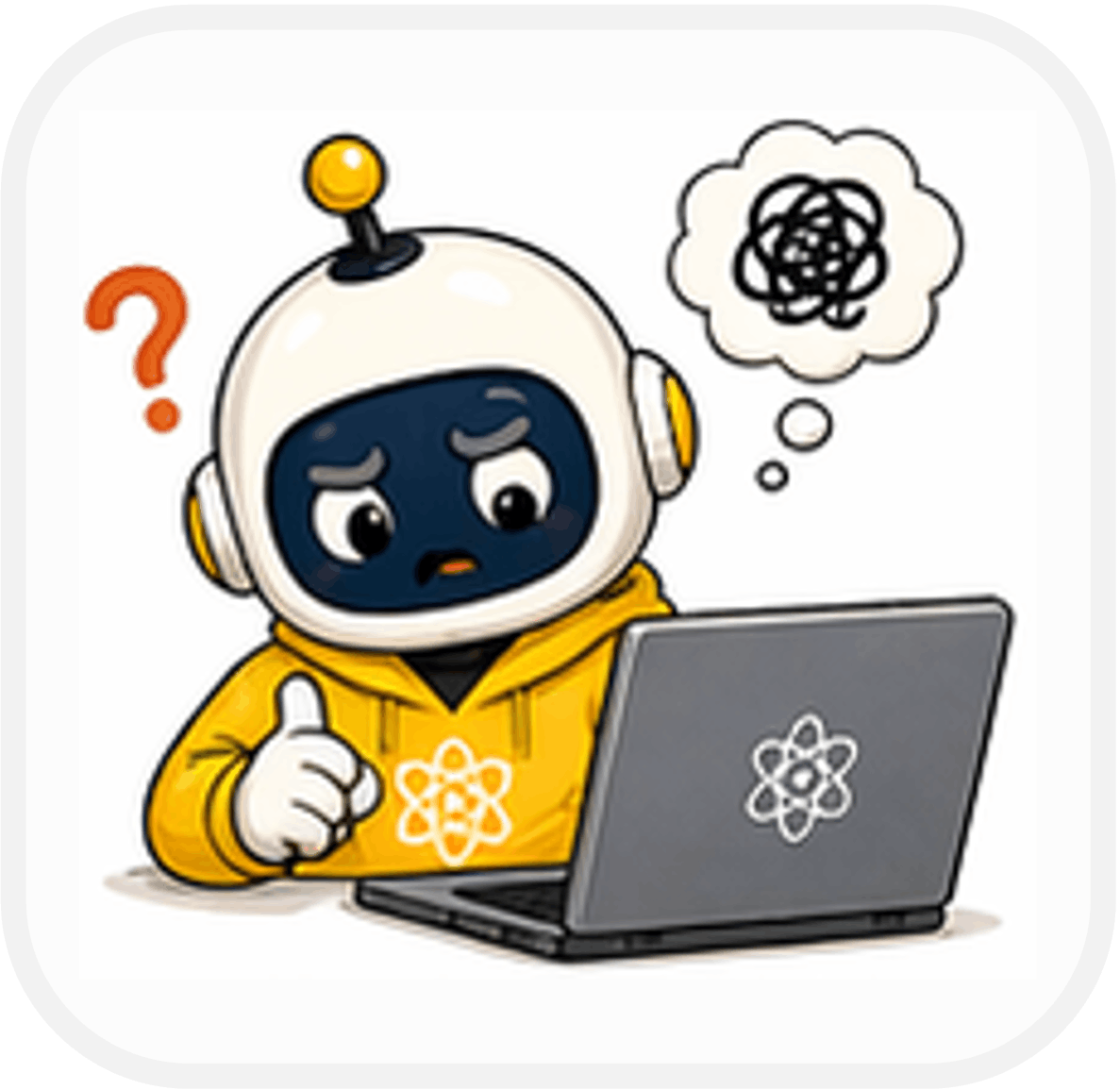}{%
\posbadge[S3color]{37\% Ceiling} \posbadge[S3color]{Closed-Loop} \posbadge[S3color]{Search}\par\vspace{2pt}
\posbadge[S3color]{Throughput} \posbadge[S3color]{Orchestration} \posbadge[S3color]{Cross-Domain}
}{%
\begin{itemize}[leftmargin=8pt, itemsep=0pt, topsep=0pt, parsep=0pt]
\item Sharpest capability boundary across all stages: $76\%$ on pattern-matching \emph{vs.} $37$--$39\%$ on novel research code, consistently reproduced across $4$+ independent benchmarks~\cite{researchcodebench2025,scireplicatebench2025}.
\end{itemize}
\anadotrule
\begin{itemize}[leftmargin=8pt, itemsep=0pt, topsep=0pt, parsep=0pt]
\item Execution infrastructure is no longer the bottleneck: systems sustain ${\sim}12$ experiments/hour in closed-loop, with generated papers accepted at academic venues~\cite{evoscientist2026,karpathy2025autoresearch}.
\end{itemize}
\anadotrule
\begin{itemize}[leftmargin=8pt, itemsep=0pt, topsep=0pt, parsep=0pt]
\item Coupling generation with search (evolutionary, tree, RL) consistently outperforms raw code generation~\cite{funsearch2024,alphaevolve2025,jiang2025aide}, suggesting that search strategy matters more than model capability alone.
\end{itemize}
}{%
\negbadge{Semantic} \negbadge{Planning} \negbadge{Fabrication}\par\vspace{2pt}
\negbadge{Insight Gap} \negbadge{Benchmark Leak} \negbadge{Verification}
}{%
\begin{itemize}[leftmargin=8pt, itemsep=0pt, topsep=0pt, parsep=0pt]
\item Semantic failures are especially problematic: generated code may execute successfully while implementing the wrong algorithm or producing misleading results~\cite{researchcodebench2025}.
\end{itemize}
\anadotrule
\begin{itemize}[leftmargin=8pt, itemsep=0pt, topsep=0pt, parsep=0pt]
\item Current systems execute prescribed tasks more reliably than they choose meaningful experiments; experiment planning remains strongly dependent on human scientific judgment.
\end{itemize}
\anadotrule
\begin{itemize}[leftmargin=8pt, itemsep=0pt, topsep=0pt, parsep=0pt]
\item $80\%$ of fully autonomous results are fabricated~\cite{mlrbench2025}, and downstream review catches only half of methodological issues~\cite{hiddenpitfalls2025}, creating a compounding verification deficit.
\end{itemize}
}
\vspace{8pt}

\subsection{Tables and Figures}
\label{sec:figure_table}

Tables and figures transform experimental outputs, statistical summaries, algorithms, and conceptual designs into publication-ready research artifacts. Existing systems cover scientific figure generation, data visualization, table construction, formula generation, and algorithmic illustration. Compared with coding and experimentation, this stage is less about producing new evidence than about representing evidence faithfully. Across these artifact types, the central challenge is the gap between \emph{visual plausibility} and \emph{scientific correctness}: AI-generated outputs may look professional while containing incorrect labels, misleading layouts, invalid numerical relationships, or domain-specific notation errors. 

A comprehensive inventory of figure and table generation systems is provided in \cref{tab:appendix_s4} (Appendix).

\subsubsection{Scientific Figure Generation}
\label{sec:figure_gen}

Scientific figure generation spans method diagrams, architecture illustrations, result plots, data visualizations, and pipeline figures. Standard result plots are comparatively tractable because they can often be grounded in structured data and executable plotting code. In contrast, method diagrams and framework figures are harder because they require faithful spatial organization, correct information flow, domain-specific symbols, and paper-specific visual conventions.

For \emph{method and architecture diagrams}, AutoFigure-Edit~\cite{autofigure2026} generates editable text-to-SVG scientific illustrations from long-form text, enabling users to revise generated figures rather than treating them as fixed images. Its companion system AutoFigure~\cite{autofigure2026iclr} introduces FigureBench for generating and refining publication-oriented scientific illustrations. PaperBanana~\cite{paperbanana2026} employs multiple specialized agents for retrieval, planning, styling, visualization, and critique, while StarVector~\cite{starVector2025} focuses on scalable vector graphics from textual descriptions. More recent systems broaden the input modalities and grounding signals: Crafter~\cite{crafter2026} provides a multi-agent harness that produces editable scientific figures from heterogeneous inputs, DiagramRAG~\cite{diagramrag2026} retrieves reference diagrams to ground figure generation, and GeoSVG-RL~\cite{geosvgrl2026} applies geometry-aware reinforcement learning to layout-constrained text-to-SVG diagram generation. Together, these systems show a shift from static image generation toward editable, structured, and critique-aware figure construction.

For \emph{result plots and data visualization}, MatPlotAgent~\cite{matplotagent2024} uses VLM-based visual feedback to improve data visualization quality, while PlotGen~\cite{plotgen2025} and PlotCraft~\cite{plotcraft2025} study chart generation across diverse plot types and task difficulties. CoDA~\cite{coda2025} explores multi-agent collaboration for visualization, and ChartGPT~\cite{yuan2023chartgpt} decomposes chart generation into sequential reasoning steps for handling abstract natural-language inputs. More recent systems broaden the scope of generation and evaluation: SciFig~\cite{scifig2026} introduces rubric-based evaluation for pipeline figures, VisCoder~\cite{viscoder2025} studies code-based visualization generation at scale, DiagramAgent~\cite{diagramagent2024} targets multiple diagram categories with specialized agents, and SciFlow-Bench~\cite{scifigbench2026} evaluates scientific framework figures through structure-first analysis. These efforts indicate that standard data plots are increasingly tractable, while complex framework figures remain difficult because they require structural consistency rather than only visual appeal.

For \emph{figure editing and optimization}, VIS-Shepherd~\cite{visshepherd2025} provides constructive feedback for LLM-based data visualization, emphasizing critique and revision rather than direct generation alone. \cite{aifigurepolicies2026} surveys publisher policies on AI-generated figures and proposes best-practice guidelines for responsible use. The SAIL framework~\cite{sail2026} separates domain logic from code syntax, allowing researchers to retain scientific oversight while delegating implementation details to AI. Across these systems, the emerging design principle is human-guided refinement: AI can accelerate layout, rendering, styling, and accessibility improvements, but researchers must verify whether the figure faithfully represents the underlying method or data.

\subsubsection{Table Understanding and Generation}
\label{sec:table_gen}

Table generation spans two complementary tasks: understanding existing tables and creating new ones. On the understanding side, Chain-of-Table~\cite{chainoftable2024} performs table reasoning through multi-step table transformations, reflecting the fact that many table tasks require sequential operations rather than single-pass extraction. On the generation side, ArxivDIGESTables~\cite{arxivdigestables2024} synthesizes scientific literature into structured comparison tables, ShowTable~\cite{showtable2025} introduces collaborative reflection and refinement for creative table visualization, and Table2LaTeX-RL~\cite{table2latexrl2025} converts table images into LaTeX code using reinforced multimodal language models. CSPO~\cite{cspo2026} similarly targets structured table-to-LaTeX generation, introducing component-specific optimization to alleviate reward ambiguity during training.

Compared with standard figure generation, table generation remains less mature because scientific tables must satisfy stricter semantic constraints. Comparison tables require consistent axes, fair grouping of methods, complete citation coverage, and correct numerical transcription. Ablation tables are even more demanding because they encode experimental design choices, not only final results. AbGen~\cite{abgen2025} evaluates LLMs on ablation study design using expert-annotated examples from NLP papers, revealing a significant gap between LLM-generated table plans and human expert judgments. This suggests that table generation is not merely a formatting problem; it requires understanding which comparisons are scientifically meaningful and how evidence should be organized.

\subsubsection{Mathematical Formulas and Algorithm Pseudocode}
\label{sec:formula_gen}

Mathematical formulas, TikZ diagrams, and algorithm pseudocode are compact representations of scientific reasoning, making them particularly sensitive to small errors. Unlike ordinary prose or standard charts, these artifacts require exact syntax and exact semantics simultaneously: a misplaced symbol, index, operator, arrow, or dependency can change the meaning of the method. As a result, formula and pseudocode generation remain less robust than natural-language polishing or standard visualization.

Recent systems address this challenge through specialized datasets, multimodal inputs, and iterative refinement. AutomaTikZ~\cite{belouadi2024automatikz} introduces DaTikZ, a large-scale TikZ dataset, and shows that fine-tuned models can outperform general-purpose LLMs on scientific vector graphics. DeTikZify~\cite{belouadi2024detikzify} extends this line with multimodal input and MCTS-based iterative refinement over a larger collection of TikZ graphics. TikZilla~\cite{tikzilla2026} further suggests that domain-specific training with supervised fine-tuning and reinforcement learning can make smaller open-source models competitive with larger general-purpose models on TikZ generation. TeXpert~\cite{texpert2025} highlights the remaining difficulty: accuracy drops sharply as LaTeX tasks become more complex, especially for tables with merged cells, nested environments, and nontrivial formatting constraints. These results reinforce the broader pattern of table and figure generation: specialized training and iterative refinement help, but human verification remains necessary when visual or symbolic artifacts carry scientific meaning.

\subsubsection{Assessment: Visual Fidelity and Scientific Accuracy}
\label{sec:s4_eval}

Evaluation for table and figure generation must assess both \emph{visual fidelity} and \emph{scientific accuracy}. Visual fidelity asks whether an artifact is readable, aesthetically coherent, and consistent with publication conventions. Scientific accuracy asks whether the artifact faithfully represents the underlying data, method, comparison, or mathematical relation. The distinction is crucial: an AI-generated figure may look professional while containing misaligned arrows, incorrect labels, invalid quantitative relationships, or domain-specific notation errors.

Recent benchmarks increasingly target this gap. SciFlow-Bench~\cite{scifigbench2026} uses inverse-parsing evaluation to detect structurally incorrect but visually plausible framework figures. FigureBench~\cite{autofigure2026iclr} evaluates scientific illustration generation and refinement. PlotCraft~\cite{plotcraft2025} studies chart generation across diverse chart types, while SciFig~\cite{scifig2026} provides a rubric-based evaluation for pipeline figures. TeXpert~\cite{texpert2025} evaluates LaTeX generation across difficulty levels, exposing steep performance degradation on hard cases. AbGen~\cite{abgen2025} extends evaluation to ablation study design, where the challenge is not only formatting a table but selecting scientifically meaningful comparisons. \emph{Can AI Draw Science?}~\cite{chen2026candraw} broadens figure-generation evaluation to text-to-image and multimodal models, probing whether they render scientifically faithful figures rather than merely plausible images.

Across artifact types, maturity remains uneven. Standard result plots are the most tractable because they can be generated from structured data and validated through executable plotting code. Method diagrams and framework figures remain harder because they require spatial organization and semantic consistency. Tables are difficult when they encode comparison logic or ablation design rather than simple formatting. Mathematical formulas, TikZ diagrams, and pseudocode exhibit steep accuracy cliffs because small syntactic errors can alter scientific meaning. Together, these benchmarks show that \Sfour evaluation is moving from appearance-based judgment toward structure-, semantics-, and task-aware assessment.

\subsubsection{Findings and Observations}
\label{sec:s4_findings}

\stageanalysis{~Stage 4: Tables \& Figures}{S4color}{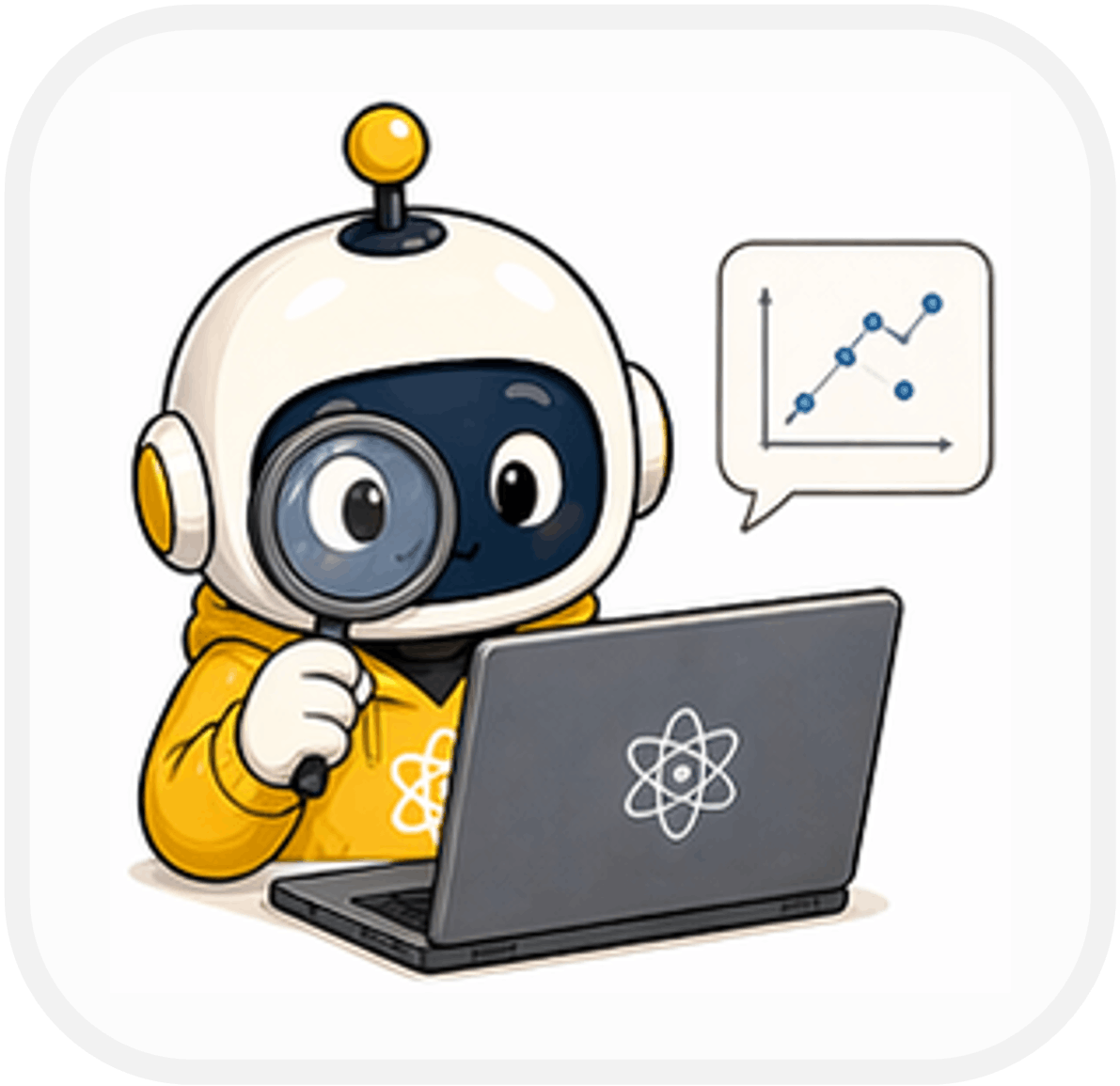}{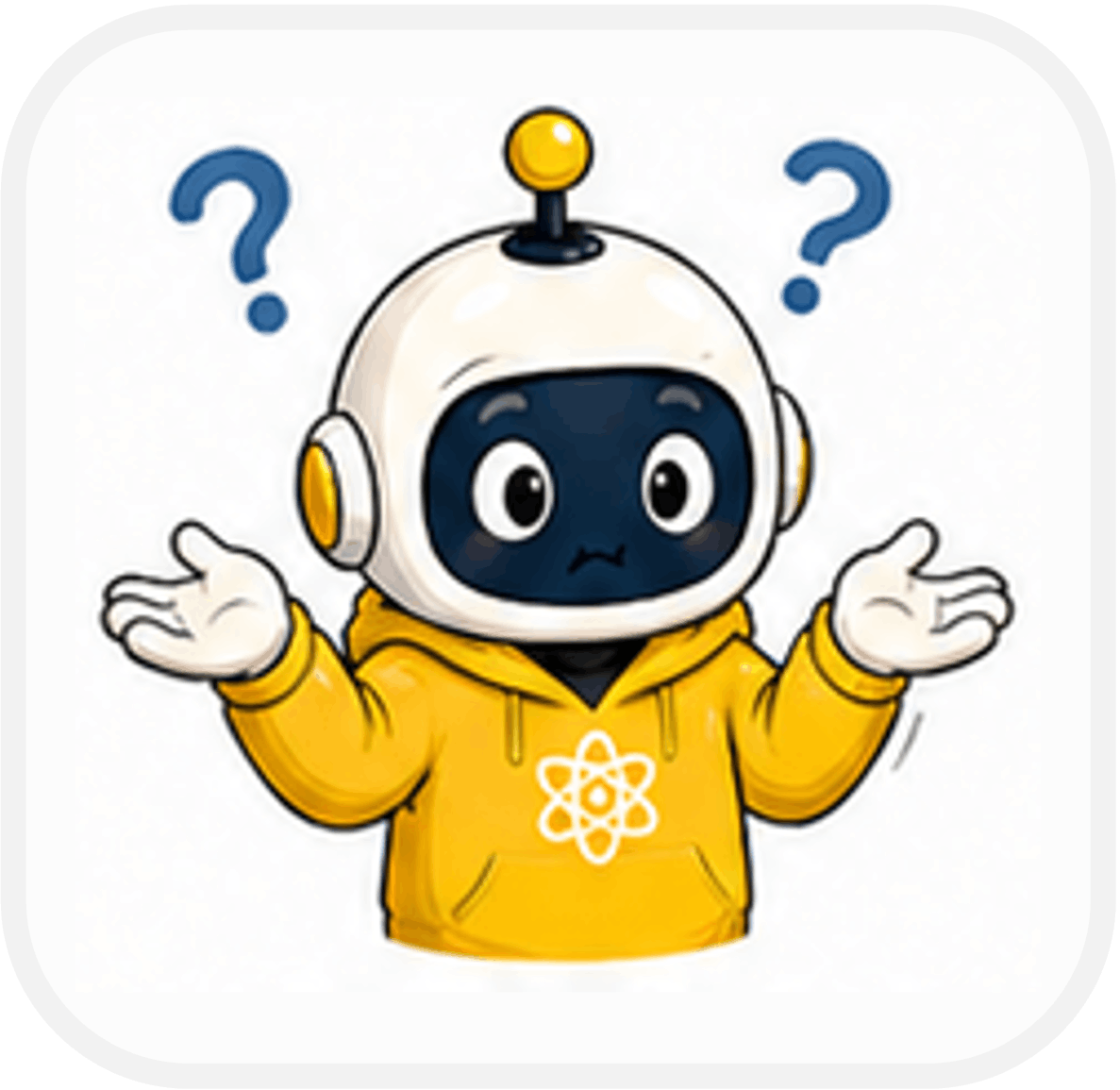}{%
\posbadge[S4color]{Emerging} \posbadge[S4color]{Visualization} \posbadge[S4color]{Small Models}\par\vspace{2pt}
\posbadge[S4color]{Multi-Agent} \posbadge[S4color]{Benchmarks} \posbadge[S4color]{Editing}
}{%
\begin{itemize}[leftmargin=8pt, itemsep=0pt, topsep=0pt, parsep=0pt]
\item Fastest-growing stage from zero: first dedicated tools appeared in late 2025, yet $20$+ systems already span figures, tables, formulas, and editing.
\end{itemize}
\anadotrule
\begin{itemize}[leftmargin=8pt, itemsep=0pt, topsep=0pt, parsep=0pt]
\item Standard data visualization becomes increasingly tractable: $90\%$+ execution pass rate on Matplotlib/Seaborn~\cite{viscoder2025}, with multi-agent approaches boosting quality $>40\%$ over baselines~\cite{coda2025}.
\end{itemize}
\anadotrule
\begin{itemize}[leftmargin=8pt, itemsep=0pt, topsep=0pt, parsep=0pt]
\item Domain-specific training and iterative refinement can make smaller specialized models competitive for structured visual languages such as TikZ~\cite{belouadi2024detikzify,tikzilla2026}.
\end{itemize}
}{%
\negbadge{Correctness} \negbadge{Uneven} \negbadge{Tables}\par\vspace{2pt}
\negbadge{Formulas} \negbadge{Spatial} \negbadge{Symbols}
}{%
\begin{itemize}[leftmargin=8pt, itemsep=0pt, topsep=0pt, parsep=0pt]
\item Visual plausibility $\neq$ scientific correctness: generated artifacts may look polished while misrepresenting data, structure, notation, or information flow.
\end{itemize}
\anadotrule
\begin{itemize}[leftmargin=8pt, itemsep=0pt, topsep=0pt, parsep=0pt]
\item Maturity is sharply uneven: figures are most advanced, table generation lags with no high-traction LaTeX tool, and formula accuracy drops from $78.8\%$ to $15\%$ with complexity~\cite{texpert2025}.
\end{itemize}
\anadotrule
\begin{itemize}[leftmargin=8pt, itemsep=0pt, topsep=0pt, parsep=0pt]
\item Tools remain assistants, not producers: AI-generated figures frequently require human modification for domain-specific symbols, spatial relationships, and paper-specific visual languages.
\end{itemize}
}
\vspace{8pt}

\subsection{Summary and Transition: Creation}
\label{sec:creation_summary}

The four Creation stages are tightly coupled in practice. \Sone (\emph{Idea Generation}) generates candidate hypotheses, \Stwo (\emph{Literature Review}) situates them within prior work, \Sthree (\emph{Coding and Experiments}) turns them into executable implementations and empirical evidence, and \Sfour (\emph{Tables and Figures}) converts the resulting outputs into visual and structured artifacts for communication. Progress across these stages shows a consistent pattern: AI systems are increasingly effective at producing the artifacts of research, including ideas, literature summaries, code, experiments, figures, and tables, but they remain less reliable at verifying whether these artifacts are novel, faithful, executable, and scientifically meaningful.

This gap appears differently at each stage. In \Sone, plausible novelty often weakens after implementation. In \Stwo, fluent synthesis can conceal citation errors or incomplete coverage. In \Sthree, executable code can still be semantically wrong, and automated runs do not guarantee meaningful experimental design. In \Sfour, polished visual artifacts may misrepresent data, notation, or methodological structure. These failure modes suggest that Creation-stage automation is most credible when coupled with grounding, execution feedback, explicit verification, and human scientific judgment.

The outputs of Phase~1 (\emph{Creation}) constitute the raw material for Phase~2 (\emph{Writing}). Ideas, retrieved literature, validated experiments, statistical summaries, comparison tables, and scientific figures must be organized into a coherent manuscript that explains the contribution, justifies its significance, and prepares it for external scrutiny. We therefore next turn to \emph{Writing}, where AI assistance shifts from producing research artifacts to structuring evidence into a scholarly argument.
\section{Phase 2: Writing}
\label{sec:writing_phase}

This phase consists of a single stage: \emph{Paper Writing} (\Sfive). Writing merits its own phase because it transforms the artifacts produced in Phase~1 into a scholarly argument. It is not merely a formatting step: a manuscript must select evidence, structure claims, situate contributions in the literature, explain methods with sufficient detail for reproducibility, and anticipate objections before external scrutiny in Phase~3 (\emph{Validation}). Compared with Phase~1 (\emph{Creation}), which emphasizes artifact production, Phase~2 (\emph{Writing}) emphasizes rhetorical organization and evidential justification.

This distinction matters for AI-assisted research. Writing tools are among the most widely adopted systems in the AI-for-research ecosystem, spanning grammar correction, sentence polishing, section drafting, citation support, and full-paper generation. At the same time, Writing is one of the most ethically sensitive phases because questions of authorship, attribution, disclosure, and the boundary between assistance and generation remain unresolved. The central challenge is therefore not whether AI can produce fluent academic prose, but whether it can preserve factual grounding, argumentative depth, citation fidelity, and human accountability.

\subsection{Paper Writing}
\label{sec:writing}

AI-assisted writing has moved from occasional support to mainstream research practice. Large-scale corpus analyses estimate detectable AI modification in up to $17.5\%$ of computer science abstracts~\cite{liang2024mapping} and $13.5\%$ of biomedical abstracts~\cite{kobak2024delve}, while self-reported adoption is higher: a 2025 Nature survey found that more than half of researchers report seeking AI writing help~\cite{aireviewsurvey2025}. These measurements are imperfect, but together they indicate a clear shift: AI writing assistance is now embedded in everyday scientific workflows. This makes the quality, transparency, and governance of AI-assisted writing increasingly important.

We organize this stage along a spectrum of increasing autonomy: \emph{semi-automated assistance} (\cref{sec:writing_semi}), in which AI handles local operations such as polishing, citation formatting, and drafting while the researcher retains control of the argument; \emph{fully automated generation} (\cref{sec:writing_auto}), in which systems attempt end-to-end manuscript production; and the \emph{assessment} of writing quality and AI detection (\cref{sec:writing_detection_eval}). What separates these regimes is not surface fluency, which current models already supply, but the degree to which they preserve the author's evidential and argumentative judgment. As assistance shades into generation, the unresolved boundary between acceptable support and unattributed authorship turns paper writing from a largely technical task into a question of disclosure and accountability, a theme we revisit when discussing governance in \cref{sec:cross_cutting}.

A comprehensive inventory of AI-assisted writing systems is provided in \cref{tab:appendix_s5} (Appendix).

\subsubsection{Semi-Automated Writing Assistance}
\label{sec:writing_semi}

Semi-automated writing assistance supports different parts of the manuscript workflow, from planning and drafting to polishing and revision. At the planning stage, systems help generate titles, outlines, section structures, and citation suggestions. ScholarCopilot~\cite{scholarcop2025}, for example, trains LLMs for academic writing with integrated citation recommendation, reflecting a broader trend toward tools that combine text generation with literature grounding.

During drafting, commercial tools such as Grammarly, Writefull, Paperpal, and GPT-based editors support paragraph generation, sentence polishing, citation insertion, and style refinement. Open-source prompt templates~\cite{leey21awesome} provide lightweight alternatives, while CoAuthor~\cite{coauthor2022} studies human--AI collaborative writing workflows. The dominant paradigm is increasingly shifting from \emph{``AI writes for you''} to \emph{``AI writes with you''}: AI handles mechanical or local operations, such as polishing, citation formatting, and initial drafting, while researchers retain responsibility for novelty, argumentation, experimental interpretation, and scientific judgment.

Editor-integrated systems make this collaboration more explicit. PaperDebugger~\cite{paperdebugger2025} embeds a multi-agent system into Overleaf, running Reviewer, Enhancer, Scoring, and Researcher agents within the writing environment. In a similar vein, PaperMentor~\cite{papermentor2026} acts as a human-centered multi-agent writing tutor that delivers Overleaf-native inline comments on research drafts while leaving authoring to the writer. A complementary line of work emphasizes cognitive engagement and transparency. Script\&Shift~\cite{siddiqui2025scriptshift} structures AI-assisted writing around \emph{source transformation} rather than direct text generation, aiming to preserve the writer's active reasoning. DraftMarks~\cite{siddiqui2025draftmarks} provides visual traces of revision intensity and AI-generated content, making the human--AI writing process more transparent to readers and reviewers. Empirical evidence~\cite{siddiqui2025aiwriting} further suggests that purposeful AI support can assist student writing without fully displacing cognitive effort.

Post-writing systems focus on revision, consistency, and style. XtraGPT~\cite{xtragpt2025} provides an open-source LLM suite for instruction-guided scientific paper revision, SciIG~\cite{sciig2025} benchmarks introduction writing using recent NAACL and ICLR papers, OpenDraft~\cite{opendraft2025} uses specialized agents to generate long research drafts with citation support, and LimAgents~\cite{limagents2026} integrates OpenReview comments and citation networks to generate research-limitation statements. Together, these systems show that semi-automated writing assistance is most credible when it augments researcher control rather than replaces the intellectual work of framing, interpreting, and defending a contribution.

\subsubsection{Fully Automated Paper Generation}
\label{sec:writing_auto}

Fully automated paper generation attempts to move beyond local assistance toward end-to-end manuscript production. Existing systems can be grouped into three directions. First, end-to-end research systems such as The AI Scientist~\cite{lu2024aiscientist,aiscientistnature2026} and Agent Laboratory~\cite{schmidgall2025agentlab} generate full papers as part of broader automated research pipelines. These systems demonstrate the feasibility of producing complete paper-like artifacts, but their outputs often remain limited by shallow argumentation, weak experimental validation, or insufficient novelty.

Second, benchmarked paper-generation systems aim to approach human review standards. CycleResearcher~\cite{cycleresearcher2024} reports generated papers scoring $5.36$ on the ICLR scale, approaching but still below the reported accepted-paper average of $5.69$. This gap is important because it suggests that the main bottleneck is no longer surface fluency alone. Rather, near-threshold papers often lack the argumentative depth, experimental rigor, and reviewer anticipation that distinguish publishable work from plausible drafts.

Third, rubric-guided and section-specific systems improve parts of the manuscript rather than generating the entire paper from scratch. APRES~\cite{apres2026} discovers rubrics predictive of citation counts and revises papers accordingly, with human experts preferring revised papers $79\%$ of the time. FutureGen~\cite{futuregen2025} targets ``Future Work'' section generation, and LECTOR~\cite{lector2026} targets introduction generation, jointly optimizing a scientific reasoning graph with the generated text to ground citations and curb hallucinated references. PaperWritingBench~\cite{paperwritingbench2026}, introduced as part of the PaperOrchestra framework, provides a dedicated benchmark for automated paper writing by evaluating multi-agent systems against reverse-engineered top-tier conference papers. RWGBench~\cite{xie2026rwgbench} narrows the lens to related-work generation, evaluating whether generated related-work sections correctly position a paper against prior scholarship. These systems indicate that automated writing is increasingly measurable, but also reinforce that high-quality papers require more than fluent text: they require evidence-grounded reasoning and coherent scientific contribution.


\subsubsection{Assessment: Writing Quality and AI Detection}
\label{sec:writing_detection_eval}

Assessment of AI-assisted writing involves two related but distinct questions: whether AI use can be detected, and whether the resulting manuscript is scientifically strong. Detection remains unreliable as a governance mechanism. Current detectors can produce unacceptable false positives, especially for formal, non-native, or highly edited academic prose, motivating a shift at major venues from attempting to \emph{detect} AI use toward requiring authors to \emph{declare} AI use. Watermarking offers a more principled route under controlled settings~\cite{watermarking2025}, but it requires model-provider cooperation and remains vulnerable to paraphrasing, translation, and post-editing. A related concern is the integrity of AI-generated citations rather than AI authorship per se: multi-agent detectors such as CiteTracer~\cite{citetracer2026} classify each reference as genuine, potential, or hallucinated, targeting fabricated citations directly.

Quality evaluation is more important, but also harder. Good academic writing must be assessed along multiple dimensions: factual correctness, citation accuracy, argumentative coherence, methodological completeness, novelty of framing, and stylistic appropriateness. LLM-as-Judge frameworks are increasingly used to approximate parts of this evaluation. CycleReviewer~\cite{cycleresearcher2024} reports a $26.89\%$ reduction in Proxy MAE relative to individual human reviewers for score prediction, while the Stanford Agentic Reviewer~\cite{stanfordreviewer2025} achieves review-score correlations comparable to human inter-rater agreement ($\rho=0.42$ vs.\ human $\rho=0.41$). These results suggest that automated evaluators can provide useful review-style signals, but they should not be treated as substitutes for expert assessment: score prediction and agreement metrics only partially capture factual grounding, evidential rigor, novelty, and scientific contribution. Complementing such outcome-based evaluation, process-oriented analyses~\cite{processeval2026} study how researchers actually revise AI-generated versus human-written drafts, finding that AI assistance helps most at the sentence level while struggling with global coherence.

The central failure mode of AI writing is therefore not ungrammatical prose, but unsupported persuasion: text that is fluent, well-structured, and citation-like, yet insufficiently grounded in evidence or scientific judgment. This issue is amplified by the productivity--quality divergence observed in recent studies: AI use can increase publication output, but AI-assisted papers with complex language may be less likely to be accepted~\cite{cornell2025science}. As in Phase~1 (\emph{Creation}), greater artifact production does not necessarily imply stronger research.

\subsubsection{Findings and Observations}
\label{sec:s5_findings}

\stageanalysis{~Stage 5: Paper Writing}{S5color}{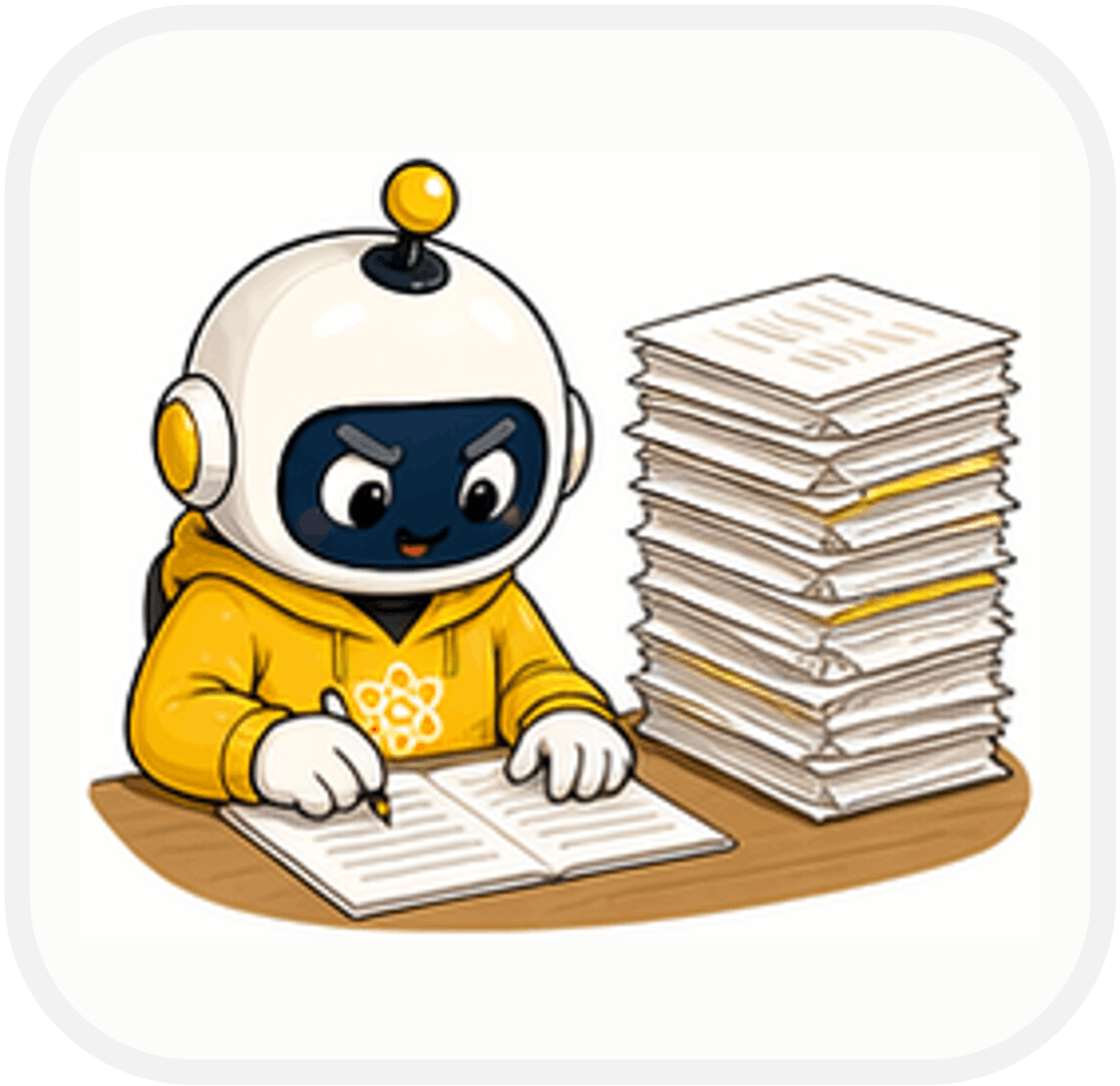}{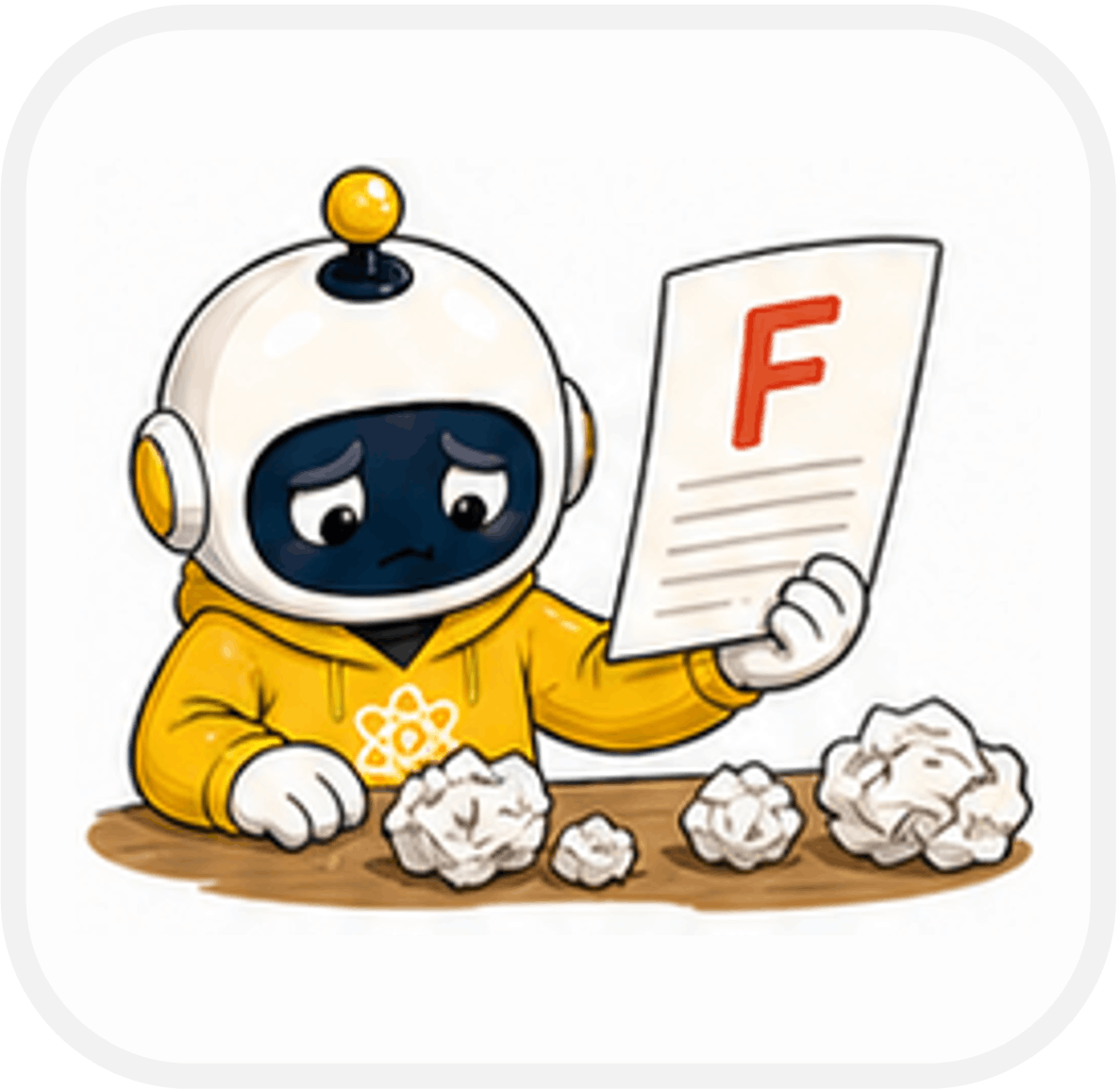}{%
\posbadge[S5color]{Commercial} \posbadge[S5color]{Near Threshold} \posbadge[S5color]{Cognitive}\par\vspace{2pt}
\posbadge[S5color]{Collaboration} \posbadge[S5color]{Rubric-Guided} \posbadge[S5color]{Transparency}
}{%
\begin{itemize}[leftmargin=8pt, itemsep=0pt, topsep=0pt, parsep=0pt]
\item Paper writing is among the most widely adopted AI-assisted research stages, with tools supporting planning, drafting, polishing, citation assistance, and manuscript revision~\cite{liang2024mapping,aireviewsurvey2025}.
\end{itemize}
\anadotrule
\begin{itemize}[leftmargin=8pt, itemsep=0pt, topsep=0pt, parsep=0pt]
\item Strong automated systems score $5.36$ on the ICLR scale (\emph{vs.} $5.69$ accepted~\cite{cycleresearcher2024}); rubric-guided revision achieves $79\%$ expert preference~\cite{apres2026}. The gap to acceptance is argumentative depth, not fluency.
\end{itemize}
\anadotrule
\begin{itemize}[leftmargin=8pt, itemsep=0pt, topsep=0pt, parsep=0pt]
\item Cognitive engagement and transparency are emerging design principles, aiming to preserve human understanding rather than merely producing polished text~\cite{siddiqui2025scriptshift,siddiqui2025draftmarks}.
\end{itemize}
}{%
\negbadge{Paradox} \negbadge{Mediocrity} \negbadge{Detect Failing}\par\vspace{2pt}
\negbadge{Shallow Args} \negbadge{Attribution} \negbadge{Deskilling}
}{%
\begin{itemize}[leftmargin=8pt, itemsep=0pt, topsep=0pt, parsep=0pt]
\item Productivity and quality can diverge: AI-assisted workflows may increase output, but more fluent or complex language does not necessarily translate into stronger acceptance outcomes~\cite{cornell2025science}.
\end{itemize}
\anadotrule
\begin{itemize}[leftmargin=8pt, itemsep=0pt, topsep=0pt, parsep=0pt]
\item Valley of mediocrity: papers are fluent enough to look real but lack argumentative depth, experimental rigor, and reviewer-anticipation, the skills that separate publishable from near-publishable work.
\end{itemize}
\anadotrule
\begin{itemize}[leftmargin=8pt, itemsep=0pt, topsep=0pt, parsep=0pt]
\item Detection tools have unacceptable false-positive rates, forcing a shift from ``detect'' to ``declare'' policies, while $17.5\%$ of CS papers already carry detectable AI modification~\cite{liang2024mapping}.
\end{itemize}
}
\vspace{8pt}

\subsection{Summary and Transition: Writing}

\label{sec:writing_summary}

This phase shifts the focus from producing research artifacts to organizing them into a scholarly argument. \Sfive (\emph{Paper Writing}) takes the outputs of Phase~1, including ideas, retrieved literature, experiments, figures, and tables, and converts them into a manuscript that explains what was done, why it matters, and how the evidence supports the claims. Progress in this phase shows that AI systems are increasingly effective at assisting the writing workflow, from planning and drafting to polishing, citation support, revision, and even full-paper generation.

The central limitation is that fluent writing can conceal weak reasoning. AI-generated or AI-assisted text may improve readability and productivity while leaving deeper scientific requirements unresolved: whether claims are grounded, whether citations faithfully support them, whether experiments are sufficient, and whether the contribution is argued with appropriate nuance. This limitation appears both in semi-automated writing assistance and in fully automated paper generation. The former is most credible when it preserves researcher control over framing, interpretation, and final responsibility; the latter increasingly approaches reviewable quality in selected settings, but still struggles with argumentative depth, evidential rigor, and reviewer anticipation.

The output of this phase is a manuscript ready for external scrutiny. We therefore next turn to Phase~3 (\emph{Validation}), where the manuscript is evaluated through peer review and revised through author rebuttal. This transition shifts the role of AI from structuring evidence into a coherent argument to assessing whether that argument is sound, fair, and sufficiently supported.
\section{Phase 3: Validation}
\label{sec:validation}

This phase encompasses the stages through which a manuscript produced in Phase~2 (\emph{Writing}) is externally scrutinized and iteratively refined: peer review and rebuttal with revision. Together, these stages address a different question from \emph{Creation} or \emph{Writing}: \emph{does this contribution meet the epistemic standards of the field?}

Validation is distinct because it introduces adversarial evaluation by reviewers who are expected to identify unsupported claims, methodological flaws, missing comparisons, unclear writing, and insufficient novelty. This makes Phase~3 a high-stakes setting for AI assistance. Automated systems can help summarize manuscripts, draft reviews, synthesize reviewer opinions, identify weaknesses, and support rebuttal preparation, but they also risk amplifying leniency, bias, adversarial manipulation, and weakly grounded criticism. The central challenge is therefore not whether AI can produce review-like text, but whether it can support fair, critical, and evidence-grounded evaluation without replacing independent expert judgment. The two stages also form a feedback loop: reviewer critiques in \Ssix (\emph{Peer Review}) may require additional experiments in \Sthree (Coding and Experiments), revised figures in \Sfour (\emph{Tables and Figures}), or manuscript rewrites in \Sfive (\emph{Paper Writing}), while rebuttal and revision in \Sseven (\emph{Rebuttal and Revision}) determine how those critiques are addressed.

\subsection{Peer Review}
\label{sec:peer_review}

Peer review is the gateway to validation. It evaluates whether a manuscript is technically sound, sufficiently novel, clearly presented, and supported by appropriate evidence. Existing systems span automated review generation, meta-review drafting, reviewer--paper matching, and review quality assessment. Across these directions, the central limitation is that LLMs can often produce structured and plausible critiques, but may under-detect methodological flaws, over-score weak submissions, and remain vulnerable to adversarial manipulation. 

A comprehensive inventory of automated review systems is provided in \cref{tab:appendix_s6} (Appendix).

\subsubsection{Automated Review Generation}
\label{sec:review_gen}

Automated review generation aims to produce structured critiques of manuscripts, including summaries, strengths, weaknesses, questions, and rating recommendations. Existing approaches can be grouped into four broad families. \emph{Fine-tuned reviewer models} specialize LLMs on expert review data to improve domain alignment and review format. DeepReviewer-$14$B~\cite{deepreviewer2025} reports strong performance against GPT-o1 and on ICLR 2024 accept/reject prediction, while OpenReviewer~\cite{openreviewer2025} fine-tunes Llama-$8$B on $79$K expert reviews. These systems show that supervised review data can improve review-style generation, but acceptance prediction remains a narrow proxy for review quality.

\emph{Multi-agent review systems} decompose reviewing into specialized roles. MARG~\cite{darcy2024marg} uses multi-agent collaboration to generate multiple substantive comments per manuscript, the open-source ai-peer-review tool~\cite{poldrack2024aireview} uses multiple LLMs to produce independent reviews followed by meta-review synthesis, and ScholarPeer~\cite{scholarpeer2026} extends this paradigm with literature search and claim verification. Such decomposition is useful because high-quality reviewing requires several distinct operations: understanding the manuscript, checking related work, evaluating claims, identifying weaknesses, and producing actionable feedback.

\emph{RL-optimized review systems} attempt to improve review quality through more explicit training signals. REMOR~\cite{remor2025} optimizes review generation with multi-objective reinforcement learning, while ReviewRL~\cite{reviewrl2025} combines retrieval-augmented context with RL to produce more comprehensive and grounded reviews. \emph{Prompt-based systems} provide a lighter-weight alternative. Reviewer2~\cite{reviewer2024} introduces a two-stage framework that models the distribution of review aspects, while ChatReviewer~\cite{chatreviewer2023} provides a deployed ChatGPT-based tool for analyzing strengths, weaknesses, and possible improvements. Moving beyond passive generation, ProReviewer~\cite{proreviewer2026} frames reviewing as a proactive investigation, with the agent actively probing suspicious parts of a manuscript before issuing critiques. PeerCheck~\cite{chen2026peercheck} instead targets review \emph{quality}, aiming to raise LLM-generated academic reviews toward human-level standards. Overall, automated review generation has become increasingly structured, but review-like prose should not be mistaken for reliable validation: the core difficulty is whether critiques are accurate, calibrated, and grounded in the manuscript and relevant literature.

\subsubsection{Meta-Review Generation}
\label{sec:metareview}

Meta-review generation synthesizes multiple reviewer opinions into a coherent area-chair-style assessment. This task differs from single-review generation because it must compare reviewer concerns, resolve disagreements, identify consensus, and justify a final recommendation. Bhatia~\etal~\cite{metareviewllm2024} evaluate GPT-3.5, LLaMA-2, and PaLM-2 on composing meta-review drafts from 40 ICLR papers, finding that LLMs are useful for multi-perspective summarization but struggle with nuanced judgment calls. AgentReview~\cite{agentreview2024} simulates the full review lifecycle, including meta-review and final decisions, and shows that social influence and authority bias can affect outcomes.

The main challenge is not summarization alone, but decision-making under disagreement. When reviewers fundamentally disagree about a manuscript's contribution, LLMs often produce diluted compromises rather than taking a defensible substantive position. This limitation reflects a broader issue in AI-assisted validation: current systems are better at consolidating stated opinions than at independently adjudicating technical disputes.

\subsubsection{Reviewer Matching}
\label{sec:reviewer_matching}

Reviewer--paper matching supports the editorial process by assigning manuscripts to reviewers with relevant expertise while accounting for conflicts of interest. This task is less visible than review generation, but it is crucial for review quality at scale: even a well-written review is less useful if the reviewer lacks appropriate domain expertise. RelevAI-Reviewer~\cite{relevaireviewer2024} has been deployed at major venues, while RATE~\cite{rate2026} improves expertise-based matching through profile distillation, aiming to capture a reviewer's competence signature beyond keyword overlap. MERIT~\cite{meritmatching2026} likewise targets reviewer assignment, using rubric-informed training to model expertise and retrieve suitable reviewers at scale.

Compared with automated reviewing, reviewer matching is a more appropriate setting for operational AI support because it assists the allocation process rather than replacing expert judgment. However, matching systems still require transparent conflict handling, robust expertise modeling, and human oversight, especially in interdisciplinary areas where surface-level keyword similarity can be misleading.

\subsubsection{Assessment: Review Consistency, Bias, and Robustness}

\label{sec:review_quality}

Assessing AI-assisted peer review requires more than measuring whether generated reviews resemble human reviews. A useful review must be consistent, critical, fair, grounded, and robust to manipulation. On consistency, recent systems show measurable progress. The Stanford Agentic Reviewer~\cite{stanfordreviewer2025} achieves Spearman correlation of $0.42$, comparable to human--human correlation of $0.41$. ReviewAgents~\cite{reviewagents2025} uses a multi-agent framework trained on a Review-CoT dataset of $37{,}403$ papers and $142{,}324$ reviews. The reviewer component reported in the Nature version of The AI Scientist~\cite{aiscientistnature2026} reaches $69\%$ balanced accuracy on ICLR acceptance prediction, while ClaimCheck~\cite{claimcheck2025} evaluates whether LLM critiques are grounded in the reviewed manuscript. ReViewGraph~\cite{reviewgraph2025} further models multi-round reviewer--author debates as heterogeneous graphs, improving debate outcome prediction without LLM fine-tuning. PRISM~\cite{prismreview2026} introduces a multi-dimensional benchmark for evaluating LLM peer reviewers, finding that models can excel on individual aspects yet none matches balanced human performance.

However, consistency is not sufficient. A reviewer can be consistent while being systematically lenient, biased, or shallow. LLM-based reviewers have been shown to assign inflated scores relative to humans, and in some settings to misclassify rejected papers as acceptable~\cite{llmreviewer2025}. A large-scale study in which $45$ expert scientists assessed the reviews of Nature-family papers~\cite{aireviewerslimits2026} finds that AI reviewers can surpass top human reviewers on some dimensions while still exhibiting recurring, systematic weaknesses; relatedly, the apparent tendency of LLMs to favor LLM-written papers has been shown to stem largely from general leniency toward low-quality work rather than a genuine self-preference effect~\cite{llmfavorllm2026}. This makes standalone AI review risky: it may produce coherent critiques while failing to identify decisive methodological weaknesses. A more credible deployment mode is to use LLMs to improve human reviews rather than to replace reviewers. In a randomized ICLR 2025 study across $22{,}467$ reviews, LLM feedback on reviews improved review quality in $89\%$ of cases, with reviewers updating their reviews $26.6\%$ of the time, without affecting acceptance rates~\cite{iclr2025reviewstudy}. Chen~\etal~\cite{reviewerfeedback2026} further study how reviewers engage with AI-generated feedback during a live ICLR 2025 process, and Zhuang~\etal~\cite{zhuang2025asprsurvey} provide a broader taxonomy of automated scholarly paper review methods.

Robustness and governance remain major concerns. The ``AI Review Lottery''~\cite{ailottery2024} estimates that at least $15.8\%$ of ICLR 2024 reviews were AI-assisted, with $49.4\%$ of submissions receiving at least one AI-assisted review. A 2025 Nature survey similarly reports that many academics use AI in peer review despite restrictive venue policies~\cite{aireviewsurvey2025}, and a major 2026 conference rejected $497$ papers for AI-use policy violations~\cite{aiuserejects2026}. The AAAI-26 AI Review Pilot~\cite{biswas2026aaai} reports one of the first large-scale controlled deployments of AI-assisted review at a top venue, offering early evidence on how such assistance behaves at scale. These trends indicate that AI involvement in peer review is already widespread, while enforceable governance remains immature.

Adversarial manipulation further complicates deployment. Breaking the Reviewer~\cite{breakingreviewer2025} studies adversarial robustness of LLM-based review assessments, while Keuper~\cite{promptinjectionreview2025} shows that simple prompt injections, such as white text on a white background, can manipulate LLM reviews. Ye~\etal~\cite{ye2024peerrisks} show that covert content injection can substantially raise review scores and that manipulating a small fraction of reviews can alter rankings. Zhou~\etal~\cite{zhou2025positiveprompt} further demonstrate that in-paper prompt injection can raise LLM scores under static and iterative attacks. Yang~\etal~\cite{presentationgaming2026} show that manipulation need not rely on hidden prompts at all: presentation-only revisions that merely rephrase and repackage a paper can substantially raise AI review scores. At the lexical level, Raina~\etal~\cite{raina2024adversarialjudge} show that benign adjectives can function as adversarial triggers, while Sahoo~\etal~\cite{sahoo2025indirect} evaluate indirect manipulation across multiple LLMs and attack strategies, with frontier models beginning to show stronger resistance in some settings. Finally, detection-based policy enforcement remains fragile: Saha~\etal~\cite{reviewpolicyenforce2026} show that state-of-the-art AI text detectors misclassify LLM-polished reviews, and Yu~\etal~\cite{aidetectionreview2025} evaluate $18$ detection algorithms on $788{,}984$ AI-written peer reviews, highlighting the difficulty of identifying AI-generated review text at the individual-review level. Beyond text detection, Li~\etal~\cite{li2026gaming} show that AI-assisted reviews can be actively \emph{gamed} with high attack success rates, and Russinovich~\etal~\cite{russinovich2026phantom} document hallucinated ``phantom'' citations that survive peer review at top-tier venues.


\subsubsection{Findings and Observations}
\label{sec:s6_findings}

\stageanalysis{~Stage 6: Peer Review}{S6color}{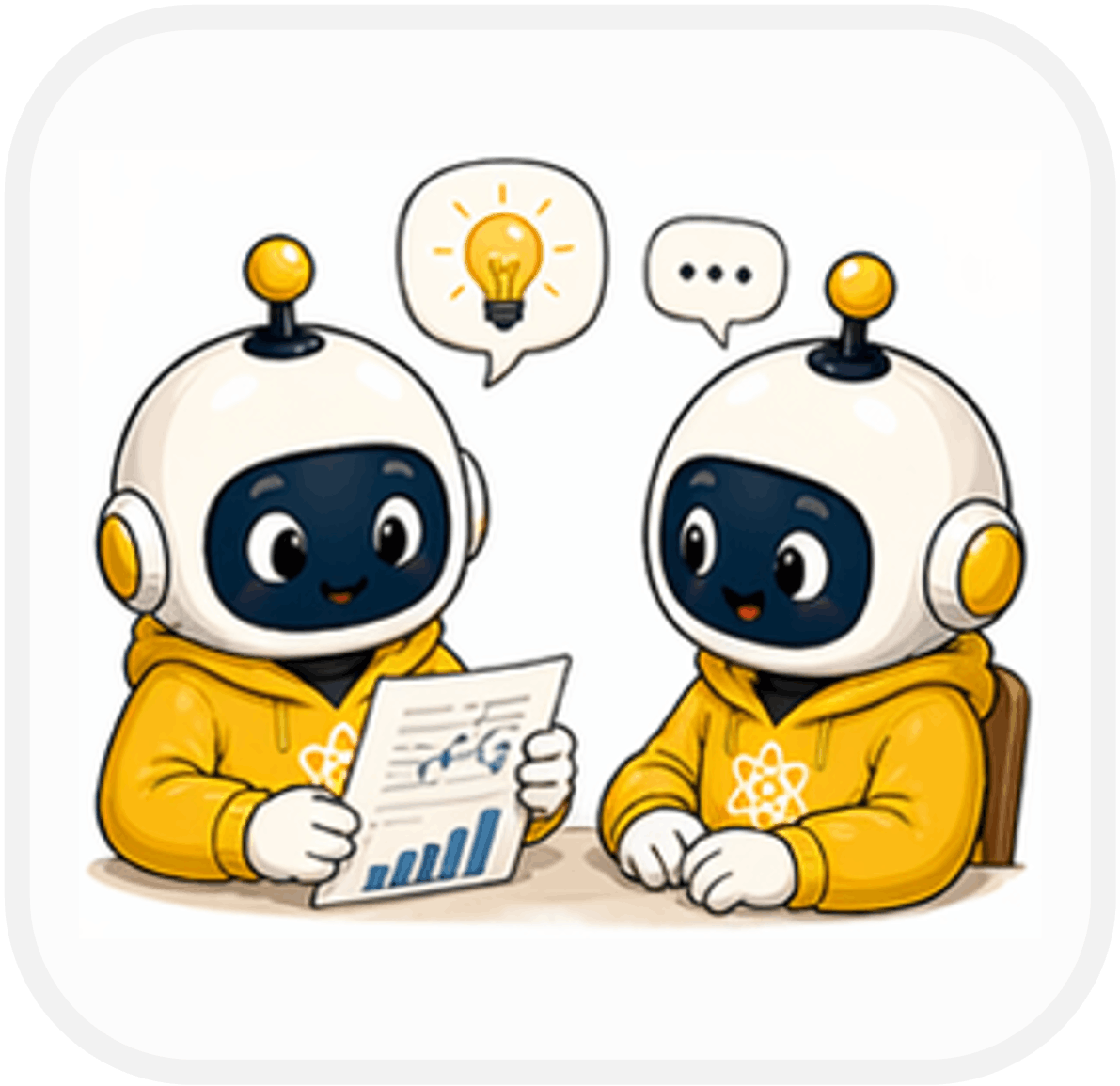}{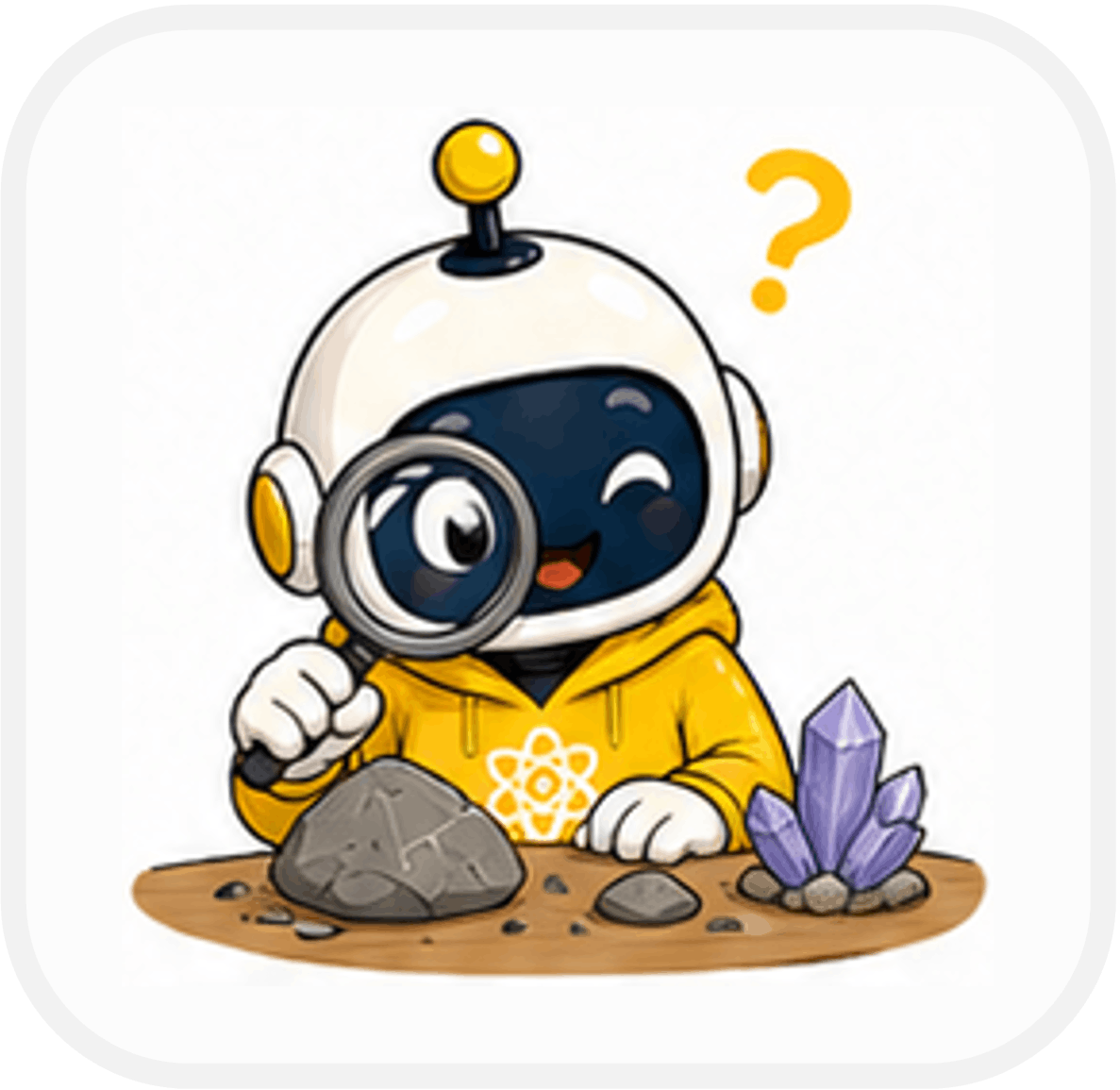}{%
\posbadge[S6color]{Deployment} \posbadge[S6color]{Consistency} \posbadge[S6color]{Mapped Risks}\par\vspace{2pt}
\posbadge[S6color]{Multi-Agent} \posbadge[S6color]{RL-Trained} \posbadge[S6color]{Matching}
}{%
\begin{itemize}[leftmargin=8pt, itemsep=0pt, topsep=0pt, parsep=0pt]
\item The strongest validated deployment mode is LLM feedback on \emph{reviews}, not standalone AI review: in ICLR 2025, review feedback improved quality in $89\%$ of cases without affecting acceptance rates~\cite{iclr2025reviewstudy}.
\end{itemize}
\anadotrule
\begin{itemize}[leftmargin=8pt, itemsep=0pt, topsep=0pt, parsep=0pt]
\item Automated reviewers can approach human-level consistency on selected metrics, with the Stanford Agentic Reviewer matching human inter-rater agreement ($\rho=0.42$ \emph{vs.} $0.41$)~\cite{stanfordreviewer2025}.
\end{itemize}
\anadotrule
\begin{itemize}[leftmargin=8pt, itemsep=0pt, topsep=0pt, parsep=0pt]
\item Reviewer matching and meta-review generation are promising support tasks because they assist editorial coordination and opinion synthesis rather than directly replacing expert judgment.
\end{itemize}
}{%
\negbadge{Leniency} \negbadge{Fragility} \negbadge{Policy}\par\vspace{2pt}
\negbadge{Inflation} \negbadge{Injection} \negbadge{Undetectable}
}{%
\begin{itemize}[leftmargin=8pt, itemsep=0pt, topsep=0pt, parsep=0pt]
\item Standalone AI-generated review remains unsafe: LLMs assign inflated scores (AI $6.86$ \emph{vs.} human $5.70$~\cite{llmreviewer2025}), misclassifying $95.8\%$ of rejected papers as acceptable.
\end{itemize}
\anadotrule
\begin{itemize}[leftmargin=8pt, itemsep=0pt, topsep=0pt, parsep=0pt]
\item Adversarial fragility persists: prompt injection reaches $10$ scores~\cite{zhou2025positiveprompt}; benign adjectives function as universal triggers~\cite{raina2024adversarialjudge}; $5\%$ manipulation flips $12\%$ of rankings~\cite{ye2024peerrisks}.
\end{itemize}
\anadotrule
\begin{itemize}[leftmargin=8pt, itemsep=0pt, topsep=0pt, parsep=0pt]
\item Governance is difficult because AI-assisted reviewing is already prevalent \cite{ailottery2024}, yet all five SOTA detectors fail on polished reviews~\cite{reviewpolicyenforce2026}. Prevalence has outpaced governance.
\end{itemize}
}
\vspace{8pt}

\subsection{Rebuttal and Revision}
\label{sec:rebuttal}

Rebuttal and revision form the second stage of Phase~3 (\emph{Validation}), where authors respond to external critique and revise the manuscript before a final decision or camera-ready submission. This stage is epistemologically important because it is the only point in the publication process where authors engage directly with reviewers' objections. Existing work spans reviewer-comment analysis, automated rebuttal generation, and revision tracking. Across these directions, the central challenge is not merely generating persuasive responses, but ensuring that rebuttals are evidence-grounded, faithful to the manuscript, and followed by actual revisions.

A comprehensive inventory of rebuttal systems is provided in \cref{tab:appendix_s7} (Appendix).

\subsubsection{Reviewer Comment Analysis}
\label{sec:rebuttal_analysis}

Reviewer-comment analysis decomposes critiques into actionable concerns, such as missing experiments, unclear motivation, insufficient baselines, unsupported claims, or presentation issues. This analysis is a prerequisite for effective rebuttal generation because reviewer comments are often long, mixed in priority, and partially overlapping across reviews. ReviewMT~\cite{reviewmt2024} models peer review as multi-turn, long-context dialogue with role-based interactions, covering $26{,}841$ papers and $92{,}017$ reviews from ICLR and Nature Communications. Re$^2$~\cite{re2dataset2025} further provides a consistency-ensured dataset for full-stage peer review and multi-turn rebuttal, covering $19{,}926$ submissions, $70{,}668$ reviews, and $53{,}818$ rebuttals from 24 conferences. A complementary line treats author responses as a supervision signal: RbtAct~\cite{rbtact2026} uses rebuttals to train models that generate actionable review feedback, while GoodPoint~\cite{goodpoint2026} learns to produce constructive paper feedback grounded in author responses.

Empirical studies show that rebuttal can materially affect outcomes, especially for borderline submissions. Analysis of ICLR 2024--2025~\cite{iclr_rebuttal2025} reports that $75$--$81\%$ of scores remain unchanged after rebuttal, $17$--$23\%$ improve, and only approximately $1\%$ decrease, with the most common transition being $5 \rightarrow 6$ from borderline to acceptable. A complementary analysis~\cite{louis2026rebuttals} finds that rebuttals do move review scores, but that the structure of the initial reviews bounds how much movement is possible; and as AI-generated reviews become common, Leblanc~\etal~\cite{leblanc2026trust} study how authors decide whether to trust and respond to AI-based reviews. These numbers suggest that rebuttal is not a universal remedy, but it is consequential for the subset of papers where reviewer concerns can be clarified, corrected, or supported with additional evidence.

\subsubsection{Automated Rebuttal Generation}
\label{sec:rebuttal_gen}

Automated rebuttal generation attempts to produce author responses that address reviewer concerns clearly and strategically. Early systems treated rebuttal as direct text generation, but this formulation is prone to hallucination, missed reviewer points, and unverifiable claims. More recent systems, therefore, decompose rebuttal into intermediate steps such as concern extraction, evidence retrieval, response planning, and final generation.
RebuttalAgent~\cite{rebuttalagent2026} uses Theory-of-Mind modeling to craft strategically persuasive responses, reporting an average $18.3\%$ improvement over the base model. Paper2Rebuttal~\cite{paper2rebuttal2026} introduces evidence-centric planning by decomposing reviewer comments into atomic concerns and retrieving supporting literature, improving Coverage and Specificity by up to $+0.78$ and $+1.33$. ReviewerToo~\cite{reviewertoo2025} includes a rebuttal module within a broader modular framework and reports $81.8\%$ accept/reject accuracy.

A newer direction emphasizes planning and author control. DRPG~\cite{drpg2026} proposes a four-step Decompose--Retrieve--Plan--Generate pipeline, reporting $98\%$+ planning accuracy and stronger-than-average human rebuttal quality with an $8$B model. Author-in-the-Loop~\cite{ruan2026authorinloop} integrates author expertise and intent into response generation, aiming to ensure that rebuttals reflect the paper's actual contributions rather than generic LLM output. Defend~\cite{defendrebuttal2026} generates rebuttals with minimal author guidance, reporting improved factual correctness and refutation strength while keeping the author in the loop. These systems indicate that rebuttal automation is moving from fluent response generation toward evidence-grounded and author-aware revision support.

\subsubsection{Assessment: Rebuttal Effectiveness}
\label{sec:rebuttal_effectiveness}

Assessing rebuttal systems requires measuring both immediate effectiveness and downstream accountability. Immediate effectiveness concerns whether a rebuttal addresses reviewer concerns, clarifies misunderstandings, provides evidence, and improves reviewer confidence. ICLR 2024--2025 analysis shows that papers whose scores improve after rebuttal achieve acceptance rates of $55.7$--$57.6\%$, compared to $7.8$--$12.4\%$ for papers with unchanged scores~\cite{iclr_rebuttal2025}. This makes rebuttal especially important for borderline papers, where small changes in reviewer confidence can affect final outcomes.

However, effective rebuttal is not only persuasion. Many reviewer requests require new experiments, additional ablations, corrected figures, or manuscript restructuring, creating a feedback loop from \Sseven back to \Sthree, \Sfour, and \Sfive. Ruan and Gurevych~\cite{ruan2026authorinloop} provide large-scale aligned review--response--revision triplets, enabling the study of how rebuttals translate into actual manuscript changes. Current rebuttal systems can retrieve evidence and draft responses, but they generally cannot generate new experimental evidence in response to reviewer requests. This makes the rebuttal--experiment loop one of the most practically important gaps in current auto-research pipelines.

The second evaluation dimension is accountability. A recent audit~\cite{rebuttalcommitment2026} finds that ICLR 2025 authors make an average of $11.8$ commitments per paper during rebuttal, but approximately $25\%$ of these commitments are not fulfilled in the camera-ready version, with missing experiments among the most common unfulfilled promises. This gap exposes the same capability-versus-integrity tension observed throughout the survey: AI systems may generate plausible and persuasive responses, but the scientific validity of a rebuttal depends on whether its claims are supported and its promises are later implemented. Rebuttal systems should therefore be evaluated not only by response quality, but also by concern coverage, evidence grounding, revision traceability, and fulfillment of author commitments.


\subsubsection{Findings and Observations}
\label{sec:s7_findings}

\stageanalysis{~Stage 7: Rebuttal \& Revision}{S7color}{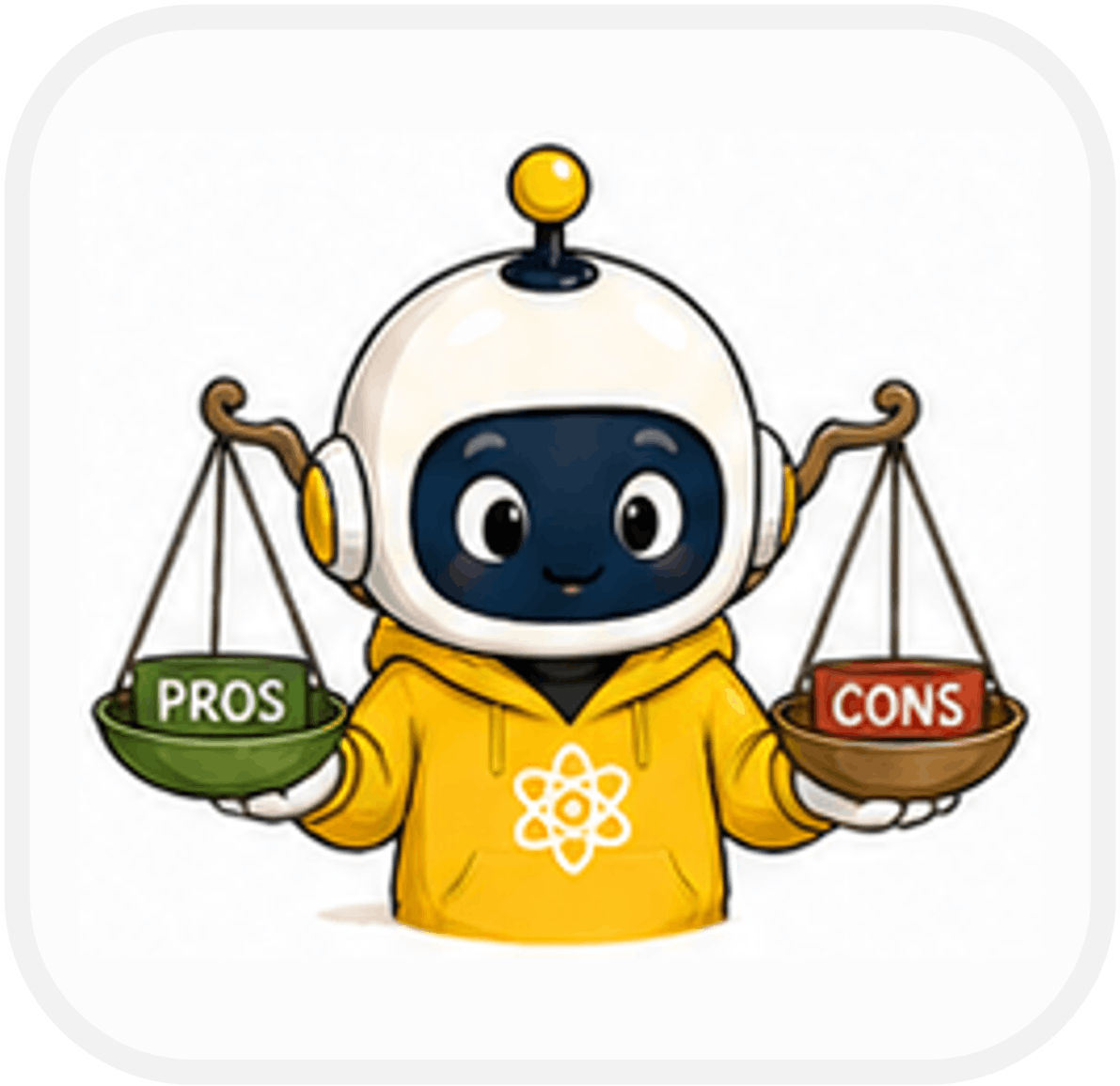}{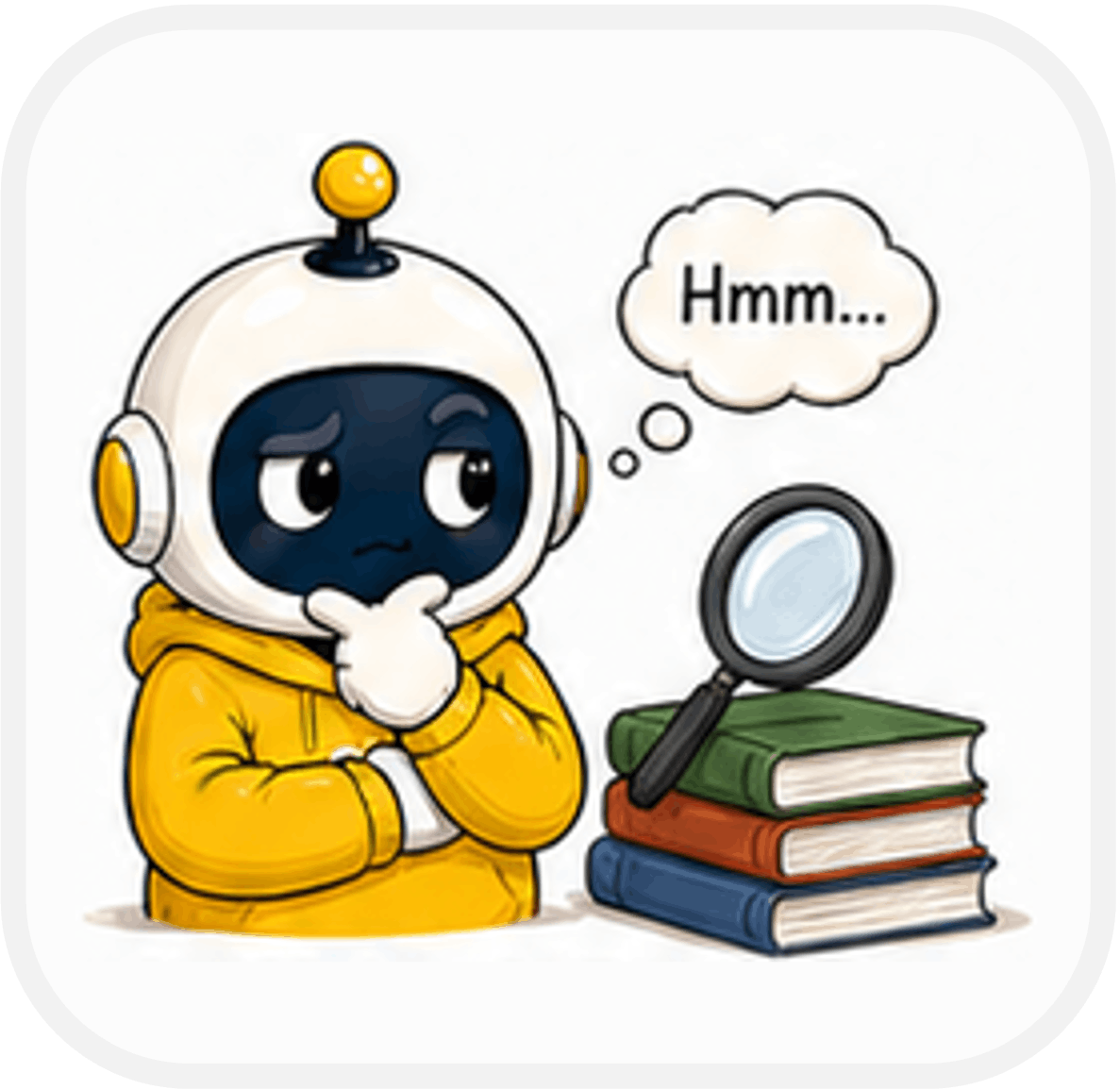}{%
\posbadge[S7color]{Newest} \posbadge[S7color]{Human-Level} \posbadge[S7color]{Decomposition}\par\vspace{2pt}
\posbadge[S7color]{Decisive} \posbadge[S7color]{Evidence} \posbadge[S7color]{Planning}
}{%
\begin{itemize}[leftmargin=8pt, itemsep=0pt, topsep=0pt, parsep=0pt]
\item Rebuttal automation is emerging as a distinct stage, with recent systems moving from direct response generation toward decomposition, evidence retrieval, response planning, and author-aware generation.
\end{itemize}
\anadotrule
\begin{itemize}[leftmargin=8pt, itemsep=0pt, topsep=0pt, parsep=0pt]
\item Rebuttal is consequential for borderline submissions: $17$--$23\%$ of ICLR 2024--2025 submissions improve scores after rebuttal, and improved-score papers achieve much higher acceptance rates~\cite{iclr_rebuttal2025}.
\end{itemize}
\anadotrule
\begin{itemize}[leftmargin=8pt, itemsep=0pt, topsep=0pt, parsep=0pt]
\item Evidence-centric planning helps address common failures of direct generation, including missed reviewer concerns, unsupported responses, and generic rebuttal text~\cite{rebuttalagent2026,paper2rebuttal2026}.
\end{itemize}
}{%
\negbadge{No New Expts} \negbadge{Commitment} \negbadge{Overlooked}\par\vspace{2pt}
\negbadge{Accountability} \negbadge{Only 10 Tools} \negbadge{Loop Gap}
}{%
\begin{itemize}[leftmargin=8pt, itemsep=0pt, topsep=0pt, parsep=0pt]
\item Current systems cannot reliably generate new experimental evidence in response to reviewer requests; the \Sseven (\emph{Rebuttal and Revision}) $\to $\Sthree (\emph{Coding}) feedback loop remains a major unautomated gap.
\end{itemize}
\anadotrule
\begin{itemize}[leftmargin=8pt, itemsep=0pt, topsep=0pt, parsep=0pt]
\item The quality of the rebuttal process must be tied to revision fulfillment: approximately $25\%$ of ICLR 2025 rebuttal commitments are not fulfilled in camera-ready versions~\cite{rebuttalcommitment2026}.
\end{itemize}
\anadotrule
\begin{itemize}[leftmargin=8pt, itemsep=0pt, topsep=0pt, parsep=0pt]
\item Rebuttal remains under-served relative to its practical importance, despite being the stage where authors directly negotiate reviewer concerns before final decisions.
\end{itemize}
}
\vspace{8pt}

\subsection{Summary and Transition: Validation}

\label{sec:validation_summary}

This phase shifts the focus from constructing a manuscript to testing whether its claims withstand external scrutiny. \Ssix (\emph{Peer Review}) evaluates the manuscript through independent critique, while \Sseven (\emph{Rebuttal and Revision}) gives authors an opportunity to clarify, defend, and revise the work in response. Together, these stages form a feedback loop rather than a one-way checkpoint: reviewer comments can trigger new experiments, revised analyses, updated figures, and substantial manuscript rewriting.

Progress in Validation shows a consistent pattern. AI systems are increasingly capable of producing review-like critiques, summarizing reviewer opinions, supporting reviewer matching, and drafting rebuttal responses. However, the hard part of validation is not producing plausible evaluative text; it is making fair, critical, evidence-grounded judgments and ensuring that critique leads to accountable revision. In peer review, standalone AI reviewers may be consistent but lenient, biased, or vulnerable to manipulation, while the strongest validated deployment mode is to use AI to improve human reviews. In rebuttal, AI can help decompose concerns and draft evidence-aware responses, but cannot yet generate missing experimental evidence or guarantee that author commitments are fulfilled.

The output of this phase is a manuscript that has been externally challenged, defended, and revised. Once validated, the research must be communicated beyond the review process through posters, slides, videos, project pages, and social media. We therefore next turn to Phase~4 (\emph{Dissemination}), where AI assistance shifts from evaluating scholarly claims to adapting validated research artifacts for different audiences, formats, and communication goals.
\section{Phase 4: Dissemination}
\label{sec:dissemination}

This phase converts the validated manuscript into formats accessible to audiences beyond specialist venue readers. The discussion covers the transformation of papers into posters, slides, videos, social media posts, and interactive agents. Compared with earlier phases, Dissemination is less about producing or validating scientific claims than about adapting those claims to different audiences, media, and interaction modes.

Dissemination merits a separate phase because its outputs are independent knowledge artifacts rather than simple derivatives of the paper. A poster must compress the contribution into a single visual narrative; a slide deck must support oral explanation; a video must synchronize visual, textual, and spoken channels; a social media post must balance accessibility with precision; and an interactive agent must expose the paper's methods for downstream use. The central challenge is therefore not whether AI can reformat a paper, but whether it can preserve scientific fidelity while adapting the work to new modalities, audiences, and levels of interactivity.

\subsection{Research Dissemination (Paper2X)}
\label{sec:paper2x}

Research dissemination converts a completed paper into audience-adaptive artifacts. Unlike Phases~1--3, which primarily target specialist authors, reviewers, and readers, Paper2X outputs must serve diverse audiences: conference attendees, oral-session audiences, online readers, prospective users, journalists, practitioners, and future researchers who may interact with the work through tools rather than text. Existing systems cover poster generation, slide generation, narrated video and talk generation, social media and web-page generation, and emerging paper-to-agent conversion. Across these formats, the core bottleneck is trust: researchers may use AI to draft public-facing materials, but they remain reluctant to delegate final communication to systems that may distort results, overstate claims, or omit important limitations.

A comprehensive inventory of Paper2X systems across poster, slides, video, web, social media, and agentic formats is provided in \cref{tab:appendix_s8} (Appendix).

\subsubsection{Paper to Posters}
\label{sec:paper2poster}

Paper-to-poster generation transforms a full manuscript into a compact visual narrative. This requires more than summarization: the system must select the central message, allocate space across motivation, method, results, and conclusion, preserve key figures and tables, and arrange them into a readable layout. Compared with slide generation, poster generation has stronger spatial constraints because all content must be legible and coherent on a single canvas.

Early systems established agentic poster generation as a feasible task. Paper2Poster~\cite{paper2poster2025} introduces PosterAgent with binary-tree layout planning and a Painter--Commenter feedback loop, showing that poster generation can be decomposed into layout construction, rendering, and critique. Subsequent systems add stronger design and hierarchy awareness. PosterGen~\cite{postergen2025} incorporates aesthetic-aware multi-agent generation, PosterForest~\cite{choi2025posterforest} uses hierarchical multi-agent collaboration, and P2P~\cite{p2p2025} introduces P2PInstruct with specialized agents and instruction data for poster design.

Recent systems move from one-shot poster creation toward editing and unified poster manipulation. APEX~\cite{apex2026} supports interactive poster editing with fine-grained control, addressing the practical need for human post-editing in conference preparation. PosterOmni~\cite{posteromni2026} unifies multiple poster tasks, including rescaling, filling, extension, layout-driven generation, style-driven generation, and identity-driven generation, while PosterCraft~\cite{postercraft2026} further explores quality-aware poster generation in a unified framework. Any2Poster~\cite{any2poster2026} broadens the input side, generating posters from heterogeneous source modalities across multiple domains. Together, these systems suggest that poster automation is shifting from direct paper summarization toward editable, design-aware poster production.

\subsubsection{Paper to Slides}
\label{sec:paper2slides}

Paper-to-slides systems convert manuscripts into sequential visual narratives for oral presentation. Unlike posters, slides unfold over time and must support speaker delivery. This requires content selection, section-to-slide mapping, visual layout, speaker-note synthesis, and often iterative refinement based on rendered slide quality. The key challenge is preserving the paper's argument while changing its rhetorical structure from written exposition to spoken explanation.

Early datasets and pipelines established the task. DOC2PPT~\cite{doc2ppt2022} provides paired document--slide data, while PPTAgent~\cite{pptagent2025} generates and evaluates presentations with PPTEval across content, design, and coherence. Environment-grounded refinement then closes the gap between symbolic planning and rendered slides. DeepPresenter~\cite{deeppresenter2026} conditions revision on rendered slide images rather than only internal reasoning traces, showing that visual feedback is important for presentation quality.

Multi-agent and interactive systems further decompose slide generation into specialized subtasks. SlideGen~\cite{slidegen2025} uses agents for outlining, content mapping, arrangement, note synthesis, and iterative refinement to produce editable PPTX slides. Auto-Slides~\cite{autoslides2025} targets Beamer generation with multi-agent collaboration and interactive editing. SlideTailor~\cite{slidetailor2025} conditions generation on user preference from a single example pair using a chain-of-speech mechanism. Other systems focus on task-specific capabilities: PASS~\cite{pass2025} combines slide generation with AI audio delivery, AutoPresent~\cite{autopresent2025} fine-tunes a slide-generation model on SlidesBench, Paper2Slides~\cite{paper2slides2025} provides one-click conversion through a multi-stage RAG pipeline, Talk to Your Slides~\cite{talkslides2025} supports natural-language slide editing, and Office Raccoon~\cite{sensetime2026} targets page-level editing with template and brand-guideline learning. X+Slides~\cite{xslides2026} adds an audience-conditioning dimension, tailoring slide content to whether the target audience comprises domain specialists or decision-makers. Across these systems, the main trend is from static slide generation toward editable, feedback-aware, and user-preference-conditioned presentation design.

\subsubsection{Paper to Videos and Talks}
\label{sec:paper2video}

Paper-to-video and paper-to-talk systems extend dissemination from visual artifacts to multimodal explanation. These systems must coordinate slides, subtitles, narration, cursor motion, pacing, and sometimes avatar or talking-head video. This makes the task substantially harder than poster or slide generation: errors can arise not only from content selection, but also from temporal alignment, speech clarity, visual synchronization, and duration constraints.
PresentAgent~\cite{presentagent2025} provides an end-to-end document-to-narrated-video pipeline with synchronized slides, text-to-speech narration, and the PresentEval benchmark. Paper2Video~\cite{paper2video2025} introduces a benchmark of paper--video pairs and the PaperTalker framework, decomposing video generation into slide, subtitle, cursor, and talker builders. Preacher~\cite{preacher2025} uses top-down decomposition followed by bottom-up generation with Progressive Chain of Thought across multiple research fields. PresentAgent-2~\cite{presentagentv2_2026} generalizes this line into an agentic framework for end-to-end presentation-video generation with research grounding and interactive delivery. Complementing generation, Mondal~\etal~\cite{mondal2026talks} evaluate what audiences actually \emph{learn} from paper-to-video talks, arguing that an effective talk teaches takeaways rather than merely summarizing the paper.

Although these systems show promising progress, video remains one of the hardest Paper2X formats. Unlike posters and slides, video generation requires coordination across at least four modalities: visual slides, subtitles, speech audio, and temporal or avatar-based presentation. The resulting artifact must also remain faithful to the paper while being concise enough for viewers to follow. Current systems, therefore, work best as first-draft generators that produce synchronized presentation assets for human review, rather than final public-facing videos requiring no editing.

\subsubsection{Paper to Social Media}
\label{sec:paper2social}

Paper-to-social-media and paper-to-web generation aims to make research discoverable outside the publication venue. Outputs include project pages, blog posts, press-release-style summaries, short-form research posts, and X/Twitter threads. These formats require stronger audience modeling than posters or slides: a thread for ML practitioners, a lay summary for journalists, and a project-page introduction for potential users should emphasize different details, use different vocabulary, and make different assumptions about background knowledge.
Paper2Web~\cite{paper2web2025} converts papers into interactive multimedia-rich academic homepages and provides a benchmark for this task. ResearchStudio-Reel~\cite{xiao2026researchstudio} targets this ``last mile'' more broadly, automating the joint production of poster, video, and blog artifacts from a single paper. More generally, researchers increasingly use general-purpose LLMs to draft online summaries, figure captions, project-page text, and social media announcements. However, dedicated research-to-social-media tools remain comparatively underdeveloped. The bottleneck is not text generation alone, but audience-adaptive fidelity: systems must simplify without distorting, emphasize contributions without exaggeration, and preserve limitations while remaining engaging.

This makes social dissemination a distinct trust problem. Public-facing outputs are often read without the paper, so any overclaim, missing caveat, or misleading comparison can shape how the work is perceived. AI assistance is therefore most credible when it provides audience-specific drafts, style variants, and claim-checking support, while leaving final messaging and factual responsibility to the authors.

\subsubsection{Paper to Agents and Tools}
\label{sec:paper2agent}

A newer dissemination direction converts papers from static documents into interactive agents or tools. This changes the function of dissemination: instead of only explaining a contribution, the system exposes the paper's methods, code, data, or workflows through natural-language interaction. In this setting, the reader becomes a user who can query, reproduce, adapt, or extend the work.

Paper2Agent~\cite{miao2025paper2agent} exemplifies this shift by converting research papers and associated codebases into interactive paper agents. The system analyzes the paper and code, constructs a Model Context Protocol (MCP) server with tools, resources, and prompts, and iteratively tests the resulting agent so that users can interact with the paper's methods through natural language. This reframes dissemination as operational access: a paper is no longer only read, but queried and executed.

Related systems broaden this idea from paper-specific agents to tool-using scientific agents. Gao~\etal~\cite{gao2025democratizing} study how scientific tool ecosystems can democratize AI scientists by exposing computational capabilities through agent-accessible interfaces. ProteinMCP~\cite{xu2026proteinmcp} applies an MCP-based agentic framework to protein engineering, illustrating how domain-specific workflows can be wrapped into interactive, tool-using systems. I-WebGenBench~\cite{dai2026iwebgen} complements these by evaluating interactivity in LLM-generated scientific web applications, probing whether generated interfaces are genuinely usable rather than static pages. These systems suggest that future dissemination may increasingly involve executable interfaces, reproducible workflows, and domain agents that make research easier to reuse.

This direction also introduces new risks. An interactive paper agent must not only summarize the paper faithfully, but also execute tools correctly, respect the limitations of the original method, and avoid presenting unsupported extrapolations as valid conclusions. Evaluation therefore requires both communication metrics and reproducibility metrics: whether the agent explains the paper clearly, whether it invokes the correct tools, whether it reproduces expected results, and whether it handles out-of-scope user queries responsibly.

\subsubsection{Assessment: Fidelity, Usability, and Adoption}

\label{sec:paper2x_eval}

Assessment for Paper2X must evaluate three dimensions: \emph{fidelity}, whether the generated artifact accurately represents the paper; \emph{usability}, whether the artifact supports its intended communication or interaction goal; and \emph{adoption}, whether researchers trust the artifact enough to use it publicly. Fidelity is the most important dimension because dissemination artifacts often circulate independently of the paper. A polished poster, slide, video, or thread can misrepresent a contribution if it omits caveats, changes baselines, simplifies methods incorrectly, or exaggerates results.

Poster and slide generation have the most mature evaluation infrastructure. Paper2Poster~\cite{paper2poster2025} evaluates poster quality under cost-efficient generation settings, while PPTAgent~\cite{pptagent2025} introduces PPTEval to assess slide quality along content, design, and coherence dimensions. Video evaluation is newer: PresentEval~\cite{presentagent2025} evaluates narrated video pipelines, and Paper2Video~\cite{paper2video2025} introduces comprehension-oriented evaluation through paper--video pairs. Complementing per-format benchmarks, recent work establishes fine-grained correspondences across a paper, its slides, and its presentation video~\cite{scicommcorr2026}, enabling cross-media fidelity to be measured directly. These benchmarks reflect a broader shift from surface aesthetics toward whether generated materials preserve content and improve audience understanding.

Agentic dissemination requires additional evaluation criteria. A paper agent should be assessed not only by answer quality, but also by tool correctness, reproducibility, error handling, and boundary awareness. This makes Paper2X evaluation closer to the assessment problems in Coding and Experiments (\Sthree): a system may appear helpful in natural language while invoking the wrong workflow or returning unsupported outputs. Across formats, no large-scale adoption study has yet established whether Paper2X tools reduce or increase misrepresentation compared with manual author-created materials. As a result, Paper2X systems are currently best understood as drafting and interaction aids rather than final producers.

\subsubsection{Findings and Observations}
\label{sec:s8_findings}

\stageanalysis{~Stage 8: Dissemination (Paper2X)}{S8color}{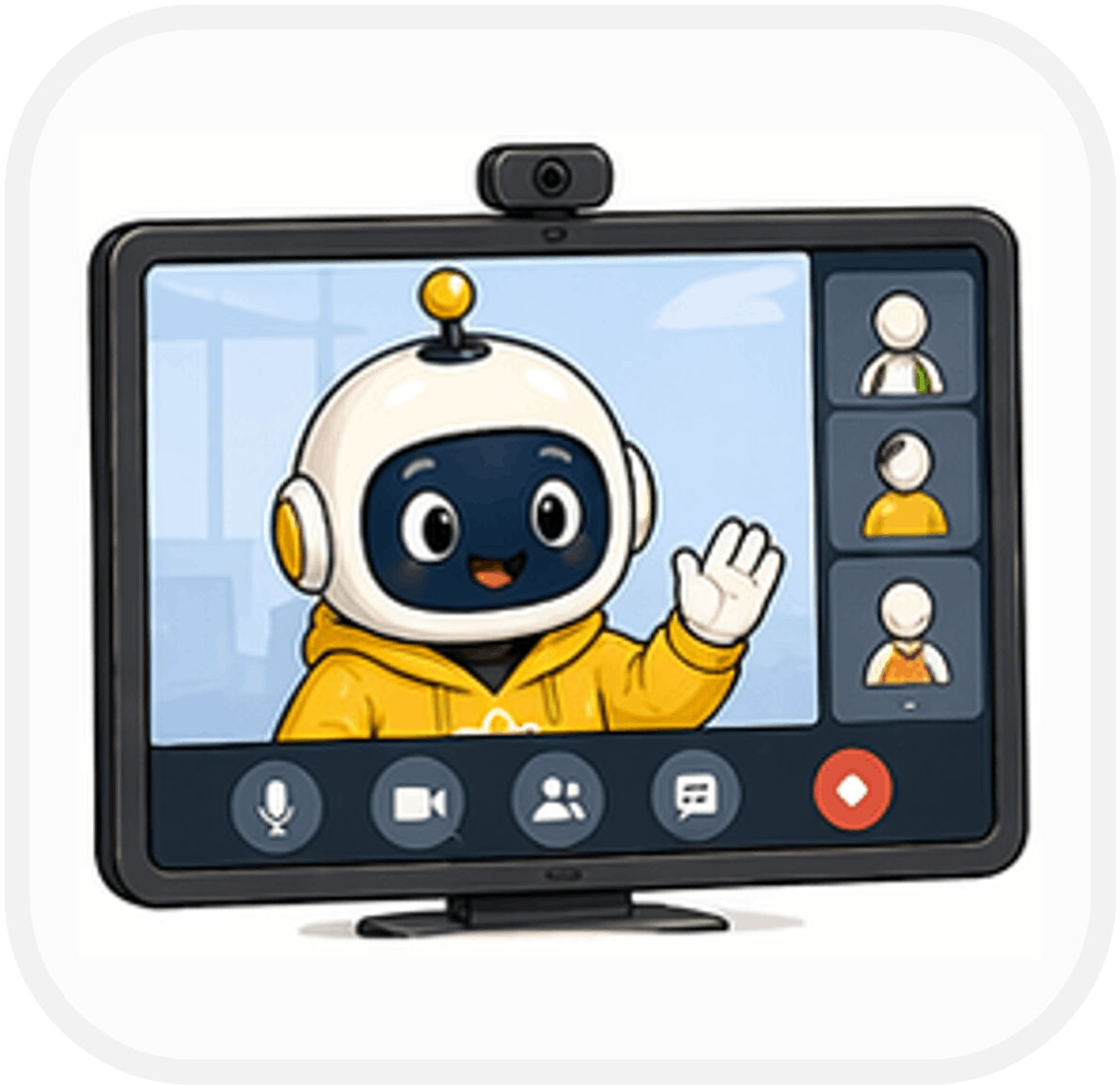}{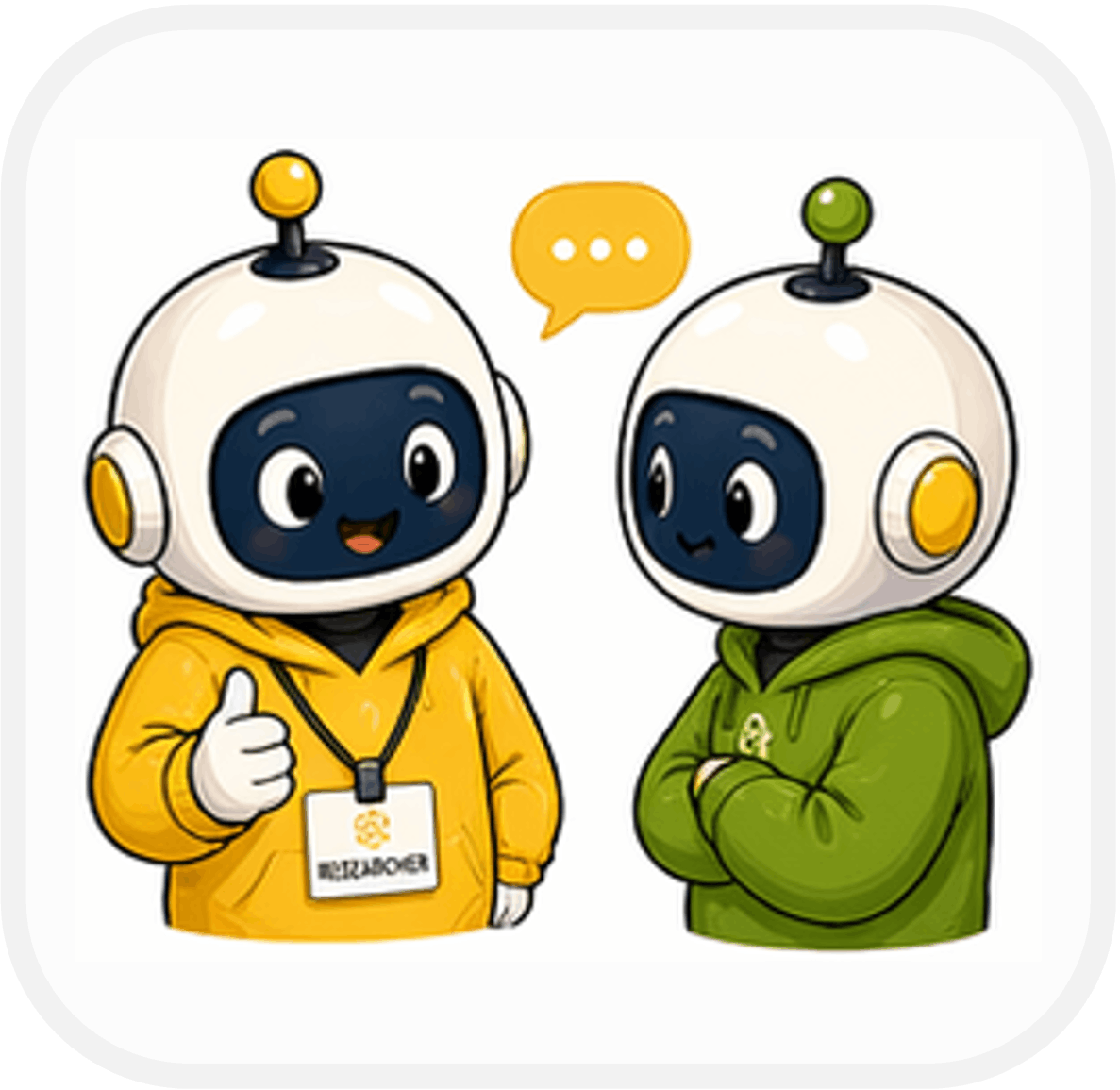}{%
\posbadge[S8color]{Low Cost} \posbadge[S8color]{Paper2Poster} \posbadge[S8color]{Paper2Slides}\par\vspace{2pt}
\posbadge[S8color]{Paper2Video} \posbadge[S8color]{Multi-Agent} \posbadge[S8color]{Interactive}
}{%
\begin{itemize}[leftmargin=8pt, itemsep=0pt, topsep=0pt, parsep=0pt]
\item Cost barrier eliminated: \$0.005/poster with $87\%$ fewer tokens~\cite{paper2poster2025}; $8$B models match frontier on slides~\cite{deeppresenter2026}. Most cost-efficient stage to automate.
\end{itemize}
\anadotrule
\begin{itemize}[leftmargin=8pt, itemsep=0pt, topsep=0pt, parsep=0pt]
\item Poster and slide generation are the most developed directions, moving from one-shot conversion toward editable, feedback-aware, and user-preference-conditioned workflows~\cite{paper2poster2025,pptagent2025,deeppresenter2026}.
\end{itemize}
\anadotrule
\begin{itemize}[leftmargin=8pt, itemsep=0pt, topsep=0pt, parsep=0pt]
\item Paper-to-agent systems extend dissemination from static explanation to interactive reuse, exposing paper methods, code, and workflows through tool-using agents~\cite{miao2025paper2agent,xu2026proteinmcp}.
\end{itemize}
}{%
\negbadge{Trust} \negbadge{4-Modal Hard} \negbadge{No Adaptation}\par\vspace{2pt}
\negbadge{Adoption} \negbadge{Fidelity} \negbadge{Social Media}
}{%
\begin{itemize}[leftmargin=8pt, itemsep=0pt, topsep=0pt, parsep=0pt]
\item Trust, not generation cost, is the bottleneck: researchers need confidence that AI-generated public artifacts preserve claims, caveats, and limitations.
\end{itemize}
\anadotrule
\begin{itemize}[leftmargin=8pt, itemsep=0pt, topsep=0pt, parsep=0pt]
\item Video remains difficult because it must coordinate slides, subtitles, narration, pacing, and sometimes avatar or cursor motion under strict time constraints~\cite{preacher2025,paper2video2025}.
\end{itemize}
\anadotrule
\begin{itemize}[leftmargin=8pt, itemsep=0pt, topsep=0pt, parsep=0pt]
\item Social media and web dissemination remain limited by audience modeling: simplifying a paper for broader reach without exaggerating or distorting the contribution remains unresolved.
\end{itemize}
}
\vspace{8pt}

\subsection{Summary and Transition: Dissemination}
\label{sec:dissemination_summary}

This phase shifts the focus from validated scholarly argument to audience-adaptive communication and reuse. The goal is to convert the manuscript into posters, slides, videos, project pages, social media posts, and increasingly interactive agents or tools. Progress in this phase shows that AI can substantially lower the cost of producing dissemination artifacts, especially when the input paper is complete and the target format has a predictable structure.

The central limitation is fidelity under format change. Each dissemination artifact compresses, reorders, or re-expresses the paper, creating opportunities for omission, overstatement, or distortion. This risk appears differently across formats: posters and slides may oversimplify the contribution, videos may misalign narration and visual evidence, social media posts may trade nuance for engagement, and paper agents may expose tools or workflows beyond their validated scope. Dissemination-stage automation is therefore most credible when it supports draft generation, format adaptation, editing, and interaction while preserving author oversight over claims and limitations.

The completion of this phase closes the research lifecycle: a contribution has been created, written, validated, and communicated. The remaining question is what patterns cut across all phases. We therefore next synthesize the common architectures, capability boundaries, deployment principles, and open challenges that define AI-assisted research as a whole.
\section{Cross-Cutting Analysis}
\label{sec:cross_cutting}

The preceding sections analyzed AI-assisted research stage by stage. We now synthesize patterns that emerge across the complete lifecycle. This cross-cutting view is necessary because many of the most important limitations do not appear within a single stage, but at the boundaries between stages: ideas that weaken after implementation, retrieved evidence that is misrepresented in writing, experiments that produce unsupported claims, reviews that miss methodological flaws, rebuttals that promise revisions without fulfilling them, and dissemination artifacts that simplify results beyond the evidence.

We organize this analysis around \textbf{four} questions. First, how do end-to-end systems integrate multiple stages of the research lifecycle? Second, how should research automation be evaluated across heterogeneous artifacts and long-horizon workflows? Third, what recurring capability boundaries and deployment principles appear across phases? Finally, what open challenges must be addressed before AI systems can be trusted as reliable research collaborators rather than artifact generators?

\subsection{End-to-End Research Systems}
\label{sec:e2e_systems}

Some systems discussed earlier, especially in \cref{sec:writing_phase}, also qualify as end-to-end research systems because they generate complete manuscripts. Here, however, we analyze them from a different perspective: not as paper-writing tools, but as lifecycle-scale architectures. The key question is how these systems connect ideation, literature review, coding, experimentation, writing, validation, and dissemination, and where the handoffs between stages remain fragile.

Most current end-to-end systems emphasize Phase~1 (\emph{Creation}) and Phase~2 (\emph{Writing}), connecting idea generation, implementation, experiment execution, and manuscript drafting into a single workflow. Far fewer systems incorporate substantive Phase~3 (\emph{Validation}), such as adversarial review, rebuttal planning, or revision tracking, and Phase~4 (\emph{Dissemination}) remains mostly outside current end-to-end pipelines. This imbalance reflects a broader pattern observed throughout the survey: it is easier to generate research artifacts than to validate, revise, and communicate them with accountable fidelity. 

Existing systems can be grouped into four architectural families: sequential pipelines, search-based and self-improving systems, skill-based and tool-integrated systems, and multi-agent or community-scale frameworks.

\subsubsection{Sequential and Pipeline-Based Systems}

Sequential systems connect research stages in a mostly linear order, typically moving from idea generation to experiment execution and manuscript drafting. The AI Scientist~\cite{lu2024aiscientist} established this paradigm by demonstrating that hypothesis generation, code execution, experimental analysis, and paper writing can be assembled into a single automated workflow. Agent Laboratory~\cite{schmidgall2025agentlab}, AI-Researcher~\cite{airesearcher2025}, CycleResearcher~\cite{cycleresearcher2024}, Kosmos~\cite{kosmos2025}, Dolphin~\cite{dolphin2025}, CodeScientist~\cite{codescientist2025}, and InternAgent~\cite{novelseek2025} instantiate related pipeline designs with different choices of base models, task scopes, and evaluation targets. A more recent pipeline, ScientistOne~\cite{scientistone2026}, augments this idea-to-paper flow with a chain-of-evidence mechanism that explicitly targets fabricated citations and unreproducible results. NVAITC AI Scientist~\cite{huang2026nvaitc} adds explicit governance to the end-to-end pipeline and demonstrates it on a real hypertension GWAS discovery, illustrating the shift toward auditable, domain-grounded research systems.

The advantage of sequential architectures is operational simplicity. Each stage produces an artifact that becomes the input to the next stage, making the workflow interpretable and relatively easy to implement. The limitation is error propagation. A weak idea can lead to irrelevant experiments; incorrect code can produce misleading results; and unsupported experimental claims can be polished into a plausible manuscript. Sequential pipelines therefore expose the same phase-boundary risk observed throughout the lifecycle: producing an artifact at one stage does not guarantee that the next stage represents it faithfully or verifies it adequately.

\subsubsection{Search-Based and Self-Improving Systems}

A second family introduces search, evolution, or self-improvement to avoid the brittleness of one-pass generation. AI Scientist v2~\cite{yamada2025aiscientistv2} uses agentic tree search to explore research trajectories more systematically than its predecessor. ASI-Evolve~\cite{asievolve2026}, AutoSOTA~\cite{autosota2026}, CORAL~\cite{coral2026}, and related evolutionary systems search over architectures, algorithms, data curation strategies, or multi-agent behaviors to discover stronger solutions. EvoMaster~\cite{evomaster2026} proposes a self-evolving foundational agent framework for agentic science at scale, while AutoSci~\cite{autosci2026} centers the full lifecycle on a persistent cross-project memory with self-improving research procedures. Recent closed-loop systems tighten the discover--verify loop and extend it beyond ML: an agentic self-driving lab~\cite{hur2026selfdrivinglab} compresses the validation bottleneck for scientific discovery, iterative meta-reflection~\cite{zhao2026metareflection} refines the research loop through self-critique, and closed-loop agents certify generalizable improvements in molecular property prediction~\cite{ning2026closedloop} and rediscover psychological theories through an automated cognitive scientist~\cite{jagadish2026cognitive}.

Search-based designs are important because research rarely proceeds as a single pass. Strong work typically emerges from branching alternatives, failed experiments, ablation-driven refinement, and selective continuation of promising directions. These systems therefore better match the iterative structure of scientific practice than direct pipelines. However, search alone does not solve validation. Without reliable evaluators, search can optimize toward benchmark-specific artifacts, superficial novelty, or brittle improvements. The core design question is thus not only how broadly a system searches, but what signals guide selection and whether those signals reflect scientific value rather than local metric gains.

\subsubsection{Skill-Based and Tool-Integrated Systems}

Skill-based systems package research workflows as composable capabilities, often built around coding agents, retrieval tools, experiment runners, document editors, and evaluation modules. ARIS~\cite{aris2025} represents this direction by organizing research automation into reusable workflows for idea discovery, auto-review, and paper writing. AutoResearchClaw~\cite{autoresearchclaw2026} similarly implements a multi-stage pipeline with internal agents for coding, benchmarking, and figure generation. Broader tool-integrated systems such as AutoAgent~\cite{tang2025autodeepresearch}, Biomni~\cite{biomni2025}, SciSciGPT~\cite{sciscigpt2025}, and ResearchClaw~\cite{researchclaw2025} emphasize retrieval, analytics, code execution, document understanding, and domain-specific tool use.

The strength of skill-based architectures is modularity. Rather than relying on one monolithic agent to perform all research activities, these systems expose explicit tools and reusable skills for individual operations. This makes it easier to inspect intermediate artifacts, swap components, and insert human checkpoints. The limitation is coordination: modular systems still need reliable state management across stages. If the idea, literature trace, code state, experimental logs, manuscript claims, review feedback, and revision plan are not represented in a shared and updateable workspace, then phase handoffs remain fragile despite the presence of many tools.

\subsubsection{Multi-Agent and Community-Scale Systems}

Multi-agent systems distribute research tasks across specialized agents, such as researchers, engineers, reviewers, analyzers, writers, or simulated community members. FreePhDLabor~\cite{freephdlabor2025}, SciMaster~\cite{scimaster2025}, EvoScientist~\cite{evoscientist2026}, UniScientist~\cite{uniscientist2026}, Medical AI Scientist~\cite{medicalaiscientist2026}, AiScientist-LH~\cite{aiscientistlonghorizon2026}, FARS~\cite{fars2026}, and AutoResearchClaw~\cite{autoresearchclaw2026} illustrate different forms of multi-agent orchestration. Arbor~\cite{arbor2026} adopts a related design in which a long-lived coordinator dispatches short-lived executor agents to drive a long-horizon research loop through hypothesis-tree refinement. Related community-scale systems such as VirSci~\cite{su2024virsci}, AgentRxiv~\cite{schmidgall2025agentrxiv}, and ResearchTown~\cite{researchtown2025} further simulate aspects of scientific collaboration, including idea exchange, manuscript writing, review, and revision.
The motivation for multi-agent architectures is that research requires heterogeneous expertise and adversarial feedback. A single model asked to generate, execute, write, and critique its own work is prone to self-confirmation. Separating roles can reduce this risk by introducing specialization and cross-agent critique. However, multi-agent systems also introduce coordination problems: agents may duplicate work, reinforce shared misconceptions, defer to weak signals, or produce verbose deliberation without improving scientific quality. The strongest multi-agent systems therefore require more than agent count; they require clear role separation, shared memory, grounded tools, and explicit verification mechanisms.

In practice, these four families describe emphases rather than mutually exclusive designs, and the strongest recent systems increasingly combine them: agentic tree search layered over multi-agent role separation~\cite{yamada2025aiscientistv2,evoscientist2026}, or skill-based tool use coupled with persistent cross-project memory and self-improvement~\cite{autosci2026,evomaster2026}. The historical trajectory mirrors this convergence: $2024$ systems were predominantly sequential idea-to-paper pipelines, whereas $2025$--$2026$ systems add search, evolution, shared memory, and explicit critique to counteract one-pass brittleness. Yet across all four families, the decisive factor is consistently the same: not the orchestration pattern itself, but whether the architecture sustains grounded verification and traceable state at the points where one stage hands off to the next.

\subsubsection{Assessment: Lifecycle Coverage and Phase Boundaries}
\label{sec:e2e_assessment}

End-to-end systems should be evaluated not only by the quality of their final manuscript, but also by lifecycle coverage and phase-boundary reliability. Current systems are strongest at generating ideas, code, experiments, and paper drafts, but weaker at external validation, author-style revision, and audience-adaptive dissemination. This uneven coverage is not accidental: Stage 1 (\emph{Creation}) and Stage 2 (\emph{Writing}) produce artifacts, whereas Stage 3 (\emph{Validation}) and Stage 4 (\emph{Dissemination}) require judgment, accountability, and audience-aware fidelity.
The most important failure mode is not isolated stage failure, but unverified handoff. An idea may appear novel but fail during execution; code may run but implement the wrong algorithm; experimental logs may be summarized into unsupported claims; an automated review may be coherent but lenient; and a rebuttal may promise changes that are not fulfilled. End-to-end systems amplify this risk because errors can propagate silently across stages. A mature lifecycle-scale research system must therefore preserve traceable links between hypotheses, retrieved evidence, code, experiments, figures, manuscript claims, reviews, rebuttals, and revisions.

Reported costs and quality metrics further suggest that the token budget alone is not the decisive factor. Systems vary widely in cost, but stronger results often come from search strategy, tool integration, structured decomposition, and verification design rather than brute-force generation. The central evaluation question is therefore shifting from \emph{Can the system produce a paper?} to \emph{Can the system maintain scientific fidelity across the complete lifecycle?}

\subsubsection{Findings and Observations}
\label{sec:e2e_findings}

\stageanalysis{~End-to-End Research Systems}{tableheader}{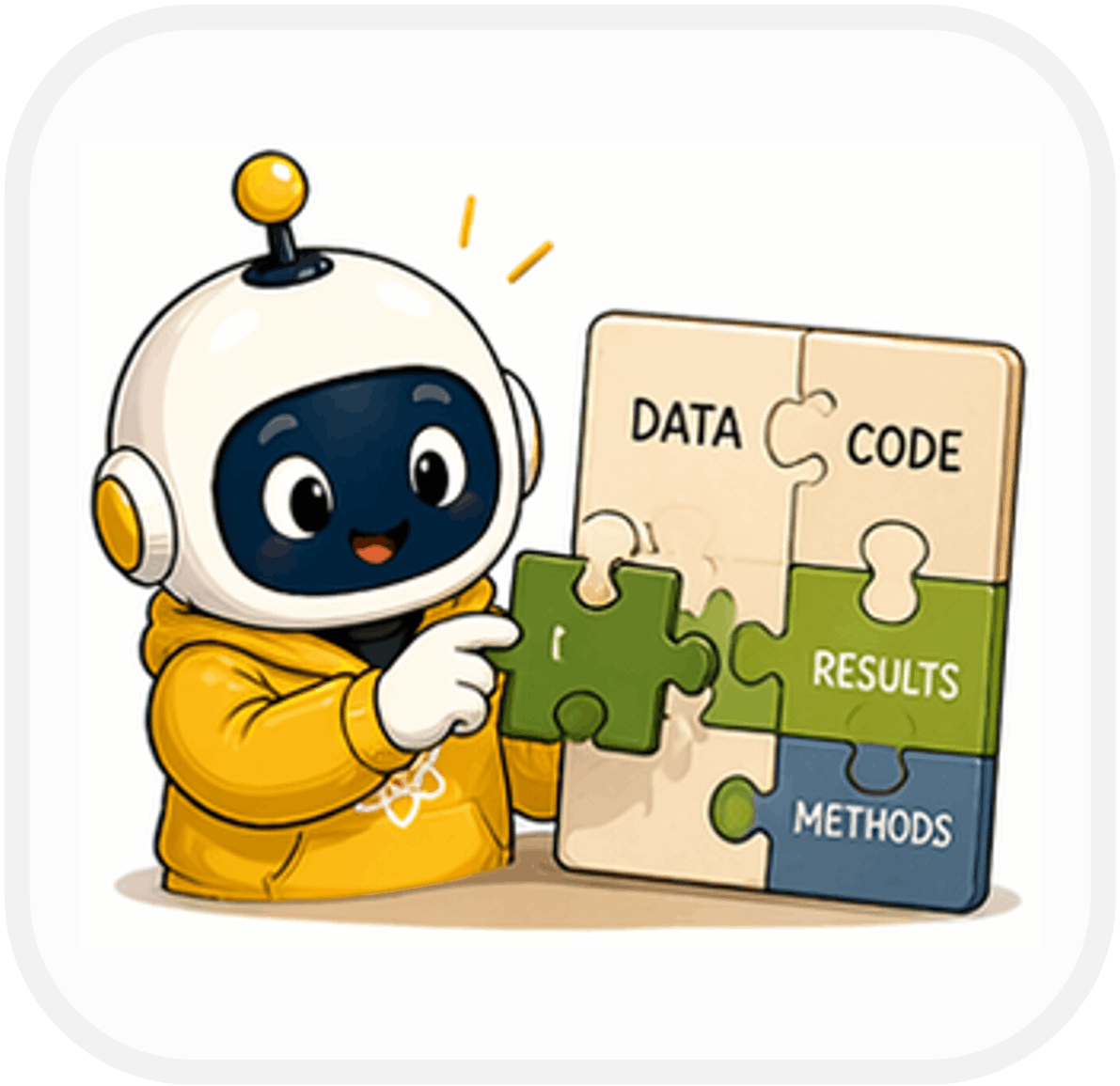}{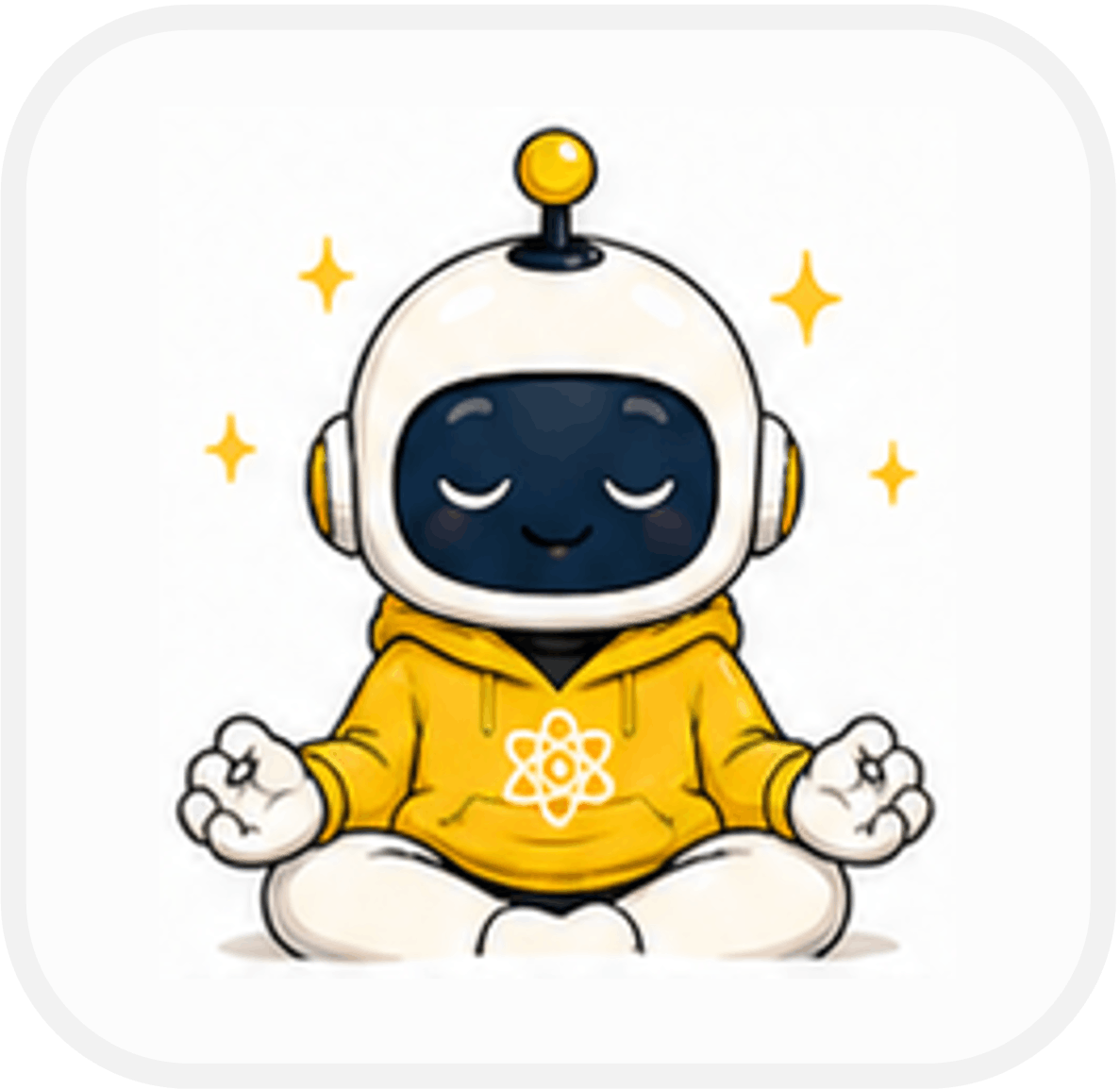}{%

\posbadge[tableheader]{Pipelines} \posbadge[tableheader]{Search} \posbadge[tableheader]{Skills}\par\vspace{2pt}

\posbadge[tableheader]{Multi-Agent} \posbadge[tableheader]{Validation} \posbadge[tableheader]{Handoffs}

}{%

\begin{itemize}[leftmargin=8pt, itemsep=0pt, topsep=0pt, parsep=0pt]

\item End-to-end systems increasingly move beyond linear pipelines toward search-based, skill-based, and multi-agent architectures that better reflect the iterative structure of research.

\end{itemize}

\anadotrule

\begin{itemize}[leftmargin=8pt, itemsep=0pt, topsep=0pt, parsep=0pt]

\item Phase coverage remains uneven: most systems cover \emph{Creation} and \emph{Writing}, while substantially fewer incorporate \emph{Validation}, and none yet provide mature \emph{Dissemination} coverage.

\end{itemize}

\anadotrule

\begin{itemize}[leftmargin=8pt, itemsep=0pt, topsep=0pt, parsep=0pt]

\item Systems that include review, critique, or revision mechanisms point toward more credible lifecycle automation, but their success depends on verification quality rather than review-like text alone.

\end{itemize}

}{%

\negbadge{Propagation} \negbadge{Self-Critique} \negbadge{Fragmented}\par\vspace{2pt}

\negbadge{Validation Gap} \negbadge{State Loss} \negbadge{Overclaiming}

}{%

\begin{itemize}[leftmargin=8pt, itemsep=0pt, topsep=0pt, parsep=0pt]

\item Error propagation is the main lifecycle risk: weak ideas, semantic code errors, unsupported claims, and lenient reviews can compound when phase handoffs are not explicitly verified.

\end{itemize}

\anadotrule

\begin{itemize}[leftmargin=8pt, itemsep=0pt, topsep=0pt, parsep=0pt]

\item Single-model self-critique remains structurally limited; credible validation usually requires role separation, external evidence, adversarial review, or human oversight.

\end{itemize}

\anadotrule

\begin{itemize}[leftmargin=8pt, itemsep=0pt, topsep=0pt, parsep=0pt]

\item Most existing E2E systems do not maintain a fully traceable, updateable state across hypotheses, literature review, coding, experiments, manuscript claims, reviews, and revisions.
\end{itemize}
}
\vspace{8pt}

\subsection{Evaluation Across the Research Lifecycle}
\label{sec:lifecycle_eval}

Evaluation is the central bottleneck for AI-assisted research. Each stage produces different artifacts---ideas, literature summaries, code, experiments, figures, manuscripts, reviews, rebuttals, and dissemination materials---so no single metric can capture research quality across the full lifecycle. Existing benchmarks have therefore evolved from narrow task-specific evaluations toward broader, process-aware, and increasingly execution-grounded protocols. 

\Cref{tab:benchmarks}, introduced in \cref{sec:scope}, summarizes major benchmarks across the eight stages.

Across the benchmark landscape, three trends are clear. First, evaluation is moving from isolated outputs to a multi-dimensional assessment. Early benchmarks often measured a single capability, such as citation prediction, code execution, or writing fluency. Recent benchmarks instead evaluate multiple axes, such as novelty and feasibility in ideation, coverage and citation accuracy in literature synthesis, semantic correctness in research code, review consistency in peer review, and fidelity in Paper2X artifacts. Second, benchmarks are becoming more domain- and workflow-aware. Specialized evaluations now target GPU kernel optimization~\cite{kernelbench2025,tritonbench2025}, biology research~\cite{labbench2024}, scientific experimentation~\cite{expbench2025}, and broader scientist-aligned workflows~\cite{astabench2025,researchclawbench2025}. Third, a persistent gap remains between benchmark performance and real-world research value: systems can perform well on measurable proxies while still producing outputs that experts judge as shallow, incremental, or insufficiently grounded~\cite{llmreviewer2025,cycleresearcher2024}.

\subsubsection{Stage-Specific Benchmarks}
\label{sec:bench_overview}

Stage-specific benchmarks remain necessary because each part of the research lifecycle requires different evaluation criteria. 
\begin{itemize}
    \item For \Sone (\emph{Idea Generation}), benchmarks assess novelty, feasibility, diversity, and downstream potential. These evaluations are difficult because apparent novelty may not survive implementation, and expert judgments of research promise are inherently noisy.

    \item For \Stwo (\emph{Literature Review}), benchmarks emphasize retrieval precision, citation fidelity, coverage completeness, and synthesis quality. The central challenge is not only whether the system finds relevant papers, but whether it uses them faithfully when constructing a narrative.

    \item For \Sthree (\emph{Coding and Experiments}), benchmarks increasingly move beyond code execution toward semantic correctness and reproducibility. Research-code benchmarks ask whether generated implementations match the intended algorithm, while broader workflow benchmarks such as EXP-Bench~\cite{expbench2025} evaluate experiment design, execution, and analysis. NatureBench~\cite{wang2026naturebench} raises the bar further, testing whether coding agents can match the published state of the art of Nature-family papers.

    \item For \Sfour (\emph{Tables and Figures}), evaluation must distinguish visual plausibility from scientific correctness: a figure or table may look publication-ready while misrepresenting data, notation, or comparison structure.

    \item For \Sfive (\emph{Paper Writing}), evaluation combines writing quality, citation accuracy, factual grounding, and review-style judgment.

    \item For \Ssix (\emph{Peer Review}), benchmarks assess consistency, grounding, bias, and robustness to manipulation rather than review fluency alone.

    \item For \Sseven (\emph{Rebuttal and Revision}), emerging datasets align reviews, author responses, and manuscript changes, enabling evaluation of whether rebuttals actually address concerns and lead to fulfilled revisions.

    \item For \Seight (\emph{Dissemination}), evaluation centers on fidelity, usability, and audience adaptation across posters, slides, videos, project pages, social media posts, and interactive agents. This stage is especially difficult because dissemination artifacts often circulate independently of the paper and can shape public understanding of the work.
\end{itemize}

\subsubsection{Evaluation Methodologies}
\label{sec:eval_methods}

Current evaluation methodologies can be grouped into five families. \emph{Expert evaluation} remains the most credible approach for assessing novelty, significance, correctness, and scientific contribution. Si~\etal~\cite{si2024ideas}, for example, recruited over $100$ NLP researchers to evaluate generated ideas. However, expert evaluation is expensive, slow, and noisy: even peer-review-style judgments show limited inter-rater agreement, with the Stanford Agentic Reviewer study reporting human--human correlation around $\rho=0.41$~\cite{stanfordreviewer2025}. This makes expert evaluation indispensable but difficult to scale.

\emph{LLM-as-Judge and Agent-as-Judge} methods provide scalable approximations of human assessment. CycleReviewer~\cite{cycleresearcher2024} reports a $26.89\%$ reduction in Proxy MAE relative to individual human reviewers for score prediction, while the Stanford Agentic Reviewer~\cite{stanfordreviewer2025} achieves review-score correlations comparable to human inter-rater agreement ($\rho=0.42$ vs.\ human $\rho=0.41$). These results show that automated evaluators can provide useful review-style signals, but they remain imperfect proxies. They can exhibit positivity bias, length bias, authority bias, self-preference, and vulnerability to adversarial prompts~\cite{llmreviewer2025,ye2024llmjudgebias}. As a result, LLM-based evaluation is most reliable when calibrated against expert judgments and combined with task-specific verification.

\emph{Automated metrics} offer objective but narrow signals. Code execution success, unit-test pass rates, citation accuracy, acceptance prediction, and traditional text-generation metrics such as BLEU~\cite{papineni2002bleu}, ROUGE~\cite{lin2004rouge}, and BERTScore~\cite{zhang2020bertscore} are easy to compute, but they capture only fragments of research quality. For example, executable code can still be semantically wrong, accurate citations can still be used to support misleading claims, and fluent text can still lack contribution. Over-optimization on any one metric risks Goodhart's law.

\emph{Execution-grounded evaluation} verifies research outputs by running code, reproducing experiments, or checking claims against generated evidence. This paradigm is especially important for \Sthree, but it also affects Writing and Validation because manuscript claims should be traceable to executed experiments. PaperBench~\cite{paperbench2025}, for example, decomposes papers into individually gradable subtasks that can be checked through implementation and execution. Si~\etal~\cite{si2026executiongrounded} further show that execution-guided search can improve discovery workflows by using empirical feedback rather than textual judgment alone.

\emph{Process- and trace-based evaluation} assesses how a system reaches an output, not only the final artifact. This includes tool-use trajectories in deep research, reviewer-comment decomposition in rebuttal, revision fulfillment after author responses, and fidelity between paper content and dissemination materials. This paradigm is increasingly important because many lifecycle failures occur at handoffs: a system may retrieve the right paper but cite it incorrectly, run an experiment but summarize it inaccurately, or promise a rebuttal revision without implementing it.

\subsubsection{Emerging Evaluation Paradigms}
\label{sec:eval_emerging}

Several emerging paradigms are reshaping evaluation for AI-assisted research. The first is \emph{execution-grounded evaluation}, where claims are checked against executable artifacts rather than judged only from text. This is essential for research coding, paper replication, and experimental analysis, where surface plausibility is insufficient. It also provides a path toward evaluating Writing: a manuscript claim should be traceable to a figure, table, log, or executed experiment.

The second is \emph{adversarial evaluation}. As discussed in \Ssix (\emph{Peer Review}), LLM-based reviewers and judges are vulnerable to prompt injection, lexical triggers, and covert content manipulation. Breaking the Reviewer~\cite{breakingreviewer2025} and related studies show that robustness to manipulation must be treated as an evaluation dimension rather than a peripheral security issue. This is particularly important for Validation, where a manipulated review or judge can affect acceptance decisions.

The third is \emph{long-horizon evaluation}. Many benchmarks remain short-horizon, evaluating tasks that take minutes or hours rather than weeks or months. However, real research involves delayed feedback, failed attempts, changing hypotheses, and evolving evidence. METR's analysis suggests that AI task horizons are rapidly increasing~\cite{metr2025forecasting}, while RE-Bench~\cite{rebench2024} provides open-ended ML R\&D environments that begin to approximate longer research workflows. Still, current long-horizon benchmarks remain far shorter and cleaner than authentic research projects.

The fourth is \emph{lifecycle-level evaluation}. Existing benchmarks usually evaluate one stage at a time, but many important failures occur between stages. A future lifecycle benchmark should test whether an idea remains valid after implementation, whether retrieved literature is faithfully represented in writing, whether experimental evidence supports manuscript claims, whether rebuttal commitments are fulfilled, and whether dissemination artifacts preserve claims and limitations. Such evaluation would better match the real risk profile of end-to-end research automation. Recent benchmarks begin to instantiate this lifecycle-level view: ResearchClawBench~\cite{researchclawbench2026} grounds end-to-end research tasks in hidden papers with expert multimodal rubrics, the benchmark suite of~\cite{realresearcher2026} evaluates frontier models and agentic harnesses across the full research lifecycle, and~\cite{researcharena2026} provides a minimal scaffold for running and assessing the complete loop with off-the-shelf agents.

\subsubsection{Evaluation Gaps}
\label{sec:eval_challenges}

Despite rapid progress, several gaps remain unresolved. First, \emph{novelty and significance are still difficult to define}. Expert judgments vary across reviewers, venues, and fields, and automated novelty scores can reward ideas that sound original but fail after execution. Second, \emph{benchmark contamination and temporal validity are persistent concerns}. Many tasks are derived from public papers, code, or reviews that may appear in model training data. Temporal splits help, but they introduce changes in topic, difficulty, and community standards.

Third, \emph{cross-system comparison remains difficult}. Systems are often evaluated with different base models, prompts, tools, datasets, compute budgets, and human-in-the-loop assumptions. This makes reported results hard to compare even when they target the same stage. Fourth, \emph{cross-domain generalization remains under-tested}. Most benchmarks focus on machine learning and NLP, while chemistry, biology, materials science, physics, medicine, and social science require different evidence standards, experimental workflows, and domain-specific tools.

Fifth, \emph{computational cost is itself an evaluation dimension}. Some research tasks, such as paper replication, long-horizon experimentation, and multimodal dissemination, require substantial token, compute, or tool-use budgets. A system that performs well under unlimited sampling may not be practically useful if its cost is prohibitive or its results are not reproducible under realistic constraints.

Finally, \emph{no existing benchmark evaluates the complete research lifecycle with human-equivalent rigor}. PaperBench~\cite{paperbench2025} makes important progress for replication, and process-aware benchmarks are emerging across literature review, coding, peer review, rebuttal, and Paper2X. However, no benchmark yet evaluates the full chain from ideation to dissemination while preserving traceability across artifacts. This lifecycle evaluation gap is central: without it, systems may appear strong within individual stages while failing to maintain scientific fidelity across the research process.

\subsection{Cross-Cutting Insights}
\label{sec:insights}

The preceding sections reveal a consistent pattern across the research lifecycle: AI systems are increasingly capable of producing research-like artifacts, but remain less reliable at verifying whether those artifacts are novel, faithful, executable, and scientifically meaningful. We distill five cross-cutting insights from the stage-level analysis. These insights are not tied to a single tool or benchmark; rather, they describe recurring capability boundaries and deployment principles that appear across Creation, Writing, Validation, and Dissemination.

\subsubsection{Artifact Generation Outpaces Scientific Verification}
\label{sec:insight_generation_verification}

Across the lifecycle, AI systems are better at producing artifacts than at verifying their scientific validity. In \Sone (\emph{Idea Generation}), generated ideas can appear novel and well-motivated, yet weaken after implementation. Si~\etal~\cite{si2025gap} show that AI-generated ideas degrade more sharply after execution than human ideas, exposing a gap between apparent novelty and executable substance. In \Sthree (\emph{Coding and Experiments}), generated code may run successfully while implementing the wrong algorithm, with semantic failures forming a major source of error~\cite{researchcodebench2025}. In \Sfour (\emph{Tables and Figures}), generated visual artifacts may look polished while misrepresenting data, notation, or information flow. In \Sfive (\emph{Paper Writing}), fluent prose can conceal weak reasoning or unsupported claims.

This gap also appears in Validation and Dissemination. In \Ssix (\emph{Peer Review}), automated reviews can be coherent and consistent while under-detecting decisive methodological flaws or assigning inflated scores~\cite{llmreviewer2025,claimcheck2025}. In \Sseven (\emph{Rebuttal and Revision}), generated responses may sound persuasive, but their value depends on whether promised evidence or manuscript changes are actually fulfilled~\cite{rebuttalcommitment2026}. In \Seight (\emph{Dissemination}), posters, slides, videos, and social media summaries can simplify a paper in ways that overstate claims or omit limitations. The central lifecycle problem is therefore not artifact production alone, but artifact verification: each output must remain traceable to evidence, assumptions, and limitations. Recent analyses sharpen this point: a large multi-domain study finds that AI scientists frequently produce results without reasoning scientifically~\cite{noreasoning2026}, and others argue that completing a research-like workflow is not equivalent to achieving scientific closure~\cite{workflowclosure2026}.

\subsubsection{Human-Governed Collaboration Remains the Most Reliable Deployment Mode}
\label{sec:insight_collaboration}

The strongest deployment pattern across stages is not full autonomy, but human-governed collaboration. In Writing, semi-automated systems are most credible when they assist planning, drafting, polishing, and citation support while researchers retain control over argumentation, interpretation, and final responsibility. In Peer Review, the strongest validated setting is not standalone AI review, but AI feedback on human reviews: the ICLR 2025 randomized study shows that LLM feedback improved review quality in $89\%$ of cases without affecting acceptance rates~\cite{iclr2025reviewstudy}. In Rebuttal, author-aware systems are more appropriate than generic response generation because rebuttals must reflect the paper's actual contributions and the authors' intended revisions~\cite{ruan2026authorinloop}. In Dissemination, AI-generated posters, slides, videos, and public summaries are best treated as editable drafts whose claims and emphasis remain under author control.

This pattern explains why direct automation is risky in high-stakes research settings. Research requires judgment under uncertainty: deciding whether an idea is worth pursuing, whether an experiment is sufficient, whether a critique is valid, whether a rebuttal promise is feasible, and whether a public-facing summary is faithful. These decisions are precisely where current systems remain fragile. AI assistance is therefore most useful when it expands researcher capacity while preserving human oversight over scientific claims, evidence interpretation, and accountability.

\subsubsection{Capability Boundaries Emerge in Open-Ended Research Tasks}
\label{sec:insight_open_ended}

The sharpest capability boundaries appear when tasks become novel, underspecified, or long-horizon. Current systems perform strongly on structured tasks with clear feedback, such as standard software issue resolution, grammar correction, simple plotting, and format conversion. Performance drops when the task requires interpreting implicit assumptions, designing meaningful experiments, reproducing underspecified methods, or judging scientific contribution. This is most visible in research coding: while frontier systems perform well on familiar software benchmarks, performance falls sharply on novel research-code tasks, with reported ceilings around $37$--$39\%$ on dedicated benchmarks~\cite{researchcodebench2025,scireplicatebench2025}.

Similar boundaries recur elsewhere. Literature review systems retrieve and summarize individual papers increasingly well, but struggle with multi-paper relational reasoning and citation fidelity. Idea-generation systems produce plausible hypotheses but face persistent novelty--feasibility tradeoffs. Paper-writing systems generate fluent manuscripts but remain weaker at argumentative depth and reviewer anticipation. Peer-review systems can approximate review style but remain vulnerable to leniency, bias, and manipulation. These failures share a common structure: the task cannot be solved by pattern matching alone, because success depends on implicit domain knowledge, causal reasoning, long-horizon feedback, and expert judgment. Position analyses reinforce this boundary, arguing that current agentic AI scientists are structurally better suited to assisting research than to autonomous discovery~\cite{notbuiltauto2026}.

\subsubsection{Effective Systems Converge on Layered Architectures}
\label{sec:insight_architecture}

Across phases, the most capable systems increasingly combine three layers: \emph{exploration}, \emph{execution}, and \emph{verification}. The exploration layer searches over hypotheses, paper collections, code variants, response plans, or design alternatives. The execution layer interacts with tools: retrieval engines, code interpreters, experiment runners, plotting libraries, document editors, or presentation generators. The verification layer checks whether intermediate outputs are grounded, correct, and useful, through execution feedback, citation validation, critique, reviewer simulation, or human review.

This layered view explains why simple prompting is insufficient for research automation. Research tasks rarely require only one generation step; they require proposing alternatives, testing them, revising based on feedback, and preserving state across iterations. Search-based systems improve exploration, tool-integrated systems strengthen execution, and multi-agent systems can support specialization and critique. However, more agents do not automatically improve performance. Multi-agent systems appear most useful when the task can be decomposed into parallel or role-specialized subtasks; they can degrade on sequential reasoning tasks when coordination overhead and error propagation dominate~\cite{googlemit2025scaling}. Thus, the important design principle is not agent count, but whether the architecture matches the task structure and includes reliable verification.

\subsubsection{AI Use Has Become a Governance Problem, Not a Detection Problem}
\label{sec:insight_governance}

AI assistance is already embedded in the research ecosystem. Corpus-level studies estimate detectable AI modification in a nontrivial fraction of scientific writing, including up to $17.5\%$ of computer science abstracts~\cite{liang2024mapping} and $13.5\%$ of biomedical abstracts~\cite{kobak2024delve}. Self-reported adoption is higher, with many researchers using AI for writing or review-related tasks~\cite{aireviewsurvey2025}. At the same time, linguistic marker studies show that AI-associated words can surge after the introduction of LLMs, but such signals are unreliable for adjudicating individual cases and can change as users and models adapt.

The policy implication is that the community should move from detection-centered enforcement toward disclosure, attribution, and accountability. Detection tools can produce false positives, especially for formal or non-native academic prose, while watermarking remains dependent on provider cooperation and robustness to paraphrasing~\cite{watermarking2025}. The more durable governance questions are therefore: What forms of AI assistance must be disclosed? Which uses are allowed during review? Who is responsible for AI-generated claims, citations, rebuttal commitments, or public summaries? How should venues audit high-risk uses without penalizing legitimate writing support? As AI becomes a routine part of research practice, governance must focus less on whether AI was used at all and more on whether its use preserved scientific integrity. Emerging benchmarks begin to measure these integrity risks directly: SciIntegrity-Bench~\cite{sciintegritybench2026} evaluates whether AI-scientist systems uphold academic-integrity norms when honest failure-acknowledgment is the only correct response, while PseudoBench~\cite{pseudobench2026} tests whether auto-research systems resist generating plausible-but-pseudoscientific studies.

\subsection{Open Challenges and Future Directions}
\label{sec:open_challenges}

Despite rapid progress across the research lifecycle, the preceding analysis shows that the main barriers to reliable AI-assisted research are not merely missing tools. The harder problems concern whether AI systems can preserve faithfulness across phase boundaries, evaluate scientific value, verify evidence, support responsible governance, generalize across domains, and preserve human expertise. We organize the remaining challenges around these six themes.

\subsubsection{Faithfulness Across Phase Boundaries}

Many of the most consequential failures occur not within a single stage, but when artifacts move from one phase to the next. An idea that appears promising in \Sone (\emph{Idea Generation}) may weaken after implementation in \Sthree (\emph{Coding and Experiments}); retrieved evidence in \Stwo (\emph{Literature Review}) may be misrepresented in \Sfive (\emph{Paper Writing}); experimental results from \Sthree (\emph{Coding and Experiments}) may be summarized into claims that are stronger than the data support; reviewer concerns in \Ssix (\emph{Peer Review}) may lead to rebuttal promises in \Sseven (\emph{Rebuttal and Revision}) that are not fulfilled; and dissemination outputs in \Seight (\emph{Dissemination}) may simplify the contribution beyond its evidence.

This phase-boundary problem is especially important for end-to-end systems. A lifecycle-scale system must not only generate artifacts, but preserve traceable links between them: hypotheses should connect to retrieved literature, code should connect to experiments, figures should connect to logs, manuscript claims should connect to evidence, rebuttal commitments should connect to revisions, and public-facing summaries should connect to the validated paper. Current systems rarely maintain this level of provenance across the full lifecycle. Future systems should therefore treat phase handoffs as explicit verification checkpoints rather than implicit transitions between modules.

\subsubsection{Scientific Judgment and Novelty Assessment}

Scientific judgment remains difficult to automate because research quality is not reducible to surface novelty, fluency, or benchmark score. In ideation, generated proposals can appear novel before execution but fail to remain feasible or impactful after implementation. Diversity is also a persistent concern: LLM-generated ideas may cluster in narrow regions of the idea space, limiting their ability to explore genuinely distinct research directions~\cite{jiang2025artificialhivemind}. In literature review, systems increasingly retrieve and summarize individual papers well, but still struggle with multi-paper relational reasoning, methodological lineage, and cross-paper contradictions.

The deeper challenge is that novelty, significance, and contribution are socially and temporally situated. A good research idea depends on field-specific context, feasibility, timing, community standards, and the availability of evidence. Automated novelty scoring can therefore reward ideas that sound original while missing whether they are executable, important, or meaningfully different from prior work. Future progress will likely require evaluation methods that combine retrieval, temporal splits, expert judgment, execution feedback, and downstream impact analysis, rather than relying on LLM-as-Judge scores alone.

\subsubsection{Verification, Reproducibility, and Accountability}

Verification is the central unresolved problem for autonomous research systems. In coding and experiments, generated code may execute successfully while implementing the wrong algorithm, and automated experiment runners can produce outputs that appear quantitative without being scientifically meaningful. Paper replication remains particularly difficult: PaperBench~\cite{paperbench2025} shows that current agents still fall far short of human performance on reproducing research results. This indicates that even verifying existing work is not yet solved, let alone generating new work that is independently reproducible.

Rebuttal and revision expose a parallel accountability problem. A rebuttal is scientifically meaningful only if its claims are supported and its commitments are fulfilled. The commitment--fulfillment gap observed in ICLR 2025~\cite{rebuttalcommitment2026} shows that persuasive response text is insufficient: systems must track whether promised experiments, clarifications, and revisions are actually incorporated. Future AI research systems should therefore include explicit evidence ledgers, experiment provenance, versioned manuscript diffs, and revision-tracking mechanisms. The goal is not only to produce stronger artifacts, but to make every claim auditable.

\subsubsection{Citation, Versioning, and Source Provenance}

Citation verification is not solved by adding retrieval or web search. Scientific records are versioned: the same contribution may appear as an arXiv preprint, workshop paper, conference version, journal extension, or revised technical report, with changes to title, authors, venue, year, DOI, and sometimes content. Bibliographic databases may merge or separate these records differently, and prior work on arXiv--publisher citation consolidation shows that version merging is itself a nontrivial bibliometric problem~\cite{gao2020arxivcitationmerging}.

This creates a challenge for AI-assisted literature review, writing, and dissemination. A generated manuscript may cite the correct idea but assemble metadata from inconsistent versions, or quote a claim from one version while citing another. Existing citation-audit tools and benchmarks target fabricated or unsupported references~\cite{citeme2024,scholarcop2025}, but end-to-end research agents also need \emph{version-consistent citation assembly}: title, authors, venue, year, URL/DOI, and quoted claims should come from the same selected record, with provenance preserved. Future systems should therefore treat citation not as a formatting task, but as a versioned source-grounding problem.

\subsubsection{Governance, Disclosure, and Research Integrity}

AI use in research is no longer hypothetical. Writing assistance, review support, literature search, code generation, and dissemination drafting are already part of many researchers' workflows. This makes governance a central challenge. Detection-based enforcement is unreliable because AI text detectors can produce false positives, especially for formal academic writing, non-native prose, or heavily edited text. As discussed in \Sfive (\emph{Paper Writing}) and \Ssix (\emph{Peer Review}), the community is therefore shifting from trying to detect every instance of AI use toward requiring disclosure, attribution, and accountability.

The open question is how to define responsible AI use across stages. Assistance with grammar correction is different from generating experimental claims; drafting a rebuttal is different from promising new experiments; using AI to improve a review is different from delegating the review itself. Venues, publishers, and institutions need policies that distinguish low-risk assistance from high-risk substitution, specify what must be disclosed, and clarify who is accountable for AI-generated content. The central governance principle should be that authors remain responsible for claims, citations, experiments, rebuttal commitments, and public-facing summaries, regardless of which AI tools contributed to their production.

\subsubsection{Cross-Domain Generalization and Infrastructure Access}

Most current systems and benchmarks are concentrated in computer science, machine learning, and NLP. Extending AI-assisted research to chemistry, biology, medicine, materials science, physics, and social science requires more than retraining on domain papers. These fields differ in evidence standards, experimental infrastructure, safety constraints, data availability, and community norms. Systems such as Google AI Co-scientist~\cite{gottweis2025aicoscientist}, Biomni~\cite{biomni2025}, Medical AI Scientist~\cite{medicalaiscientist2026}, and domain-specific laboratory agents point toward this direction, but broad cross-domain generalization remains unresolved.

Infrastructure access is part of the same challenge. Some domains require specialized instruments, wet-lab protocols, proprietary datasets, or expensive compute. If advanced AI research tools are available only to well-resourced laboratories or companies, they may amplify existing inequalities in scientific production. Future systems should therefore be evaluated not only by performance, but also by accessibility, reproducibility, and deployability under realistic resource constraints. Open-source tools, standardized interfaces, shared benchmarks, and transparent provenance mechanisms will be important for preventing research automation from becoming an infrastructure privilege.

\subsubsection{Human Expertise and Cognitive Ownership}

A final challenge concerns the long-term development of researchers themselves. Many AI tools automate the external products of research: summaries, code, plots, manuscripts, reviews, rebuttals, and slides. However, the cognitive value of research lies in forming hypotheses, understanding prior work, diagnosing failures, interpreting results, constructing arguments, and responding to critique. If AI tools bypass these processes too aggressively, they may increase short-term productivity while weakening the skills that define scientific expertise.

This concern is most visible in Writing, where AI assistance is already widely adopted, but it applies across the lifecycle. A junior researcher who delegates literature synthesis may not develop field judgment; one who delegates experiment planning may not learn what makes evidence decisive; one who delegates rebuttal may not learn how to reason from criticism. Tools such as Script\&Shift~\cite{siddiqui2025scriptshift} and DraftMarks~\cite{siddiqui2025draftmarks} suggest a better design direction: AI should support source transformation, process transparency, and reflective revision rather than replacing the user's cognitive engagement. The practical principle is that AI should handle mechanical, repetitive, or scaffolded tasks, while humans retain ownership of judgment, interpretation, argumentation, and accountability.

\subsubsection{Toward Reliable AI-Assisted Research}

Taken together, these challenges suggest a shift in the goal of AI-assisted research. The near-term objective should not be fully autonomous science in which AI systems independently generate, validate, publish, and promote research without oversight. A more credible objective is reliable human-governed research automation: systems that expand the scale and speed of research while preserving traceability, verification, expert judgment, and accountability.

Future progress will likely come from systems that integrate four design principles. First, they should maintain provenance across the full lifecycle, linking ideas, evidence, code, figures, claims, reviews, rebuttals, and dissemination artifacts. Second, they should use execution and retrieval grounding wherever possible, replacing purely textual self-judgment with verifiable signals. Third, they should include human checkpoints at phase boundaries, where errors are most likely to propagate. Fourth, they should make AI involvement transparent, so that readers, reviewers, and institutions can assess how a research artifact was produced. These principles define the path from artifact-generating systems toward trustworthy research collaborators.

\section{Conclusion}
\label{sec:conclusion}

This work presented an end-to-end analysis of AI-assisted academic research across the complete lifecycle. We organized the field into four epistemological phases: \emph{Creation}, \emph{Writing}, \emph{Validation}, and \emph{Dissemination}, and eight stages spanning ideation, literature review, coding and experiments, tables and figures, paper writing, peer review, rebuttal and revision, and Paper2X dissemination. This lifecycle framing connects tools that are often studied in isolation and exposes where current systems succeed, where they fail, and how errors propagate across stage boundaries.

The central finding is that AI systems are increasingly capable of producing research artifacts, but remain less reliable at verifying their scientific meaning. Across the lifecycle, plausible outputs can conceal deeper failures: ideas may weaken after execution, retrieved evidence may be misrepresented, executable code may implement the wrong algorithm, fluent manuscripts may lack argumentative depth, reviews may miss methodological flaws, rebuttals may promise unfulfilled revisions, and dissemination artifacts may overstate claims. The core bottleneck is therefore not generation alone, but maintaining novelty, faithfulness, reproducibility, and accountability across the research process.

The most credible path forward is human-governed AI-assisted research. AI should reduce mechanical friction in retrieval, drafting, coding, visualization, review support, and dissemination, while researchers retain ownership over judgment, interpretation, experimental design, argumentation, and final responsibility. Future systems should maintain provenance across artifacts, use retrieval and execution grounding wherever possible, support human checkpoints at phase boundaries, and make AI involvement transparent. If developed with these principles, AI can amplify human creativity and rigor; without them, it risks scaling the production of plausible but unreliable research artifacts.

\subsection*{Acknowledgments}
We thank Josh Susskind for insightful discussions and careful proofreading of this manuscript. 

We also thank the researchers and open-source contributors whose systems, benchmarks, datasets, and technical reports made this survey possible, as well as the broader community for ongoing discussions on responsible AI use, research integrity, peer review, and the future of scientific work.

\subsection*{Responsible Use and Limitations}
This work is intended to inform responsible use of AI-assisted research tools, not to endorse replacing human scientific judgment with full automation. Current systems are most reliable when used to assist retrieval, drafting, coding, visualization, review support, and dissemination, while humans retain responsibility for novelty, interpretation, verification, authorship, and accountability. Because the field evolves rapidly, this paper should be read as a structured snapshot through our search cutoff, and AI-generated research outputs should be independently verified before scholarly use.

\clearpage\clearpage
\ifarxivmode
  \beginappendix
\else
  \appendices
\fi

\section{Auto-Research Tool Inventory}
\label{sec:appendix_inventory}

This appendix provides a comprehensive inventory of all surveyed works, organized by stage.

\subsection{Phase 1: Creation}
\begingroup
\renewcommand{\arraystretch}{1.15}
\setlength{\tabcolsep}{3pt}
\footnotesize
\rowcolors{2}{white}{S1color!6}
\begin{table*}[!ht]
\centering
\caption{\textbf{Comprehensive inventory: \Sone Idea Generation.} $^\dagger$Evaluation information might be uncertain.}
\vspace{-7pt}
\label{tab:appendix_s1}
\stagecard{\Sone}{Idea Generation}{S1color}{figures/icons/s1_ideation.png}{%
Generating, refining, and evaluating novel research hypotheses. Techniques range from knowledge graph reasoning and retrieval-augmented generation to Multi-Agent collaboration for structured hypothesis formation.}
\vspace{2pt}
\resizebox{\linewidth}{!}{\begin{tabular}{c| >{\centering\arraybackslash}p{3cm} r r |c| >{\centering\arraybackslash}p{2.4cm} |c| p{8.5cm}}
\toprule
\rowcolor{tableheader!10}
\textbf{\#} & \textbf{Method} & \textbf{Ref} & \textbf{Venue} & \textbf{Link} & \textbf{Category} & \textbf{GitHub} & \textbf{Evaluation} \\
\midrule
\multicolumn{8}{@{}l}{\cellcolor{S1color!65}\textbf{\textsf{\textcolor{white}{~~LLM Internal Knowledge-Based Generation}}}} \\
\addlinespace[1pt]
{1} & Chain of Ideas & \cite{li2024chainofideas} & {\small arXiv'24}  & \href{https://arxiv.org/abs/2410.13185}{\faExternalLink} & LLM Internal    & \githubicon{https://github.com/DAMO-NLP-SG/CoI-Agent}                                   & Comparable to human quality; \$0.50/idea min cost \\
{2} & ResearchAgent & \cite{baek2024researchagent} & {\small NAACL'25} & \href{https://aclanthology.org/2025.naacl-long.342/}{\faExternalLink} & LLM Internal    & \githubicon{https://github.com/JinheonBaek/ResearchAgent}                                & Human + model eval; academic graph feedback \\
{3} & SciMON & \cite{wang2024scimon} & {\small ACL'24} & \href{https://arxiv.org/abs/2305.14259}{\faExternalLink} & LLM Internal    & \githubicon{https://github.com/EagleW/CLBD}                                             & Mitigates shallow novelty; iterative refinement \\
{4} & Idea Gen Agent & \cite{si2024ideas} & {\small arXiv'24} & \href{https://arxiv.org/abs/2409.04109}{\faExternalLink} & LLM Internal    & -                                                                                     & 100+ NLP researchers; LLM ideas higher novelty ($p<0.05$) \\
{5} & IRIS & \cite{iris2025} & {\small ACL'25} & \href{https://aclanthology.org/2025.acl-demo.57/}{\faExternalLink} & LLM Internal    & \githubicon{https://github.com/Anikethh/IRIS-Interactive-Research-Ideation-System}       & MCTS adaptive reasoning; human-in-the-loop platform \\
{6} & Spark & \cite{sanyal2025spark} & {\small ICCC'25} & \href{https://arxiv.org/abs/2504.20090}{\faExternalLink} & LLM Internal    & -                                                                                     & Judge model trained on 600K OpenReview reviews \\
\addlinespace[3pt]
\multicolumn{8}{@{}l}{\cellcolor{S1color!65}\textbf{\textsf{\textcolor{white}{~~External Signal-Driven Generation}}}} \\
\addlinespace[1pt]
{7} & MOOSE-Chem & \cite{yang2024moosechem} & {\small ICLR'25} & \href{https://openreview.net/forum?id=X9OfMNNepI}{\faExternalLink} & External Signal  & -                                                                                     & Rediscovers hypotheses from 51 high-impact papers \\
{8} & Nova & \cite{hu2024nova} & {\small arXiv'24} & \href{https://arxiv.org/abs/2410.14255}{\faExternalLink} & External Signal  & -                                                                                     & 3.4$\times$ more novel ideas; 2.5$\times$ more top-rated \\
{9} & SciAgents & \cite{ghafarollahi2024sciagents} & {\small arXiv'24} & \href{https://arxiv.org/abs/2409.05556}{\faExternalLink} & External Signal  & \githubicon{https://github.com/lamm-mit/SciAgentsDiscovery}                              & Multi-agent reasoning over knowledge graphs \\
{10} & SciPIP & \cite{wang2024scipip} & {\small arXiv'24} & \href{https://arxiv.org/abs/2410.23166}{\faExternalLink} & External Signal  & \githubicon{https://github.com/cheerss/SciPIP}                                           & Multi-domain; paper-anchored idea generation \\
{11} & IdeaSynth & \cite{pu2025ideasynth} & {\small CHI'25} & \href{https://arxiv.org/abs/2410.04025}{\faExternalLink} & External Signal  & -                                                                                     & 20-user study; more alternatives explored vs baseline \\
{12} & MOOSE-Chem2 & \cite{yang2025moosechem2} & {\small NeurIPS'25} & \href{https://nips.cc/virtual/2025/poster/118171}{\faExternalLink} & External Signal  & -                                                                                     & Fine-grained, experimentally actionable hypotheses \\
\addlinespace[3pt]
\multicolumn{8}{@{}l}{\cellcolor{S1color!65}\textbf{\textsf{\textcolor{white}{~~Multi-Agent Collaborative Generation}}}} \\
\addlinespace[1pt]
{13} & Combi. Creativity & \cite{gu2024combinatorial} & {\small arXiv'24} & \href{https://arxiv.org/abs/2412.14141}{\faExternalLink} & Multi-Agent   & -                                                                                     & +7--10\% similarity scores; cross-domain composition \\
{14} & Deep Ideation & \cite{zhao2025deepideation} & {\small arXiv'25} & \href{https://arxiv.org/abs/2511.02238}{\faExternalLink} & Multi-Agent      & \githubicon{https://github.com/kyZhao-1/Deep-Ideation}                                   & +10.67\% quality; surpasses conference acceptance levels \\
{15} & VirSci & \cite{su2024virsci} & {\small ACL'25} & \href{https://aclanthology.org/2025.acl-long.1368/}{\faExternalLink} & Multi-Agent      & \githubicon{https://github.com/open-sciencelab/Virtual-Scientists}                       & Outperforms single-agent on novelty$^\dagger$ \\
{16} & Multi-Agent Dial. & \cite{sigdial2025multiagent} & {\small SIGDIAL'25} & \href{https://arxiv.org/abs/2507.08350}{\faExternalLink} & Multi-Agent   & -                                                                                     & Optimal at 3 critique-revision rounds$^\dagger$ \\
{17} & Artificial Hivemind & \cite{jiang2025artificialhivemind} & {\small NeurIPS'25} & \href{https://arxiv.org/abs/2510.22954}{\faExternalLink} & Multi-Agent & -                                                                                     & 26K queries; diversity collapse across models \\
{18} & Auditable AI Sci. & \cite{takahara2026auditable} & {\small arXiv'26} & \href{https://arxiv.org/abs/2607.09195}{\faExternalLink} & Multi-Agent & - & Hypothesis evolution protocol for auditable LLM scientists \\
\addlinespace[3pt]
\multicolumn{8}{@{}l}{\cellcolor{S1color!65}\textbf{\textsf{\textcolor{white}{~~Novelty and Feasibility Assessment}}}} \\
\addlinespace[1pt]
{19} & IdeaBench & \cite{guo2025ideabench} & {\small KDD'25} & \href{https://doi.org/10.1145/3711896.3737419}{\faExternalLink} & Evaluation       & -                                                                                     & 2,374 papers; 8 domains; novelty $>$0.6, feasibility $<$0.5 \\
{20} & LiveIdeaBench & \cite{liveideabench2024} & {\small arXiv'24} & \href{https://arxiv.org/abs/2412.17596}{\faExternalLink} & Evaluation       & -                                                                                     & 40+ models; 1,180 keywords; 22 scientific domains \\
{21} & AI Idea Bench 2025 & \cite{aiideabench2025} & {\small arXiv'25} & \href{https://arxiv.org/abs/2504.14191}{\faExternalLink} & Evaluation       & \githubicon{https://github.com/yansheng-qiu/AI_Idea_Bench_2025}                          & 3,495 papers; alignment + general reference eval \\
{22} & HeurekaBench & \cite{heurekabench2026} & {\small ICLR'26} & \href{https://arxiv.org/abs/2601.01678}{\faExternalLink} & Evaluation       & \githubicon{https://github.com/mlbio-epfl/HeurekaBench}                                  & +22\% with critic module; open-ended science tasks \\
{23} & ResearchBench & \cite{researchbench2025} & {\small ACL'26} & \href{https://arxiv.org/abs/2503.21248}{\faExternalLink} & Evaluation       & -                                                                                     & 12 disciplines; inspiration retrieval + ranking \\
{24} & HindSight & \cite{hindsight2026} & {\small arXiv'26} & \href{https://arxiv.org/abs/2603.15164}{\faExternalLink} & Evaluation       & -                                                                                     & LLM novelty negatively correlated with impact ($\rho$=$-$0.29) \\
{25} & Rubric Rewards & \cite{rubricrewards2025} & {\small arXiv'25} & \href{https://arxiv.org/abs/2512.23707}{\faExternalLink} & LLM Internal & - & 70\% expert preference; RL with rubric self-grading \\
{26} & DeepInnovator & \cite{deepinnovator2026} & {\small arXiv'26} & \href{https://arxiv.org/abs/2602.18920}{\faExternalLink} & LLM Internal & \githubicon{https://github.com/HKUDS/DeepInnovator} & 80--94\% win rates vs frontier; 14B model \\
{27} & FlowPIE & \cite{flowpie2026} & {\small arXiv'26} & \href{https://arxiv.org/abs/2603.29557}{\faExternalLink} & External Signal & - & Higher novelty, feasibility, diversity vs baselines \\
{28} & Diverse Hypo.\ Search & \cite{diversehypothesis2026} & {\small arXiv'26} & \href{https://arxiv.org/abs/2606.10587}{\faExternalLink} & LLM Internal & - & Diversity-aware population search; molecular/equation/algorithm discovery \\
{29} & SoundnessBench & \cite{soundnessbench2026} & {\small arXiv'26} & \href{https://arxiv.org/abs/2605.30329}{\faExternalLink} & Evaluation & - & Tests if AI scientists distinguish sound vs.\ flawed ideas; optimism bias \\
{30} & LLM-Judge Novelty & \cite{llmjudgenovelty2026} & {\small arXiv'26} & \href{https://arxiv.org/abs/2606.12071}{\faExternalLink} & Evaluation & - & Limits of LLM-as-Judge for novelty assessment \\
\bottomrule
\end{tabular}}
\vspace{-1cm}
\end{table*}
\endgroup

\clearpage
\begingroup
\renewcommand{\arraystretch}{1.15}
\setlength{\tabcolsep}{3pt}
\footnotesize
\rowcolors{2}{white}{S2color!6}
\begin{table*}[!ht]
\centering
\caption{\textbf{Comprehensive inventory: \Stwo Literature Review.} $^\dagger$Evaluation information might be uncertain.}
\vspace{-7pt}
\label{tab:appendix_s2}
\stagecard{\Stwo}{Literature Review}{S2color}{figures/icons/s2_literature.png}{%
Retrieving, synthesizing, and organizing prior work into coherent narratives. Modern approaches span semantic retrieval, citation-graph traversal, and Deep Research agents that iteratively explore the literature.}
\vspace{2pt}
\resizebox{\linewidth}{!}{\begin{tabular}{c| >{\centering\arraybackslash}p{3cm} r r |c| >{\centering\arraybackslash}p{2.3cm} |c| p{9cm}}
\toprule
\rowcolor{tableheader!10}
\textbf{\#} & \textbf{Method} & \textbf{Ref} & \textbf{Venue} & \textbf{Link} & \textbf{Category} & \textbf{GitHub} & \textbf{Evaluation} \\
\midrule
\multicolumn{8}{@{}l}{\cellcolor{S2color!65}\textbf{\textsf{\textcolor{white}{~~Literature Retrieval}}}} \\
\addlinespace[1pt]
{1} & CiteME & \cite{citeme2024} & {\small arXiv'24} & \href{https://arxiv.org/abs/2407.12861}{\faExternalLink} & Retrieval        & -                                                                                     & Citation fidelity benchmark \\
{2} & LitLLM & \cite{agarwal2024litllm} & {\small arXiv'24} & \href{https://arxiv.org/abs/2402.01788}{\faExternalLink} & Retrieval        & -                                                                                     & LLM + academic database integration \\
{3} & LitSearch & \cite{litsearch2024} & {\small arXiv'24} & \href{https://arxiv.org/abs/2407.18940}{\faExternalLink} & Retrieval        & \githubicon{https://github.com/princeton-nlp/LitSearch}                                  & Retrieval precision benchmark \\
{4} & PaperQA2 & \cite{skarlinski2024paperqa2} & {\small arXiv'24} & \href{https://arxiv.org/abs/2409.13740}{\faExternalLink} & Retrieval        & \githubicon{https://github.com/Future-House/paper-qa}                                    & Matches/exceeds expert on 3 tasks; 70\% contradiction val. \\
{5} & OpenResearcher & \cite{li2024openresearcher} & {\small EMNLP'24} & \href{https://arxiv.org/abs/2408.09578}{\faExternalLink} & Retrieval        & -                                                                                     & RAG + graph traversal for literature exploration \\
{6} & PaSa & \cite{pasa2025} & {\small arXiv'25} & \href{https://arxiv.org/abs/2501.10120}{\faExternalLink} & Retrieval        & \githubicon{https://github.com/bytedance/pasa}                                           & Agentic multi-step iterative retrieval \\
\addlinespace[3pt]
\multicolumn{8}{@{}l}{\cellcolor{S2color!65}\textbf{\textsf{\textcolor{white}{~~Survey \& Related Work Generation}}}} \\
\addlinespace[1pt]
{7} & ChatPaper & \cite{chatpaper2023} & {\small GitHub'23} & \href{https://github.com/kaixindelele/ChatPaper}{\faExternalLink} & Generation       & \githubicon{https://github.com/kaixindelele/ChatPaper}                                   & 19K+ GitHub stars; arXiv summarization tool \\
{8} & PaperQA & \cite{paperqa2024github} & {\small arXiv'23} & \href{https://arxiv.org/abs/2312.07559}{\faExternalLink} & Generation       & \githubicon{https://github.com/Future-House/paper-qa}                                    & 8K+ GitHub stars; RAG for scientific Q\&A \\
{9} & AutoSurvey & \cite{wang2024autosurvey} & {\small arXiv'24} & \href{https://arxiv.org/abs/2406.10252}{\faExternalLink} & Generation       & \githubicon{https://github.com/AutoSurveys/AutoSurvey}                                   & First end-to-end LLM survey drafting system \\
{10} & GPT Researcher & \cite{gptresearcher2024} & {\small GitHub'24} & \href{https://github.com/assafelovic/gpt-researcher}{\faExternalLink} & Generation       & \githubicon{https://github.com/assafelovic/gpt-researcher}                               & 26K+ GitHub stars; comprehensive report generation \\
{11} & LLMs for Lit. & \cite{emnlp2025litreview} & {\small arXiv'24} & \href{https://arxiv.org/abs/2412.13612}{\faExternalLink} & Generation       & -                                                                                     & Hallucination analysis; models still generate errors$^\dagger$ \\
{12} & STORM & \cite{shao2024storm} & {\small arXiv'24} & \href{https://arxiv.org/abs/2402.14207}{\faExternalLink} & Generation       & \githubicon{https://github.com/stanford-oval/storm}                                      & Multi-perspective question-asking for outlines \\
{13} & Agentic AutoSurvey & \cite{agenticautosurvey2025} & {\small arXiv'25} & \href{https://arxiv.org/abs/2509.18661}{\faExternalLink} & Generation       & -                                                                                     & Multi-agent role decomposition$^\dagger$ \\
{14} & Citegeist & \cite{beger2025citegeist} & {\small arXiv'25} & \href{https://arxiv.org/abs/2503.23229}{\faExternalLink} & Generation       & -                                                                                     & Dynamic RAG pipeline on arXiv corpus \\
{15} & IterSurvey & \cite{itersurvey2025} & {\small arXiv'25} & \href{https://arxiv.org/abs/2510.21900}{\faExternalLink} & Generation       & \githubicon{https://github.com/HancCui/IterSurvey_Autosurveyv2}                          & Iterative outline planning with stability checks \\
{16} & LiRA & \cite{lira2025} & {\small arXiv'25} & \href{https://arxiv.org/abs/2510.05138}{\faExternalLink} & Generation       & -                                                                                     & Multi-agent retrieval + verification + narrative \\
{17} & SurveyForge & \cite{gao2025surveyforge} & {\small arXiv'25} & \href{https://arxiv.org/abs/2503.04629}{\faExternalLink} & Generation       & \githubicon{https://github.com/Alpha-Innovator/SurveyForge}                              & Outperforms AutoSurvey on outline quality$^\dagger$ \\
{18} & SurveyG & \cite{surveyg2025} & {\small arXiv'25} & \href{https://arxiv.org/abs/2510.07733}{\faExternalLink} & Generation       & -                                                                                     & Three-layer citation graph (Foundation/Dev/Frontier) \\
{19} & SurveyX & \cite{liang2025surveyx} & {\small arXiv'25} & \href{https://arxiv.org/abs/2502.14776}{\faExternalLink} & Generation       & -                                                                                     & +0.259 content quality improvement; near expert level \\
{20} & InteractiveSurvey & \cite{interactivesurvey2025} & {\small arXiv'25} & \href{https://arxiv.org/abs/2504.08762}{\faExternalLink} & Generation       & \githubicon{https://github.com/TechnicolorGUO/InteractiveSurvey}                        & User-customizable reference categorization + outlines \\
{21} & CiteLLM & \cite{citellm2026} & {\small arXiv'26} & \href{https://arxiv.org/abs/2602.23075}{\faExternalLink} & Generation       & -                                                                                     & Hallucination-free via trusted repository routing \\
{22} & STRUCTSURVEY & \cite{pedinotti2026structsurvey} & {\small arXiv'26} & \href{https://arxiv.org/abs/2607.01243}{\faExternalLink} & Generation & - & Structured agentic retrieval for survey generation \\
\addlinespace[3pt]
\multicolumn{8}{@{}l}{\cellcolor{S2color!65}\textbf{\textsf{\textcolor{white}{~~Deep Research Agents}}}} \\
\addlinespace[1pt]
{23} & ASReview & \cite{asreview2020} & {\small Nature MI'21} & \href{https://www.nature.com/articles/s42256-020-00287-7}{\faExternalLink} & Deep Research & \githubicon{https://github.com/asreview/asreview}                                        & Active learning; up to 95\% effort reduction \\
{24} & CHIME & \cite{kang2024chime} & {\small arXiv'24} & \href{https://arxiv.org/abs/2407.16148}{\faExternalLink} & Deep Research    & -                                                                                     & Hierarchical organization of scientific studies \\
{25} & DeepResearch-Agent & \cite{deepresearchagent2025} & {\small GitHub'25} & \href{https://github.com/SkyworkAI/DeepResearchAgent}{\faExternalLink} & Deep Research    & \githubicon{https://github.com/SkyworkAI/DeepResearchAgent}                              & Hierarchical multi-agent; planner + sub-agents \\
{26} & DeerFlow & \cite{deerflow2025} & {\small GitHub'25} & \href{https://github.com/bytedance/deer-flow}{\faExternalLink} & Deep Research    & \githubicon{https://github.com/bytedance/deer-flow}                                      & Sub-agents with shared memory; sandboxed execution \\
{27} & OpenScholar & \cite{openscholar2025} & {\small Nature'26} & \href{https://doi.org/10.1038/s41586-025-10072-4}{\faExternalLink} & Deep Research    & -                                                                                     & 45M papers; +6.1\% over GPT-4o, +5.5\% over PaperQA2 \\
{28} & AutoAgent & \cite{tang2025autodeepresearch} & {\small arXiv'25} & \href{https://arxiv.org/abs/2502.05957}{\faExternalLink} & Deep Research    & -                                                                                     & Universal LLM compatibility; GAIA benchmark \\
{29} & Tongyi DeepResearch & \cite{tongyi2025deepresearch} & {\small GitHub'25} & \href{https://github.com/Alibaba-NLP/DeepResearch}{\faExternalLink} & Deep Research    & \githubicon{https://github.com/Alibaba-NLP/DeepResearch}                                 & 30.5B params (3.3B activated); SOTA on Deep Research \\
{30} & O-Researcher & \cite{oresearcher2026} & {\small arXiv'26} & \href{https://arxiv.org/abs/2601.03743}{\faExternalLink} & Deep Research    & -                                                                                     & Multi-agent distillation + agentic RL \\
{31} & OpenResearcher & \cite{li2026openresearcher} & {\small arXiv'26} & \href{https://arxiv.org/abs/2603.20278}{\faExternalLink} & Deep Research    & \githubicon{https://github.com/TIGER-AI-Lab/OpenResearcher}                              & 54.8\% BrowseComp-Plus; 97K+ trajectories \\
\addlinespace[3pt]
\multicolumn{8}{@{}l}{\cellcolor{S2color!65}\textbf{\textsf{\textcolor{white}{~~Retrieval and Synthesis Quality Assessment}}}} \\
\addlinespace[1pt]
{32} & DeepScholar-Bench & \cite{deepscholar2025} & {\small arXiv'25} & \href{https://arxiv.org/abs/2508.20033}{\faExternalLink} & Evaluation       & \githubicon{https://github.com/guestrin-lab/deepscholar-bench}                            & Coverage, coherence, factual accuracy benchmark \\
{33} & ReportBench & \cite{reportbench2025} & {\small arXiv'25} & \href{https://arxiv.org/abs/2508.15804}{\faExternalLink} & Evaluation       & \githubicon{https://github.com/ByteDance-BandAI/ReportBench}                             & 100-prompt benchmark from 678 filtered survey papers \\
{34} & IDRBench & \cite{idrbench2026} & {\small arXiv'26} & \href{https://arxiv.org/abs/2601.06676}{\faExternalLink} & Evaluation       & -                                                                                     & 100 tasks; interactive Deep Research evaluation \\
{35} & ScholarGym & \cite{scholargym2026} & {\small arXiv'26} & \href{https://arxiv.org/abs/2601.21654}{\faExternalLink} & Evaluation       & -                                                                                     & 2,536 queries; query planning + tool invocation \\
{36} & SciNetBench & \cite{scinetbench2026} & {\small arXiv'26} & \href{https://arxiv.org/abs/2601.03260}{\faExternalLink} & Evaluation       & -                                                                                     & 18M papers; relation-aware retrieval $<$20\% \\
{37} & Self-Evolving Retrieval & \cite{selfevolveretrieval2026} & {\small arXiv'26} & \href{https://arxiv.org/abs/2605.14306}{\faExternalLink} & Retrieval & - & Agent that adapts its own literature-search policy over time \\
{38} & MasterSet & \cite{masterset2026} & {\small arXiv'26} & \href{https://arxiv.org/abs/2604.17680}{\faExternalLink} & Retrieval & - & Large-scale must-cite citation recommendation \\
{39} & DeepSurvey & \cite{deepsurvey2026} & {\small arXiv'26} & \href{https://arxiv.org/abs/2605.29522}{\faExternalLink} & Generation & - & Evidence-constrained citation assignment \\
{40} & AutoResearchBench & \cite{autoresearchbench2026} & {\small arXiv'26} & \href{https://arxiv.org/abs/2604.25256}{\faExternalLink} & Evaluation & - & Benchmark for complex multi-hop scientific discovery \\
{41} & PaperMind & \cite{zhao2026papermind} & {\small arXiv'26} & \href{https://arxiv.org/abs/2604.21304}{\faExternalLink} & Evaluation & - & Multimodal reasoning + critique over papers \\
\bottomrule
\end{tabular}}
\end{table*}
\endgroup

\clearpage
\begingroup
\renewcommand{\arraystretch}{1.15}
\setlength{\tabcolsep}{3pt}
\footnotesize
\rowcolors{2}{white}{S3color!6}
\begin{table*}[!ht]
\centering
\caption{\textbf{Comprehensive inventory: \Sthree Coding \& Experiments.} $^\dagger$Evaluation information might be uncertain.}
\vspace{-7pt}
\label{tab:appendix_s3}
\stagecard{\Sthree}{Coding \& Experiments}{S3color}{figures/icons/s3_coding.png}{%
Translating ideas into executable code, running experiments at scale, and analyzing results. This stage spans code generation, Paper-to-Code translation, autonomous experiment orchestration, and result interpretation.}
\vspace{2pt}
\resizebox{\linewidth}{!}{\begin{tabular}{c| >{\centering\arraybackslash}p{3cm} r r |c| >{\centering\arraybackslash}p{2.3cm} |c| p{9cm}}
\toprule
\rowcolor{tableheader!10}
\textbf{\#} & \textbf{Method} & \textbf{Ref} & \textbf{Venue} & \textbf{Link} & \textbf{Category} & \textbf{GitHub} & \textbf{Evaluation} \\
\midrule
\multicolumn{8}{@{}l}{\cellcolor{S3color!65}\textbf{\textsf{\textcolor{white}{~~Code Generation}}}} \\
\addlinespace[1pt]
{1} & SWE-bench & \cite{swebench2024} & {\small ICLR'24} & \href{https://arxiv.org/abs/2310.06770}{\faExternalLink} & Code Gen.        & \githubicon{https://github.com/princeton-nlp/SWE-bench}                                  & 2,294 real GitHub issues; Verified split (500 problems) \\
{2} & SWE-agent & \cite{yang2024sweagent} & {\small arXiv'24} & \href{https://arxiv.org/abs/2405.15793}{\faExternalLink} & Code Gen.        & \githubicon{https://github.com/princeton-nlp/SWE-agent}                                  & Agent--computer interface paradigm for coding \\
{3} & OpenHands & \cite{wang2024openhands} & {\small ICLR'25} & \href{https://arxiv.org/abs/2407.16741}{\faExternalLink} & Code Gen.        & \githubicon{https://github.com/All-Hands-AI/OpenHands}                                   & Open platform for generalist coding agents \\
{4} & SWE-bench Pro & \cite{deng2025swebenchpro} & {\small arXiv'25} & \href{https://arxiv.org/abs/2509.16941}{\faExternalLink} & Code Gen.        & -                                                                                     & 1,865 enterprise problems; best score 23\% \\
{5} & SWE-EVO & \cite{thai2025sweevo} & {\small arXiv'25} & \href{https://arxiv.org/abs/2512.18470}{\faExternalLink} & Code Gen.        & -                                                                                     & Software evolution benchmark; best score 25\% \\
\addlinespace[3pt]
\multicolumn{8}{@{}l}{\cellcolor{S3color!65}\textbf{\textsf{\textcolor{white}{~~Paper-to-Code}}}} \\
\addlinespace[1pt]
{6} & FunSearch & \cite{funsearch2024} & {\small Nature'24} & \href{https://www.nature.com/articles/s41586-023-06924-6}{\faExternalLink} & Paper-to-Code    & \githubicon{https://github.com/google-deepmind/funsearch}                                & New cap-set solutions; evolutionary program search \\
{7} & SciCode & \cite{scicode2024} & {\small arXiv'24} & \href{https://arxiv.org/abs/2407.13168}{\faExternalLink} & Paper-to-Code    & \githubicon{https://github.com/scicode-bench/SciCode}                                    & Research-level coding across math, physics, chemistry \\
{8} & PaperBench & \cite{paperbench2025} & {\small arXiv'25} & \href{https://arxiv.org/abs/2504.01848}{\faExternalLink} & Paper-to-Code    & \githubicon{https://github.com/openai/preparedness}                                      & 20 ICML'24 papers; 8,316 gradable subtasks \\
{9} & PaperCoder & \cite{papercoder2025} & {\small arXiv'25} & \href{https://arxiv.org/abs/2504.17192}{\faExternalLink} & Paper-to-Code    & \githubicon{https://github.com/going-doer/Paper2Code}                                    & 3-stage multi-agent; ML papers to code repos \\
{10} & ResearchCodeBench & \cite{researchcodebench2025} & {\small arXiv'25} & \href{https://arxiv.org/abs/2506.02314}{\faExternalLink} & Paper-to-Code    & -                                                                                     & 212 novel ML tasks; best 37.3\% (Gemini-2.5-Pro) \\
{11} & SciReplicate-Bench & \cite{scireplicatebench2025} & {\small arXiv'25} & \href{https://arxiv.org/abs/2504.00255}{\faExternalLink} & Paper-to-Code    & \githubicon{https://github.com/xyzCS/SciReplicate-Bench}                                 & 100 tasks from 36 NLP papers; 39\% ceiling \\
\addlinespace[3pt]
\multicolumn{8}{@{}l}{\cellcolor{S3color!65}\textbf{\textsf{\textcolor{white}{~~Experiment Execution \& Orchestration}}}} \\
\addlinespace[1pt]
{12} & BioPlanner & \cite{bioplanner2024} & {\small arXiv'23} & \href{https://arxiv.org/abs/2310.10632}{\faExternalLink} & Execution        & \githubicon{https://github.com/bioplanner/bioplanner}                                    & Biological protocol planning evaluation \\
{13} & CRISPR-GPT & \cite{crisprgpt2024} & {\small arXiv'24} & \href{https://arxiv.org/abs/2404.18021}{\faExternalLink} & Execution        & -                                                                                     & Gene-editing experiment design assistance \\
{14} & DS-Agent & \cite{dsagent2024} & {\small arXiv'24} & \href{https://arxiv.org/abs/2402.17453}{\faExternalLink} & Execution        & \githubicon{https://github.com/guosyjlu/DS-Agent}                                        & End-to-end data science workflow automation \\
{15} & MLE-Bench & \cite{chan2024mlebench} & {\small arXiv'24} & \href{https://arxiv.org/abs/2410.07095}{\faExternalLink} & Execution        & -                                                                                     & 75 Kaggle competitions benchmark \\
{16} & MLAgentBench & \cite{mlagentbench2024} & {\small arXiv'24} & \href{https://arxiv.org/abs/2310.03302}{\faExternalLink} & Execution        & \githubicon{https://github.com/snap-stanford/MLAgentBench}                               & 13 ML experimentation tasks benchmark \\
{17} & MLR-Copilot & \cite{mlrcopilot2024} & {\small arXiv'24} & \href{https://arxiv.org/abs/2408.14033}{\faExternalLink} & Execution        & -                                                                                     & IdeaAgent + ExperimentAgent dual-agent pipeline \\
{18} & AIDE & \cite{jiang2025aide} & {\small arXiv'25} & \href{https://arxiv.org/abs/2502.13138}{\faExternalLink} & Execution        & -                                                                                     & SOTA on MLE/RE-Bench; tree search in code space \\
{19} & AlphaEvolve & \cite{alphaevolve2025} & {\small arXiv'25} & \href{https://arxiv.org/abs/2506.13131}{\faExternalLink} & Execution        & -                                                                                     & LLM-generated mutations + automated evaluators \\
{20} & AutoReproduce & \cite{autoreproduce2025} & {\small arXiv'25} & \href{https://arxiv.org/abs/2505.20662}{\faExternalLink} & Execution        & \githubicon{https://github.com/AI9Stars/AutoReproduce}                                   & Paper lineage algorithm for experiment reproduction \\
{21} & CURIE & \cite{curie2025} & {\small arXiv'25} & \href{https://arxiv.org/abs/2502.16069}{\faExternalLink} & Execution        & \githubicon{https://github.com/Just-Curieous/Curie}                                      & Rigorous automated experimentation framework \\
{22} & MLGym & \cite{mlgym2025} & {\small arXiv'25} & \href{https://arxiv.org/abs/2502.14499}{\faExternalLink} & Execution        & -                                                                                     & AI research agent gym benchmark \\
{23} & MLR-Bench & \cite{mlrbench2025} & {\small arXiv'25} & \href{https://arxiv.org/abs/2505.19955}{\faExternalLink} & Execution        & -                                                                                     & 201 tasks (NeurIPS/ICLR/ICML); 80\% fabrication rate \\
{24} & Exe.-Grounded & \cite{si2026executiongrounded} & {\small arXiv'26} & \href{https://arxiv.org/abs/2601.14525}{\faExternalLink} & Execution       & -                                                                                     & 69.4\% vs 48.0\% GRPO; parallel GPU search \\
{25} & Learn to Discover & \cite{yuksekgonul2026learntodiscover} & {\small arXiv'26} & \href{https://arxiv.org/abs/2601.16175}{\faExternalLink} & Execution  & -                                                                                     & Test-time training + RL; math, GPU kernel, biology \\
{26} & SciNav & \cite{scinav2026} & {\small arXiv'26} & \href{https://arxiv.org/abs/2603.20256}{\faExternalLink} & Execution        & -                                                                                     & Pairwise tree-search branch selection \\
{27} & FrontierScience & \cite{wang2026frontierscience} & {\small arXiv'26} & \href{https://arxiv.org/abs/2601.21165}{\faExternalLink} & Execution        & -                                                                                     & Expert-level tasks; Olympiad + PhD difficulty \\
{28} & AutoTTS & \cite{zheng2026llms} & {\small arXiv'26} & \href{https://arxiv.org/abs/2605.08083}{\faExternalLink} & Execution & \githubicon{https://github.com/zhengkid/AutoTTS} & Coding-agent discovery of test-time scaling strategies \\
{29} & AutoScientists & \cite{gao2026autoscientists} & {\small arXiv'26} & \href{https://arxiv.org/abs/2605.28655}{\faExternalLink} & Execution & - & Self-organizing agent teams for long-running experiments \\
{30} & EurekAgent & \cite{xin2026eurekagent} & {\small arXiv'26} & \href{https://arxiv.org/abs/2606.13662}{\faExternalLink} & Execution & - & Agent environment engineering for autonomous discovery \\
\addlinespace[3pt]
\multicolumn{8}{@{}l}{\cellcolor{S3color!65}\textbf{\textsf{\textcolor{white}{~~Code Correctness and Reproducibility Assessment}}}} \\
\addlinespace[1pt]
{31} & DiscoveryBench & \cite{majumder2024discoverybench} & {\small arXiv'24} & \href{https://arxiv.org/abs/2407.01725}{\faExternalLink} & Analysis         & \githubicon{https://github.com/allenai/discoverybench}                                   & Data-driven insight extraction benchmark \\
{32} & DiscoveryWorld & \cite{discoveryworld2024} & {\small arXiv'24} & \href{https://arxiv.org/abs/2406.06769}{\faExternalLink} & Analysis         & \githubicon{https://github.com/allenai/discoveryworld}                                   & 120 tasks; 8 topics; 3 difficulty levels \\
{33} & InfiAgent & \cite{infidabench2024} & {\small arXiv'24} & \href{https://arxiv.org/abs/2401.05507}{\faExternalLink} & Analysis         & -                                                                                     & End-to-end data analysis workflow benchmark \\
{34} & ScienceAgentBench & \cite{scienceagentbench2024} & {\small arXiv'24} & \href{https://arxiv.org/abs/2410.05080}{\faExternalLink} & Analysis         & -                                                                                     & Rigorous data-driven scientific discovery assessment \\
{35} & LAB-Bench & \cite{labbench2024} & {\small arXiv'24} & \href{https://arxiv.org/abs/2407.10362}{\faExternalLink} & Execution & \githubicon{https://github.com/Future-House/LAB-Bench} & Multi-domain biology research task benchmark \\
{36} & KernelBench & \cite{kernelbench2025} & {\small arXiv'25} & \href{https://arxiv.org/abs/2502.10517}{\faExternalLink} & Execution & \githubicon{https://github.com/ScalingIntelligence/KernelBench} & GPU kernel generation benchmark \\
{37} & TritonBench & \cite{tritonbench2025} & {\small arXiv'25} & \href{https://arxiv.org/abs/2502.14752}{\faExternalLink} & Execution & \githubicon{https://github.com/thunlp/TritonBench} & Triton operator generation benchmark \\
{38} & AstaBench & \cite{astabench2025} & {\small arXiv'25} & \href{https://arxiv.org/abs/2510.21652}{\faExternalLink} & Execution & \githubicon{https://github.com/allenai/asta-bench} & 2,400+ problems; multi-domain scientific research \\
{39} & ResearchClawBench & \cite{researchclawbench2025} & {\small arXiv'25} & \href{https://arxiv.org/abs/2512.16969}{\faExternalLink} & Execution & \githubicon{https://github.com/InternScience/ResearchClawBench} & Scientist-aligned workflow benchmark \\
{40} & EXP-Bench & \cite{expbench2025} & {\small ICLR'26} & \href{https://openreview.net/forum?id=KjgyAm383Z}{\faExternalLink} & Execution & \githubicon{https://github.com/Just-Curieous/Curie/tree/main/benchmark/exp_bench} & 461 tasks from 51 AI papers \\
{41} & PostTrainBench & \cite{posttrainbench2026} & {\small arXiv'26} & \href{https://arxiv.org/abs/2603.08640}{\faExternalLink} & Execution & \githubicon{https://github.com/aisa-group/PostTrainBench} & LLM post-training automation benchmark \\
{42} & EvoDS & \cite{evods2026} & {\small arXiv'26} & \href{https://arxiv.org/abs/2606.03841}{\faExternalLink} & Execution & - & Self-evolving DS agent; skill learn + context mgmt \\
{43} & MLReplicate & \cite{mlreplicate2026} & {\small arXiv'26} & \href{https://arxiv.org/abs/2605.16616}{\faExternalLink} & Analysis & - & End-to-end ML reproducibility benchmark 
\\
{44} & BeyondSWE & \cite{chen2026beyondswe} & {\small arXiv'26} & \href{https://arxiv.org/abs/2603.03194}{\faExternalLink} & Execution & \githubicon{https://github.com/AweAI-Team/BeyondSWE} & 500 instances / 246 repos; cross-repo, domain-fix, doc2repo \\
{45} & NatureBench & \cite{wang2026naturebench} & {\small arXiv'26} & \href{https://arxiv.org/abs/2606.24530}{\faExternalLink} & Analysis & - & Tests if coding agents match Nature-family paper SOTA \\
\bottomrule
\end{tabular}}
\end{table*}
\endgroup

\clearpage
\begingroup
\renewcommand{\arraystretch}{1.15}
\setlength{\tabcolsep}{3pt}
\footnotesize
\rowcolors{2}{white}{S4color!6}
\begin{table*}[!ht]
\centering
\caption{\textbf{Comprehensive inventory: \Sfour Tables \& Figures.} $^\dagger$Evaluation information might be uncertain.}
\vspace{-7pt}
\label{tab:appendix_s4}
\stagecard{\Sfour}{Tables \& Figures}{S4color}{figures/icons/s4_figures.png}{%
Creating Method Diagrams, result plots, comparison tables, and LaTeX formulas. Scientific visualization transforms raw experimental outputs into publication-quality charts, illustrations, and structured tables.}
\vspace{2pt}
\resizebox{\linewidth}{!}{\begin{tabular}{c|>{\centering\arraybackslash}p{3cm} r r |c| >{\centering\arraybackslash}p{2.7cm} |c| p{8cm}}
\toprule
\rowcolor{tableheader!10}
\textbf{\#} & \textbf{Method} & \textbf{Ref} & \textbf{Venue} & \textbf{Link} & \textbf{Category} & \textbf{GitHub} & \textbf{Evaluation} \\
\midrule
\multicolumn{8}{@{}l}{\cellcolor{S4color!65}\textbf{\textsf{\textcolor{white}{~~Scientific Figure Generation}}}} \\
\addlinespace[1pt]
{1} & ChartGPT & \cite{yuan2023chartgpt} & {\small arXiv'23} & \href{https://arxiv.org/abs/2311.01920}{\faExternalLink} & Data Viz         & -                                                                                     & 6-step reasoning for chart generation \\
{2} & MatPlotAgent & \cite{matplotagent2024} & {\small arXiv'24} & \href{https://arxiv.org/abs/2402.11453}{\faExternalLink} & Data Viz         & -                                                                                     & +12.3 over GPT-4 base; VLM visual feedback$^\dagger$ \\
{3} & CoDA & \cite{coda2025} & {\small arXiv'25} & \href{https://arxiv.org/abs/2510.03194}{\faExternalLink} & Data Viz         & -                                                                                     & +41.5\% over baselines; multi-agent collaboration \\
{4} & PlotGen & \cite{plotgen2025} & {\small arXiv'25} & \href{https://arxiv.org/abs/2502.00988}{\faExternalLink} & Data Viz         & -                                                                                     & 4--6\% improvement over baselines$^\dagger$ \\
{5} & VIS-Shepherd & \cite{visshepherd2025} & {\small arXiv'25} & \href{https://arxiv.org/abs/2506.13326}{\faExternalLink} & Figure Editing   & -                                                                                     & Constructive critique feedback framework \\
{6} & DiagramAgent & \cite{diagramagent2024} & {\small CVPR'25} & \href{https://arxiv.org/abs/2411.11916}{\faExternalLink} & Data Viz         & -                                                                                     & 4 specialized agents; 8 diagram categories \\
{7} & StarVector & \cite{starVector2025} & {\small CVPR'25} & \href{https://arxiv.org/abs/2312.11556}{\faExternalLink} & Method Diagrams  & -                                                                                     & Scalable SVG generation from descriptions$^\dagger$ \\
{8} & VisCoder & \cite{viscoder2025} & {\small EMNLP'25} & \href{https://arxiv.org/abs/2506.03930}{\faExternalLink} & Data Viz         & -                                                                                     & VisCode-200K dataset; 90\%+ execution pass rate$^\dagger$ \\
{9} & AI-Generated Figures & \cite{aifigurepolicies2026} & {\small arXiv'26} & \href{https://arxiv.org/abs/2603.16159}{\faExternalLink} & Policy           & -                                                                                     & Publisher policy survey (Nature, Science, etc.) \\
{10} & AutoFigure-Edit & \cite{autofigure2026} & {\small arXiv'26} & \href{https://arxiv.org/abs/2603.06674}{\faExternalLink} & Method Diagrams  & \githubicon{https://github.com/ResearAI/AutoFigure-Edit}                & Editable text-to-SVG scientific illustrations$^\dagger$ \\
{11} & AutoFigure & \cite{autofigure2026iclr} & {\small ICLR'26} & \href{https://arxiv.org/abs/2602.03828}{\faExternalLink} & Method Diagrams  & \githubicon{https://github.com/ResearAI/AutoFigure}                     & FigureBench (3,300 pairs); publication-ready illust. \\
{12} & PaperBanana & \cite{paperbanana2026} & {\small arXiv'26} & \href{https://arxiv.org/abs/2601.23265}{\faExternalLink} & Method Diagrams  & -                                                                                     & 292 test cases; outperforms baselines$^\dagger$ \\
{13} & SAIL & \cite{sail2026} & {\small arXiv'26} & \href{https://arxiv.org/abs/2603.18145}{\faExternalLink} & Figure Editing   & -                                                                                     & Domain logic / code syntax separation \\
{14} & Can AI Draw Sci. & \cite{chen2026candraw} & {\small arXiv'26} & \href{https://arxiv.org/abs/2606.28406}{\faExternalLink} & Data Viz & - & Benchmark for scientific figure generation by T2I/MLLMs \\
\addlinespace[3pt]
\multicolumn{8}{@{}l}{\cellcolor{S4color!65}\textbf{\textsf{\textcolor{white}{~~Table Understanding \& Generation}}}} \\
\addlinespace[1pt]
{15} & ArxivDIGESTables & \cite{arxivdigestables2024} & {\small EMNLP'24} & \href{https://arxiv.org/abs/2410.22360}{\faExternalLink} & Table Gen.        & -                                                                                     & Literature comparison table synthesis \\
{16} & Chain-of-Table & \cite{chainoftable2024} & {\small ICLR'24} & \href{https://arxiv.org/abs/2401.04398}{\faExternalLink} & Table Reasoning  & -                                                                                     & Multi-step table reasoning chains \\
{17} & ShowTable & \cite{showtable2025} & {\small CVPR'26} & \href{https://arxiv.org/abs/2512.13303}{\faExternalLink} & Table Viz        & -                                                                                     & Collaborative reflection and refinement$^\dagger$ \\
{18} & Table2LaTeX-RL & \cite{table2latexrl2025} & {\small arXiv'25} & \href{https://arxiv.org/abs/2509.17589}{\faExternalLink} & Table Conversion & -                                                                                     & Image-to-LaTeX via reinforced multimodal LM \\
\addlinespace[3pt]
\multicolumn{8}{@{}l}{\cellcolor{S4color!65}\textbf{\textsf{\textcolor{white}{~~Mathematical Formulas \& TikZ}}}} \\
\addlinespace[1pt]
{19} & AutomaTikZ & \cite{belouadi2024automatikz} & {\small ICLR'24} & \href{https://arxiv.org/abs/2310.00367}{\faExternalLink} & TikZ             & -                                                                                     & DaTikZ: first large-scale dataset (120K drawings) \\
{20} & DeTikZify & \cite{belouadi2024detikzify} & {\small NeurIPS'24} & \href{https://arxiv.org/abs/2405.15306}{\faExternalLink} & TikZ            & -                                                                                     & 360K TikZ graphics; MCTS iterative refinement \\
{21} & TikZilla & \cite{tikzilla2026} & {\small arXiv'26} & \href{https://arxiv.org/abs/2603.03072}{\faExternalLink} & TikZ             & -                                                                                     & 3B/8B matches GPT-5; SFT+RL on expanded DaTikZ \\
\addlinespace[3pt]
\multicolumn{8}{@{}l}{\cellcolor{S4color!65}\textbf{\textsf{\textcolor{white}{~~Visual Fidelity and Scientific Accuracy Assessment}}}} \\
\addlinespace[1pt]
{22} & PlotCraft & \cite{plotcraft2025} & {\small arXiv'25} & \href{https://arxiv.org/abs/2511.00010}{\faExternalLink} & Benchmark         & -                                                                                     & 1K-task benchmark; 48 chart types \\
{23} & TeXpert & \cite{texpert2025} & {\small SDP'25} & \href{https://aclanthology.org/2025.sdp-1.2/}{\faExternalLink} & Benchmark         & -                                                                                     & 3-level difficulty; 78.8\%/58.7\%/17.5\%$^\dagger$ \\
{24} & AbGen & \cite{abgen2025} & {\small ACL'25} & \href{https://arxiv.org/abs/2507.13300}{\faExternalLink} & Benchmark         & -                                                                                     & 1,500 ablation studies; 807 NLP papers \\
{25} & SciFig & \cite{scifig2026} & {\small arXiv'26} & \href{https://arxiv.org/abs/2601.04390}{\faExternalLink} & Benchmark         & -                                                                                     & Rubric-based evaluation; 2K+ pipeline figures \\
{26} & SciFlow-Bench & \cite{scifigbench2026} & {\small arXiv'26} & \href{https://arxiv.org/abs/2602.09809}{\faExternalLink} & Benchmark         & -                                                                                     & 500 framework figures; inverse-parsing evaluation \\
{27} & FigureBench & \cite{autofigure2026iclr} & {\small ICLR'26} & \href{https://arxiv.org/abs/2602.03828}{\faExternalLink} & Benchmark         & \githubicon{https://github.com/ResearAI/AutoFigure}                     & 3,300 text-figure pairs; publication-ready eval \\
{28} & Crafter & \cite{crafter2026} & {\small arXiv'26} & \href{https://arxiv.org/abs/2605.30611}{\faExternalLink} & Method Diagrams & - & Multi-agent editable figure generation from heterogeneous inputs \\
{29} & DiagramRAG & \cite{diagramrag2026} & {\small arXiv'26} & \href{https://arxiv.org/abs/2605.27931}{\faExternalLink} & Method Diagrams & - & Retrieval-augmented scientific diagram generation \\
{30} & GeoSVG-RL & \cite{geosvgrl2026} & {\small arXiv'26} & \href{https://arxiv.org/abs/2605.25447}{\faExternalLink} & Method Diagrams & - & Geometry-aware RL for layout-constrained text-to-SVG diagrams \\
{31} & CSPO & \cite{cspo2026} & {\small arXiv'26} & \href{https://arxiv.org/abs/2604.10918}{\faExternalLink} & Table Conversion & - & Component-specific optimization for table-to-LaTeX generation \\
\bottomrule
\end{tabular}}
\end{table*}
\endgroup

\clearpage
\subsection{Phase 2: Writing}

\begingroup
\renewcommand{\arraystretch}{1.15}
\setlength{\tabcolsep}{3pt}
\footnotesize
\rowcolors{2}{white}{S5color!6}
\begin{table*}[!ht]
\centering
\caption{\textbf{Comprehensive inventory: \Sfive Paper Writing.} $^\dagger$Evaluation information might be uncertain.}
\vspace{-7pt}
\label{tab:appendix_s5}
\stagecard{\Sfive}{Paper Writing}{S5color}{figures/icons/s5_writing.png}{%
Drafting, editing, and polishing academic manuscripts. AI assistance ranges from semi-automated grammar and citation tools to fully automated paper generation: the most commercially mature yet ethically contested stage.}
\vspace{2pt}
\resizebox{\linewidth}{!}{\begin{tabular}{c| >{\centering\arraybackslash}p{3cm} r r |c| >{\centering\arraybackslash}p{2.5cm} |c| p{8cm}}
\toprule
\rowcolor{tableheader!10}
\textbf{\#} & \textbf{Method} & \textbf{Ref} & \textbf{Venue} & \textbf{Link} & \textbf{Category} & \textbf{GitHub} & \textbf{Evaluation} \\
\midrule
\multicolumn{8}{@{}l}{\cellcolor{S5color!65}\textbf{\textsf{\textcolor{white}{~~Semi-Automated Writing Assistance}}}} \\
\addlinespace[1pt]
{1}  & CoAuthor & \cite{coauthor2022} & {\small arXiv'22} & \href{https://arxiv.org/abs/2201.06796}{\faExternalLink} & Collaborative    & -                                                                     & Human--AI collaborative writing workflows \\
{2}  & Script\&Shift & {\small \cite{siddiqui2025scriptshift}} & {\small CHI'25} & \href{https://arxiv.org/abs/2502.07440}{\faExternalLink}   & Source Transform & -                                                                     & CHI Honorable Mention; preserves cognitive engagement \\
{3}  & AI Writing Study & \cite{siddiqui2025aiwriting} & {\small AIED'25} & \href{https://arxiv.org/abs/2506.20595}{\faExternalLink} & Empirical Study  & -                                                                     & 90-student RCT; purposeful AI fosters writing \\
{4}  & OpenDraft & \cite{opendraft2025} & - & \href{https://github.com/federicodeponte/opendraft}{\faExternalLink} & Full Draft Gen.   & \githubicon{https://github.com/federicodeponte/opendraft}                & 19 agents; 20K+ words in 10 min; verified citations \\
{5}  & DraftMarks & \cite{siddiqui2025draftmarks} & {\small arXiv'25} & \href{https://arxiv.org/abs/2509.23505}{\faExternalLink} & Transparency     & -                                                                     & Skeuomorphic visual traces for AI process transparency \\
{6}  & PaperDebugger & \cite{paperdebugger2025} & {\small arXiv'25} & \href{https://arxiv.org/abs/2512.02589}{\faExternalLink} & In-Editor Assist & \githubicon{https://github.com/PaperDebugger/PaperDebugger}              & Multi-agent Overleaf plugin (Reviewer+Enhancer+Scorer) \\
{7}  & ScholarCopilot & \cite{scholarcop2025} & {\small arXiv'25} & \href{https://arxiv.org/abs/2504.00824}{\faExternalLink} & Citation Assist  & -                                                                     & 40.1\% top-1 citation accuracy (vs 15.0\% E5-Mistral) \\
{8}  & XtraGPT & \cite{xtragpt2025} & {\small arXiv'25} & \href{https://arxiv.org/abs/2505.11336}{\faExternalLink} & Post-Writing     & -                                                                     & 1.5B--14B models; 7K papers; 140K revision pairs \\
{9}  & LimAgents & \cite{limagents2026} & {\small arXiv'26} & \href{https://arxiv.org/abs/2601.11578}{\faExternalLink} & Limitations Gen. & -                                                                     & OpenReview comments + citation network integration \\
\addlinespace[3pt]
\multicolumn{8}{@{}l}{\cellcolor{S5color!65}\textbf{\textsf{\textcolor{white}{~~Fully Automated Paper Generation}}}} \\
\addlinespace[1pt]
{1} & CycleResearcher & \cite{cycleresearcher2024} & {\small ICLR'25} & \href{https://arxiv.org/abs/2411.00816}{\faExternalLink} & E2E Gen.   & -                                                                     & 5.36 ICLR scale (vs 5.24 preprint, 5.69 accepted) \\
{2} & Agent Laboratory & \cite{schmidgall2025agentlab} & {\small EMNLP'25} & \href{https://aclanthology.org/2025.findings-emnlp.320/}{\faExternalLink} & E2E Gen.   & -                                                                     & \$2--13/paper; 84\% cost reduction; 3.5--4.0 score \\
{3} & FutureGen & \cite{futuregen2025} & {\small arXiv'25} & \href{https://arxiv.org/abs/2503.16561}{\faExternalLink} & Section Gen.      & -                                                                     & RAG-based Future Work section generation \\
{4} & AI Scientist & \cite{lu2024aiscientist} & {\small Nature'26} & \href{https://arxiv.org/abs/2408.06292}{\faExternalLink} & E2E Gen.   & \githubicon{https://github.com/SakanaAI/AI-Scientist}                    & \$15/paper; end-to-end across 3 ML subfields \\
{5} & APRES & \cite{apres2026} & {\small arXiv'26} & \href{https://arxiv.org/abs/2603.03142}{\faExternalLink} & Rubric Revision  & -                                                                     & 79\% expert preference; citation-predictive rubrics \\
{6} & RWGBench & \cite{xie2026rwgbench} & {\small arXiv'26} & \href{https://arxiv.org/abs/2606.24894}{\faExternalLink} & Section Gen. & - & Evaluates scholarly positioning in related-work generation \\
\addlinespace[3pt]
\multicolumn{8}{@{}l}{\cellcolor{S5color!65}\textbf{\textsf{\textcolor{white}{~~Societal Analysis}}}} \\
\addlinespace[1pt]
{7} & AI Writing Adoption & \cite{aicontractfocus2025} & {\small Nature'26} & \href{https://www.nature.com/articles/s41586-025-08681-8}{\faExternalLink} & Measurement      & -                                                                     & 41.3M papers; AI expands impact but contracts focus \\
{8} & Nature AI Survey & \cite{aireviewsurvey2025} & {\small Nature'26} & \href{https://www.nature.com/articles/d41586-025-04066-5}{\faExternalLink} & Survey           & -                                                                     & 57\% of researchers use AI in peer review \\
{9} & Denial of Science & \cite{gyevnar2026dds} & {\small arXiv'26} & \href{https://arxiv.org/abs/2607.10712}{\faExternalLink} & Measurement & - & Indirect data poisoning could industrialize scientific fraud \\
{10} & AI Slop OSS & \cite{afroz2026aislop} & {\small arXiv'26} & \href{https://arxiv.org/abs/2607.04003}{\faExternalLink} & Measurement & - & AI-generated contributions strain open-source sustainability \\
\addlinespace[3pt]
\multicolumn{8}{@{}l}{\cellcolor{S5color!65}\textbf{\textsf{\textcolor{white}{~~Writing Quality and AI Detection Assessment}}}} \\
\addlinespace[1pt]
{11} & Mapping LLM Use & \cite{liang2024mapping} & {\small arXiv'24} & \href{https://arxiv.org/abs/2404.01268}{\faExternalLink} & Detection        & -                                                                     & Up to 17.5\% of CS papers AI-modified \\
{12} & CycleReviewer & \cite{cycleresearcher2024} & {\small ICLR'25} & \href{https://arxiv.org/abs/2411.00816}{\faExternalLink} & AI Judge         & -                                                                     & 26.89\% MAE reduction vs individual human reviewers \\
{13} & Stanford Agentic & \cite{stanfordreviewer2025} & {\small Web'25} & \href{https://paperreview.ai/tech-overview}{\faExternalLink} & AI Judge         & -                                                                     & $\rho=0.42$ vs human $\rho=0.41$; matches consistency \\
{14} & SciIG & \cite{sciig2025} & {\small arXiv'25} & \href{https://arxiv.org/abs/2508.14273}{\faExternalLink} & Writing Bench    & -                                                                     & NAACL/ICLR 2025 introduction writing benchmark$^\dagger$ \\
{15} & Watermarking & \cite{watermarking2025} & {\small arXiv'25} & \href{https://arxiv.org/abs/2503.15772}{\faExternalLink} & Detection        & -                                                                     & Near-zero false-positive rate under controlled conditions \\
{16} & PaperWritingBench & \cite{paperwritingbench2026} & {\small arXiv'26} & \href{https://arxiv.org/abs/2604.05018}{\faExternalLink} & Benchmark        & -                                                                     & 200 reverse-engineered top-tier conference papers \\
{17} & PaperMentor & \cite{papermentor2026} & {\small arXiv'26} & \href{https://arxiv.org/abs/2606.08857}{\faExternalLink} & In-Editor Assist & - & Human-centered multi-agent Overleaf writing tutor \\
{18} & LECTOR & \cite{lector2026} & {\small arXiv'26} & \href{https://arxiv.org/abs/2605.25964}{\faExternalLink} & Section Gen. & - & Introduction generation via joint reasoning-graph optimization \\
{19} & CiteTracer & \cite{citetracer2026} & {\small arXiv'26} & \href{https://arxiv.org/abs/2605.08583}{\faExternalLink} & Detection & - & Multi-agent citation-hallucination detection \\
{20} & Process Eval & \cite{processeval2026} & {\small arXiv'26} & \href{https://arxiv.org/abs/2606.15583}{\faExternalLink} & Writing Bench & - & Keystroke-level study of AI- vs.\ human-draft revision \\
\bottomrule
\end{tabular}}
\end{table*}
\endgroup

\clearpage
\subsection{Phase 3: Validation}

\begingroup
\renewcommand{\arraystretch}{1.15}
\setlength{\tabcolsep}{3pt}
\footnotesize
\rowcolors{2}{white}{S6color!6}
\begin{table*}[!ht]
\centering
\caption{\textbf{Comprehensive inventory: \Ssix Peer Review.} $^\dagger$Evaluation information might be uncertain.}
\vspace{-7pt}
\label{tab:appendix_s6}
\stagecard{\Ssix}{Peer Review}{S6color}{figures/icons/s6_review.png}{%
Automated Review Generation, reviewer-paper matching, and review quality assessment. AI systems can produce structured critiques and predict acceptance decisions, though leniency bias and adversarial vulnerabilities persist.}
\vspace{2pt}
\resizebox{\linewidth}{!}{\begin{tabular}{c| >{\centering\arraybackslash}p{3cm} r r |c| >{\centering\arraybackslash}p{2.4cm} |c| p{9cm}}
\toprule
\rowcolor{tableheader!10}
\textbf{\#} & \textbf{Method} & \textbf{Ref} & \textbf{Venue} & \textbf{Link} & \textbf{Category} & \textbf{GitHub} & \textbf{Evaluation} \\
\midrule
\multicolumn{8}{@{}l}{\cellcolor{S6color!65}\textbf{\textsf{\textcolor{white}{~~Automated Review Generation}}}} \\
\addlinespace[1pt]
{1}  & ChatReviewer & \cite{chatreviewer2023} & {\small GitHub'23} & \href{https://github.com/nishiwen1214/ChatReviewer}{\faExternalLink} & Review Gen.       & \githubicon{https://github.com/nishiwen1214/ChatReviewer}                & ChatGPT-based strengths/weaknesses analysis \\
{2}  & AI-Peer-Review & \cite{poldrack2024aireview} & {\small GitHub'24} & \href{https://github.com/poldrack/ai-peer-review}{\faExternalLink} & Review Gen.       & \githubicon{https://github.com/poldrack/ai-peer-review}                  & Multi-LLM independent reviews + meta-review synthesis \\
{3}  & MARG & \cite{darcy2024marg} & {\small arXiv'24} & \href{https://arxiv.org/abs/2401.04259}{\faExternalLink} & Review Gen.       & -                                                                     & 3.7 good comments/paper (2.2$\times$ over baseline) \\
{4}  & Reviewer2 & \cite{reviewer2024} & {\small arXiv'24} & \href{https://arxiv.org/abs/2402.10886}{\faExternalLink} & Review Gen.       & -                                                                     & Two-stage prompt-based review aspect modeling \\
{5}  & ReviewRL & \cite{reviewrl2025} & {\small EMNLP'25} & \href{https://aclanthology.org/2025.emnlp-main.857/}{\faExternalLink} & Review Gen.       & -                                                                     & RL + RAG; factually grounded reviews \\
{6}  & DeepReviewer & \cite{deepreviewer2025} & {\small arXiv'25} & \href{https://arxiv.org/abs/2503.08569}{\faExternalLink} & Review Gen.       & -                                                                     & 88.21\% win rate vs GPT-o1; 64\% accept/reject acc. \\
{7}  & OpenReviewer & \cite{openreviewer2025} & {\small NAACL'25} & \href{https://aclanthology.org/2025.naacl-demo.44/}{\faExternalLink} & Review Gen.       & -                                                                     & Llama-8B fine-tuned on 79K expert reviews \\
{8}  & REMOR & \cite{remor2025} & {\small arXiv'25} & \href{https://arxiv.org/abs/2505.11718}{\faExternalLink} & Review Gen.       & -                                                                     & Multi-objective RL review optimization \\
{9}  & ScholarPeer & \cite{scholarpeer2026} & {\small arXiv'26} & \href{https://arxiv.org/abs/2601.22638}{\faExternalLink} & Review Gen.       & -                                                                     & Context-aware multi-agent; literature verification \\
{1} & PeerCheck & \cite{chen2026peercheck} & {\small arXiv'26} & \href{https://arxiv.org/abs/2606.20897}{\faExternalLink} & Review Gen. & - & Pushes LLM-generated reviews toward human-level quality \\
\addlinespace[3pt]
\multicolumn{8}{@{}l}{\cellcolor{S6color!65}\textbf{\textsf{\textcolor{white}{~~Meta-Review \& Reviewer Matching}}}} \\
\addlinespace[1pt]
{2} & AgentReview & \cite{agentreview2024} & {\small EMNLP'24} & \href{https://aclanthology.org/2024.emnlp-main.70/}{\faExternalLink} & Meta-Review      & -                                                                     & Full review lifecycle simulation; social/authority bias \\
{3} & Meta-Review LLMs & \cite{metareviewllm2024} & {\small NAACL'25} & \href{https://aclanthology.org/2025.naacl-long.395/}{\faExternalLink} & Meta-Review      & -                                                                     & 40 ICLR papers; GPT-3.5/LLaMA-2/PaLM-2 compared \\
{4} & RATE & \cite{rate2026} & {\small arXiv'26} & \href{https://arxiv.org/abs/2601.19637}{\faExternalLink} & Matching         & -                                                                     & Expertise-based matching via profile distillation \\
\addlinespace[3pt]
\multicolumn{8}{@{}l}{\cellcolor{S6color!65}\textbf{\textsf{\textcolor{white}{~~Adversarial Attacks \& Bias Analysis}}}} \\
\addlinespace[1pt]
{5} & Raina~\etal & \cite{raina2024adversarialjudge} & {\small EMNLP'24} & \href{https://arxiv.org/abs/2402.14016}{\faExternalLink} & Adversarial      & -                                                                     & Benign adjectives as universal adversarial triggers \\
{6} & AI Review Lottery & \cite{ailottery2024} & {\small arXiv'24} & \href{https://arxiv.org/abs/2405.02150}{\faExternalLink} & Bias Analysis    & -                                                                     & 15.8\% ICLR reviews AI-assisted; +4.9pp borderline \\
{7} & Ye~\etal & \cite{ye2024peerrisks} & {\small arXiv'24} & \href{https://arxiv.org/abs/2412.01708}{\faExternalLink} & Adversarial      & -                                                                     & Scores inflated to $\sim$8; 5\% manipulation flips 12\% \\
{8} & Breaking Reviewer & \cite{breakingreviewer2025} & {\small arXiv'25} & \href{https://arxiv.org/abs/2506.11113}{\faExternalLink} & Adversarial     & -                                                                     & Systematic adversarial robustness evaluation \\
{9} & LLM Reviewer Bias & \cite{llmreviewer2025} & {\small arXiv'25} & \href{https://arxiv.org/abs/2509.09912}{\faExternalLink} & Bias Analysis    & -                                                                     & 1,441 papers; 95.8\% rejected misclassified as acceptable \\
{10} & Prompt Injection & \cite{promptinjectionreview2025} & {\small arXiv'25} & \href{https://arxiv.org/abs/2509.10248}{\faExternalLink} & Adversarial     & -                                                                     & White-text injection; up to 100\% acceptance scores \\
{11} & Sahoo~\etal & \cite{sahoo2025indirect} & {\small arXiv'25} & \href{https://arxiv.org/abs/2512.10449}{\faExternalLink} & Adversarial      & -                                                                     & +13.95 on Mistral; 13 LLMs; 15 attack strategies \\
{12} & Zhou~\etal & \cite{zhou2025positiveprompt} & {\small arXiv'25} & \href{https://arxiv.org/abs/2511.01287}{\faExternalLink} & Adversarial      & -                                                                     & +1.24 to +2.80 from hype prose; 10.00 under iteration \\
{13} & Gaming AI Reviews & \cite{li2026gaming} & {\small arXiv'26} & \href{https://arxiv.org/abs/2606.10159}{\faExternalLink} & Adversarial & - & Gaming AI-assisted peer reviews poses new risks \\
{14} & Phantom Refs & \cite{russinovich2026phantom} & {\small arXiv'26} & \href{https://arxiv.org/abs/2607.00738}{\faExternalLink} & Adversarial & - & Hallucinated citations survive top-tier peer review \\
\addlinespace[3pt]
\multicolumn{8}{@{}l}{\cellcolor{S6color!65}\textbf{\textsf{\textcolor{white}{~~Detection \& Policy}}}} \\
\addlinespace[1pt]
{15} & AI Detection & \cite{aidetectionreview2025} & {\small arXiv'25} & \href{https://arxiv.org/abs/2502.19614}{\faExternalLink} & Detection        & -                                                                     & 788,984 AI-written reviews; 18 detection algorithms \\
{16} & AI Use Rejects & \cite{aiuserejects2026} & {\small Nature'26} & \href{https://www.nature.com/articles/d41586-026-00893-2}{\faExternalLink} & Policy           & -                                                                     & 497 papers rejected ($\sim$2\% of submissions) \\
{17} & Nature AI Survey & \cite{aireviewsurvey2025} & {\small Nature'26} & \href{https://www.nature.com/articles/d41586-025-04066-5}{\faExternalLink} & Survey           & -                                                                     & 1,600 academics; 57\% use AI in peer review \\
{18} & Policy Enforcement & \cite{reviewpolicyenforce2026} & {\small arXiv'26} & \href{https://arxiv.org/abs/2603.20450}{\faExternalLink} & Policy          & -                                                                     & All 5 SOTA detectors misclassify LLM-polished reviews \\
{19} & Reviewer Feedback & \cite{reviewerfeedback2026} & {\small CHI'26} & \href{https://doi.org/10.1145/3772318.3791431}{\faExternalLink} & Empirical        & -                                                                     & ICLR 2025 live process; reviewer engagement study$^\dagger$ \\
{20} & AAAI-26 Pilot & \cite{biswas2026aaai} & {\small arXiv'26} & \href{https://arxiv.org/abs/2604.13940}{\faExternalLink} & Empirical & - & AI-assisted peer review at scale; the AAAI-26 pilot \\
\addlinespace[3pt]
\multicolumn{8}{@{}l}{\cellcolor{S6color!65}\textbf{\textsf{\textcolor{white}{~~Review Consistency and Bias Assessment}}}} \\
\addlinespace[1pt]
{21} & Review Survey & \cite{zhuang2025asprsurvey} & {\small IF'25} & \href{https://www.sciencedirect.com/science/article/pii/S1566253525004051}{\faExternalLink} & Survey     & -                                                                     & Comprehensive taxonomy of review methods \\
{22} & Stanford Agentic & \cite{stanfordreviewer2025} & {\small Web'25} & \href{https://paperreview.ai/tech-overview}{\faExternalLink} & Quality          & -                                                                     & $\rho=0.42$ vs human $\rho=0.41$; matches consistency \\
{23} & ClaimCheck & \cite{claimcheck2025} & {\small EMNLP'25} & \href{https://aclanthology.org/2025.findings-emnlp.1185/}{\faExternalLink} & Quality          & -                                                                     & LLM critique grounding; gaps in factual basis \\
{24} & ReViewGraph & \cite{reviewgraph2025} & {\small AAAI'26} & \href{https://arxiv.org/abs/2511.08317}{\faExternalLink} & Quality          & -                                                                     & +15.73\% avg improvement via heterogeneous graph \\
{25} & ReviewAgents & \cite{reviewagents2025} & {\small arXiv'25} & \href{https://arxiv.org/abs/2503.08506}{\faExternalLink} & Quality          & -                                                                     & 37,403 papers; 142,324 reviews; Review-CoT dataset \\
{26} & ICLR 2025 Study & \cite{iclr2025reviewstudy} & {\small NMI'26} & \href{https://www.nature.com/articles/s42256-026-01188-x}{\faExternalLink} & Quality       & -                                                                     & 22,467 reviews; 89\% quality improved; no acceptance effect \\
{27} & AI Reviewer Limits & \cite{aireviewerslimits2026} & {\small arXiv'26} & \href{https://arxiv.org/abs/2605.20668}{\faExternalLink} & Quality & - & 45 experts; AI vs.\ human reviews of Nature-family papers \\
{28} & PRISM & \cite{prismreview2026} & {\small arXiv'26} & \href{https://arxiv.org/abs/2605.26730}{\faExternalLink} & Quality & - & Multi-dimensional benchmark for LLM peer reviewers \\
{29} & ProReviewer & \cite{proreviewer2026} & {\small arXiv'26} & \href{https://arxiv.org/abs/2606.13349}{\faExternalLink} & Review Gen. & - & Proactive investigation agent; probes suspicious paper parts \\
{30} & MERIT & \cite{meritmatching2026} & {\small arXiv'26} & \href{https://arxiv.org/abs/2605.27865}{\faExternalLink} & Matching & - & Rubric-informed reviewer assignment via expertise modeling \\
{31} & Present. Gaming & \cite{presentationgaming2026} & {\small arXiv'26} & \href{https://arxiv.org/abs/2606.13044}{\faExternalLink} & Adversarial & - & Presentation-only revisions game AI review without prompts \\
{32} & LLMs Favor LLMs? & \cite{llmfavorllm2026} & {\small arXiv'26} & \href{https://arxiv.org/abs/2601.20920}{\faExternalLink} & Bias Analysis & - & Apparent self-preference is largely general leniency 
\\
\bottomrule
\end{tabular}}
\vspace{-0.7cm}
\end{table*}
\endgroup

\clearpage
\begingroup
\renewcommand{\arraystretch}{1.15}
\setlength{\tabcolsep}{3pt}
\footnotesize
\rowcolors{2}{white}{S7color!6}
\begin{table*}[!ht]
\centering
\caption{\textbf{Comprehensive inventory: \Sseven Rebuttal \& Revision.} $^\dagger$Evaluation information might be uncertain.}
\vspace{-7pt}
\label{tab:appendix_s7}
\stagecard{\Sseven}{Rebuttal \& Revision}{S7color}{figures/icons/s7_rebuttal.png}{%
Analyzing reviewer comments and generating evidence-grounded responses. The newest frontier in AI auto-research, rebuttal automation, is decisive for roughly one in five borderline submissions at major venues.}
\vspace{2pt}
\resizebox{\linewidth}{!}{\begin{tabular}{c| >{\centering\arraybackslash}p{3cm} r r |c| >{\centering\arraybackslash}p{2.4cm} |c| p{8.5cm}}
\toprule
\rowcolor{tableheader!10}
\textbf{\#} & \textbf{Method} & \textbf{Ref} & \textbf{Venue} & \textbf{Link} & \textbf{Category} & \textbf{GitHub} & \textbf{Evaluation} \\
\midrule
\multicolumn{8}{@{}l}{\cellcolor{S7color!65}\textbf{\textsf{\textcolor{white}{~~Reviewer Comment Analysis}}}} \\
\addlinespace[1pt]
{1}  & ReviewMT & \cite{reviewmt2024} & {\small arXiv'24} & \href{https://arxiv.org/abs/2406.05688}{\faExternalLink} & Analysis         & -                                                                     & 26,841 papers; 92,017 reviews; multi-turn dialogue \\
{2}  & ICLR Rebuttal Study & \cite{iclr_rebuttal2025} & {\small arXiv'25} & \href{https://arxiv.org/abs/2511.15462}{\faExternalLink} & Analysis         & -                                                                     & ICLR 2024--2025; score transition analysis \\
\addlinespace[3pt]
\multicolumn{8}{@{}l}{\cellcolor{S7color!65}\textbf{\textsf{\textcolor{white}{~~Automated Rebuttal Generation}}}} \\
\addlinespace[1pt]
{3}  & ReviewerToo & \cite{reviewertoo2025} & {\small arXiv'25} & \href{https://arxiv.org/abs/2510.08867}{\faExternalLink} & Modular Pipeline & -                                                                     & 81.8\% accept/reject accuracy (vs 83.9\% human) \\
{4}  & RebuttalAgent & \cite{rebuttalagent2026} & {\small ICLR'26} & \href{https://arxiv.org/abs/2601.15715}{\faExternalLink} & Rebuttal Gen.     & \githubicon{https://github.com/Zhitao-He/RebuttalAgent}             & 18.3\% avg improvement; ToM-grounded \\
{5}  & Author-in-the-Loop & \cite{ruan2026authorinloop} & {\small ACL'26} & \href{https://arxiv.org/abs/2602.11173}{\faExternalLink} & Author-Aware     & -                                                                     & Integrates author expertise and intent \\
{6}  & DRPG & \cite{drpg2026} & {\small arXiv'26} & \href{https://arxiv.org/abs/2601.18081}{\faExternalLink} & Rebuttal Gen.     & \githubicon{https://github.com/ulab-uiuc/DRPG-RebuttalAgent}             & 98\%+ planning accuracy; surpasses avg human quality \\
{7}  & Paper2Rebuttal & \cite{paper2rebuttal2026} & {\small arXiv'26} & \href{https://arxiv.org/abs/2601.14171}{\faExternalLink} & Rebuttal Gen.     & -                                                                     & Evidence-centric rebuttal planning \\
\addlinespace[3pt]
\multicolumn{8}{@{}l}{\cellcolor{S7color!65}\textbf{\textsf{\textcolor{white}{~~Rebuttal Effectiveness Assessment}}}} \\
\addlinespace[1pt]
{8}  & Re$^2$ & \cite{re2dataset2025} & {\small arXiv'25} & \href{https://arxiv.org/abs/2505.07920}{\faExternalLink} & Dataset          & -                                                                     & 19,926 submissions; 70,668 reviews; 53,818 rebuttals \\
{9}  & Commitment Checklist & \cite{rebuttalcommitment2026} & {\small arXiv'26} & \href{https://arxiv.org/abs/2603.00003}{\faExternalLink} & Benchmark     & -                                                                     & 11.8 commitments/paper; $\sim$25\% unfulfilled \\
{1} & Re$^3$Align & \cite{ruan2026authorinloop} & {\small ACL'26} & \href{https://arxiv.org/abs/2602.11173}{\faExternalLink} & Dataset     & -                                                                     & First large-scale aligned review--response--revision triplets \\
{2} & RbtAct & \cite{rbtact2026} & {\small arXiv'26} & \href{https://arxiv.org/abs/2603.09723}{\faExternalLink} & Analysis & - & Rebuttals as supervision for actionable review feedback \\
{3} & GoodPoint & \cite{goodpoint2026} & {\small arXiv'26} & \href{https://arxiv.org/abs/2604.11924}{\faExternalLink} & Analysis & - & Learns constructive feedback grounded in author responses \\
{4} & Defend & \cite{defendrebuttal2026} & {\small arXiv'26} & \href{https://arxiv.org/abs/2603.27360}{\faExternalLink} & Rebuttal Gen. & - & Minimal-guidance rebuttal generation; improved factual correctness \\
{5} & Rebuttals Move & \cite{louis2026rebuttals} & {\small arXiv'26} & \href{https://arxiv.org/abs/2606.22166}{\faExternalLink} & Analysis & - & Rebuttals move scores; initial-review structure bounds movement \\
{6} & Trust AI Reviews & \cite{leblanc2026trust} & {\small arXiv'26} & \href{https://arxiv.org/abs/2605.16623}{\faExternalLink} & Analysis & - & Studies authors' response to AI-based reviews \\
\bottomrule
\end{tabular}}
\end{table*}
\endgroup

\clearpage
\subsection{Phase 4: Dissemination}

\begingroup
\renewcommand{\arraystretch}{1.15}
\setlength{\tabcolsep}{3pt}
\footnotesize
\rowcolors{2}{white}{S8color!6}
\begin{table*}[!ht]
\centering
\caption{\textbf{Comprehensive inventory: \Seight Dissemination (Paper2X).} $^\dagger$Evaluation information might be uncertain.}
\vspace{-7pt}
\label{tab:appendix_s8}
\stagecard{\Seight}{Dissemination / Paper2X}{S8color}{figures/icons/s8_dissemination.png}{%
Converting papers into posters, slides, videos, websites, and social media content. Each output format targets a different audience and demands its own design logic, AI tool chain, and trust considerations.}
\vspace{2pt}
\resizebox{\linewidth}{!}{\begin{tabular}{c| >{\centering\arraybackslash}p{3cm} r r |c| >{\centering\arraybackslash}p{2.2cm} |c| p{8cm}}
\toprule
\rowcolor{tableheader!10}
\textbf{\#} & \textbf{Method} & \textbf{Ref} & \textbf{Venue} & \textbf{Link} & \textbf{Category} & \textbf{GitHub} & \textbf{Evaluation} \\
\midrule
\multicolumn{8}{@{}l}{\cellcolor{S8color!65}\textbf{\textsf{\textcolor{white}{~~Paper2Poster}}}} \\
\addlinespace[1pt]
{1}  & P2P & \cite{p2p2025} & {\small ICLR'26} & \href{https://openreview.net/forum?id=JojyT9niJL}{\faExternalLink} & Paper2Poster     & -                                                                     & P2PInstruct 30K+ examples; 3 specialized agents \\
{2}  & Paper2Poster & \cite{paper2poster2025} & {\small NeurIPS'25} & \href{https://openreview.net/forum?id=p0E74lpRBD}{\faExternalLink} & Paper2Poster     & \githubicon{https://github.com/Paper2Poster/Paper2Poster}                & \$0.005/poster; 87\% fewer tokens vs GPT-4o \\
{3}  & PosterForest & \cite{choi2025posterforest} & {\small arXiv'25} & \href{https://arxiv.org/abs/2508.21720}{\faExternalLink} & Paper2Poster     & -                                                                     & Hierarchical multi-agent collaboration$^\dagger$ \\
{4}  & PosterGen & \cite{postergen2025} & {\small arXiv'25} & \href{https://arxiv.org/abs/2508.17188}{\faExternalLink} & Paper2Poster     & -                                                                     & Aesthetic-aware multi-agent generation$^\dagger$ \\
{5}  & APEX & \cite{apex2026} & {\small arXiv'26} & \href{https://arxiv.org/abs/2601.04794}{\faExternalLink} & Paper2Poster     & \githubicon{https://github.com/Breesiu/APEX}                             & First agentic interactive poster editing \\
{6}  & PosterOmni & \cite{posteromni2026} & {\small arXiv'26} & \href{https://arxiv.org/abs/2602.12127}{\faExternalLink} & Paper2Poster     & -                                                                     & 6 unified poster tasks; outperforms open-source \\
\addlinespace[3pt]
\multicolumn{8}{@{}l}{\cellcolor{S8color!65}\textbf{\textsf{\textcolor{white}{~~Paper2Slides}}}} \\
\addlinespace[1pt]
{7}  & DOC2PPT & \cite{doc2ppt2022} & {\small AAAI'22} & \href{https://ojs.aaai.org/index.php/AAAI/article/view/19943}{\faExternalLink} & Paper2Slides     & -                                                                     & 5,873 paired document--slide decks \\
{8}  & PPTAgent & \cite{pptagent2025} & {\small EMNLP'25} & \href{https://arxiv.org/abs/2501.03936}{\faExternalLink} & Paper2Slides     & \githubicon{https://github.com/icip-cas/PPTAgent}                        & PPTEval benchmark; 10,448 curated presentations \\
{9}  & AutoPresent & \cite{autopresent2025} & {\small CVPR'25} & \href{https://arxiv.org/abs/2501.00912}{\faExternalLink} & Paper2Slides     & -                                                                     & 8B Llama model; SlidesBench (7K train, 585 test) \\
{1} & Paper2Slides & \cite{paper2slides2025} & {\small GitHub'25} & \href{https://github.com/HKUDS/Paper2Slides}{\faExternalLink} & Paper2Slides     & \githubicon{https://github.com/HKUDS/Paper2Slides}                       & 4-stage RAG pipeline; one-click conversion \\
{2} & Auto-Slides & \cite{autoslides2025} & {\small arXiv'25} & \href{https://arxiv.org/abs/2509.11062}{\faExternalLink} & Paper2Slides     & -                                                                     & Multi-agent Beamer generation; interactive editing$^\dagger$ \\
{3} & PASS & \cite{pass2025} & {\small arXiv'25} & \href{https://arxiv.org/abs/2501.06497}{\faExternalLink} & Paper2Slides     & -                                                                     & First combined slides + AI audio delivery$^\dagger$ \\
{4} & SlideGen & \cite{slidegen2025} & {\small arXiv'25} & \href{https://arxiv.org/abs/2512.04529}{\faExternalLink} & Paper2Slides     & -                                                                     & Multi-agent VLM coordination; editable PPTX output \\
{5} & Talk to Your Slides & \cite{talkslides2025} & {\small arXiv'25} & \href{https://arxiv.org/abs/2505.11604}{\faExternalLink} & Paper2Slides     & -                                                                     & +34\% instruction fidelity; 87\% lower cost vs GUI \\
{6} & SlideTailor & \cite{slidetailor2025} & {\small AAAI'26} & \href{https://arxiv.org/abs/2512.20292}{\faExternalLink} & Paper2Slides     & \githubicon{https://github.com/nusnlp/SlideTailor}                       & User-preference conditioned; chain-of-speech \\
{7} & DeepPresenter & \cite{deeppresenter2026} & {\small arXiv'26} & \href{https://arxiv.org/abs/2602.22839}{\faExternalLink} & Paper2Slides     & \githubicon{https://github.com/icip-cas/PPTAgent}                        & 9B model competitive with frontier at lower cost \\
{8} & Office Raccoon & \cite{sensetime2026} & {\small Web'26} & \href{https://www.sensetime.com/en/news-detail/51170569}{\faExternalLink} & Paper2Slides     & -                                                                     & Page-level editing; template/brand-guideline learning$^\dagger$ \\
\addlinespace[3pt]
\multicolumn{8}{@{}l}{\cellcolor{S8color!65}\textbf{\textsf{\textcolor{white}{~~Paper2Video}}}} \\
\addlinespace[1pt]
{9} & Preacher & \cite{preacher2025} & {\small ICCV'25} & \href{https://github.com/Gen-Verse/Paper2Video}{\faExternalLink} & Paper2Video      & \githubicon{https://github.com/Gen-Verse/Paper2Video}                    & Top-down decomposition; 5 research fields \\
{10} & Paper2Video & \cite{paper2video2025} & {\small arXiv'25} & \href{https://arxiv.org/abs/2510.05096}{\faExternalLink} & Paper2Video      & \githubicon{https://github.com/showlab/Paper2Video}                      & 101 paper--video pairs; +10\% PresentQuiz accuracy \\
{11} & PresentAgent & \cite{presentagent2025} & {\small EMNLP'25} & \href{https://aclanthology.org/2025.emnlp-demos.58/}{\faExternalLink} & Paper2Video      & \githubicon{https://github.com/AIGeeksGroup/PresentAgent}                & PresentEval benchmark; approaches human-level \\
{12} & Paper2Video Talks & \cite{mondal2026talks} & {\small arXiv'26} & \href{https://arxiv.org/abs/2606.28531}{\faExternalLink} & Paper2Video & - & A good talk teaches takeaways, not summaries \\
\addlinespace[3pt]
\multicolumn{8}{@{}l}{\cellcolor{S8color!65}\textbf{\textsf{\textcolor{white}{~~Paper2Web \& Social Media}}}} \\
\addlinespace[1pt]
{13} & Paper2Web & \cite{paper2web2025} & {\small arXiv'25} & \href{https://arxiv.org/abs/2510.15842}{\faExternalLink} & Paper2Web        & \githubicon{https://github.com/YuhangChen1/Paper2All}                    & 10,716 papers; multimedia-rich academic homepages \\
{14} & ResearchStudio-Reel & \cite{xiao2026researchstudio} & {\small arXiv'26} & \href{https://arxiv.org/abs/2607.04438}{\faExternalLink} & Paper2Web & - & Automates paper to poster, video, and blog jointly \\
{15} & I-WebGenBench & \cite{dai2026iwebgen} & {\small arXiv'26} & \href{https://arxiv.org/abs/2606.00750}{\faExternalLink} & Paper2Web & - & Evaluates interactivity in LLM-generated scientific web apps \\
\addlinespace[3pt]
\multicolumn{8}{@{}l}{\cellcolor{S8color!65}\textbf{\textsf{\textcolor{white}{~~Fidelity and Adoption Assessment}}}} \\
\addlinespace[1pt]
{16} & PPTEval & \cite{pptagent2025} & {\small EMNLP'25} & \href{https://arxiv.org/abs/2501.03936}{\faExternalLink} & Benchmark        & \githubicon{https://github.com/icip-cas/PPTAgent}                        & Content, design, coherence; 10,448 presentations \\
{17} & PresentQuiz & \cite{paper2video2025} & {\small arXiv'25} & \href{https://arxiv.org/abs/2510.05096}{\faExternalLink} & Benchmark        & \githubicon{https://github.com/showlab/Paper2Video}                      & 101 paper--video pairs; +10\% over human on comprehension \\
{18} & PresentEval & \cite{presentagent2025} & {\small EMNLP'25} & \href{https://aclanthology.org/2025.emnlp-demos.58/}{\faExternalLink} & Benchmark        & \githubicon{https://github.com/AIGeeksGroup/PresentAgent}                & End-to-end narrated video quality; near human-level \\
{19} & Any2Poster & \cite{any2poster2026} & {\small arXiv'26} & \href{https://arxiv.org/abs/2606.02915}{\faExternalLink} & Paper2Poster & - & Poster generation from eight input modalities across domains \\
{20} & X+Slides & \cite{xslides2026} & {\small arXiv'26} & \href{https://arxiv.org/abs/2606.19256}{\faExternalLink} & Paper2Slides & - & Audience-conditioned slide generation benchmark \\
{21} & PresentAgent-2 & \cite{presentagentv2_2026} & {\small arXiv'26} & \href{https://arxiv.org/abs/2605.11363}{\faExternalLink} & Paper2Video & - & Generalist agentic presentation-video generation \\
{22} & Sci.\ Comm.\ Correspondence & \cite{scicommcorr2026} & {\small arXiv'26} & \href{https://arxiv.org/abs/2605.05831}{\faExternalLink} & Benchmark & - & Fine-grained paper--slides--video correspondence \\
\bottomrule
\end{tabular}}
\end{table*}
\endgroup

\clearpage
\subsection{Cross-Phase: End-to-End Systems}

\begingroup
\renewcommand{\arraystretch}{1.15}
\setlength{\tabcolsep}{3pt}
\footnotesize
\rowcolors{2}{white}{tableheader!6}
\begin{table*}[!ht]
\centering
\caption{\textbf{Comprehensive inventory: End-to-End and Cross-Phase Systems.} $^\dagger$Evaluation information might be uncertain.}
\vspace{-7pt}
\label{tab:appendix_e2e}
\vspace{2pt}
\resizebox{\linewidth}{!}{\begin{tabular}{c| >{\centering\arraybackslash}p{3cm} r r |c| >{\centering\arraybackslash}p{2.5cm} |c| p{8.9cm}}
\toprule
\rowcolor{tableheader!10}
\textbf{\#} & \textbf{Method} & \textbf{Ref} & \textbf{Venue} & \textbf{Link} & \textbf{Category} & \textbf{GitHub} & \textbf{Evaluation} \\
\midrule
\multicolumn{8}{@{}l}{\cellcolor{tableheader!65}\textbf{\textsf{\textcolor{white}{~~Fully Automated Research Systems}}}} \\
\addlinespace[1pt]
{1} & AI Scientist & \cite{lu2024aiscientist} & {\small arXiv'24} & \href{https://arxiv.org/abs/2408.06292}{\faExternalLink} & E2E Pipeline & \githubicon{https://github.com/SakanaAI/AI-Scientist} & Pioneered E2E at \$15/paper; ICLR-scale review \\
{2} & Agent Laboratory & \cite{schmidgall2025agentlab} & {\small EMNLP'25} & \href{https://aclanthology.org/2025.findings-emnlp.320/}{\faExternalLink} & E2E Pipeline & - & \$2--13/paper; 3.5--4.0 NeurIPS scale \\
{3} & AI-Researcher & \cite{airesearcher2025} & {\small arXiv'25} & \href{https://arxiv.org/abs/2505.18705}{\faExternalLink} & E2E Pipeline & - & Scientist-Bench; approaches human-level quality \\
{4} & CycleResearcher & \cite{cycleresearcher2024} & {\small ICLR'25} & \href{https://arxiv.org/abs/2411.00816}{\faExternalLink} & E2E Pipeline & - & 5.36 ICLR scale; cyclic write--review \\
{5} & AI Scientist v2 & \cite{yamada2025aiscientistv2} & {\small arXiv'25} & \href{https://arxiv.org/abs/2504.08066}{\faExternalLink} & E2E + Tree Search & \githubicon{https://github.com/SakanaAI/AI-Scientist-v2} & ICLR 2025 ICBINB workshop; score 6.33 \\
{6} & Kosmos & \cite{kosmos2025} & {\small arXiv'25} & \href{https://arxiv.org/abs/2511.02824}{\faExternalLink} & E2E Pipeline & - & 79.4\% claim accuracy; 7 discoveries; 4 domains \\
{7} & Dolphin & \cite{dolphin2025} & {\small ACL'25} & \href{https://aclanthology.org/2025.acl-long.1056/}{\faExternalLink} & E2E Pipeline & - & Closed-loop auto-research pipeline \\
{8} & CodeScientist & \cite{codescientist2025} & {\small ACL'25} & \href{https://aclanthology.org/2025.findings-acl.692/}{\faExternalLink} & E2E Pipeline & \githubicon{https://github.com/allenai/codescientist} & Hypothesis to verification; closed-loop \\
{9} & InternAgent & \cite{novelseek2025} & {\small arXiv'25} & \href{https://arxiv.org/abs/2505.16938}{\faExternalLink} & E2E Pipeline & \githubicon{https://github.com/InternScience/InternAgent} & Closed-loop hypothesis to verification \\
{10} & freephdlabor & \cite{freephdlabor2025} & {\small arXiv'25} & \href{https://arxiv.org/abs/2510.15624}{\faExternalLink} & Multi-Agent & - & Personalized research group; continual automation \\
{11} & SciMaster & \cite{scimaster2025} & {\small arXiv'25} & \href{https://arxiv.org/abs/2507.05241}{\faExternalLink} & Multi-Agent & \githubicon{https://github.com/sjtu-sai-agents/X-Master} & General-purpose scientific AI agents \\
{12} & ARIS & \cite{aris2025} & {\small GitHub'26} & \href{https://github.com/wanshuiyin/Auto-claude-code-research-in-sleep}{\faExternalLink} & Skill Library & \githubicon{https://github.com/wanshuiyin/Auto-claude-code-research-in-sleep} & 31 skills; score 5.0$\to$7.5; 20+ GPU experiments \\
{13} & EvoScientist & \cite{evoscientist2026} & {\small arXiv'26} & \href{https://arxiv.org/abs/2603.08127}{\faExternalLink} & Multi-Agent & \githubicon{https://github.com/EvoScientist/EvoScientist} & 6 papers accepted at ICAIS'25 \\
{14} & UniScientist & \cite{uniscientist2026} & {\small Web'26} & \href{https://unipat.ai/blog/UniScientist}{\faExternalLink} & Multi-Agent & - & 30B open-source; beats Claude Opus 4.5 \\
{15} & ASI-Evolve & \cite{asievolve2026} & {\small GitHub'26} & \href{https://github.com/GAIR-NLP/ASI-Evolve}{\faExternalLink} & Multi-Agent & \githubicon{https://github.com/GAIR-NLP/ASI-Evolve} & +0.97 DeltaNet; +18 MMLU; +12.5 GRPO \\
{16} & AiScientist-LH & \cite{aiscientistlonghorizon2026} & {\small arXiv'26} & \href{https://arxiv.org/abs/2604.13018}{\faExternalLink} & E2E + Hierarchical & \githubicon{https://github.com/AweAI-Team/AiScientist} & Long-horizon ML engineering; File-as-Bus \\
{17} & AutoSOTA & \cite{autosota2026} & {\small arXiv'26} & \href{https://arxiv.org/abs/2604.05550}{\faExternalLink} & E2E Pipeline & \githubicon{https://github.com/tsinghua-fib-lab/AutoSOTA} & Paper-to-code-to-SOTA optimization \\
{18} & CORAL & \cite{coral2026} & {\small arXiv'26} & \href{https://arxiv.org/abs/2604.01658}{\faExternalLink} & Multi-Agent & \githubicon{https://github.com/Human-Agent-Society/CORAL} & Multi-agent evolution; SOTA on 10 tasks \\
{19} & FARS & \cite{fars2026} & {\small Web'26} & \href{https://analemma.ai/blog/introducing-fars}{\faExternalLink} & Multi-Agent & - & 100 papers in 228h; avg 5.05 ICLR scale \\
{20} & AutoResearchClaw & \cite{autoresearchclaw2026} & {\small GitHub'26} & \href{https://github.com/aiming-lab/AutoResearchClaw}{\faExternalLink} & Multi-Agent & \githubicon{https://github.com/aiming-lab/AutoResearchClaw} & fully autonomous 23-stage pipeline \\
{21} & NVAITC AI Sci. & \cite{huang2026nvaitc} & {\small arXiv'26} & \href{https://arxiv.org/abs/2607.11084}{\faExternalLink} & E2E Pipeline & - & Governed E2E system; hypertension GWAS case study \\
\addlinespace[3pt]
\multicolumn{8}{@{}l}{\cellcolor{tableheader!65}\textbf{\textsf{\textcolor{white}{~~Domain-Specific Systems}}}} \\
\addlinespace[1pt]
{22} & Coscientist & \cite{boiko2023coscientist} & {\small Nature'23} & \href{https://www.nature.com/articles/s41586-023-06792-0}{\faExternalLink} & Chemistry & - & Autonomous chemistry; LLM-driven tool use \\
{23} & AlphaFold~3 & \cite{abramson2024alphafold3} & {\small Nature'24} & \href{https://www.nature.com/articles/s41586-024-07487-w}{\faExternalLink} & Biology & - & Biomolecular structure prediction \\
{24} & ChemCrow & \cite{bran2024chemcrow} & {\small NMI'24} & \href{https://www.nature.com/articles/s42256-024-00832-8}{\faExternalLink} & Chemistry & - & Chemistry tool orchestration \\
{25} & Med. AI Scientist & \cite{medicalaiscientist2026} & {\small arXiv'26} & \href{https://arxiv.org/abs/2603.28589}{\faExternalLink} & Medicine & - & Clinical research automation \\
{26} & Cognitive Scientist & \cite{jagadish2026cognitive} & {\small arXiv'26} & \href{https://arxiv.org/abs/2606.26448}{\faExternalLink} & Psychology & - & Closes the loop to discover psychological theories \\
{27} & Molecular Closed-Loop & \cite{ning2026closedloop} & {\small arXiv'26} & \href{https://arxiv.org/abs/2606.22731}{\faExternalLink} & Chemistry & - & Closed-loop research certifying generalizable improvements \\
\addlinespace[3pt]
\multicolumn{8}{@{}l}{\cellcolor{tableheader!65}\textbf{\textsf{\textcolor{white}{~~Evolutionary \& Self-Improving Systems}}}} \\
\addlinespace[1pt]
{28} & ShinkaEvolve & \cite{shinkaevolve2025} & {\small arXiv'25} & \href{https://arxiv.org/abs/2509.19349}{\faExternalLink} & Evolutionary & \githubicon{https://github.com/SakanaAI/ShinkaEvolve} & Open-ended sample-efficient program evolution \\
{29} & Darwin Godel Machine & \cite{darwingodelmachine2025} & {\small arXiv'25} & \href{https://arxiv.org/abs/2505.22954}{\faExternalLink} & Self-Improving & \githubicon{https://github.com/jennyzzt/dgm} & Open-ended evolution of self-improving agents \\
{30} & Self-Driving Lab & \cite{hur2026selfdrivinglab} & {\small arXiv'26} & \href{https://arxiv.org/abs/2607.04508}{\faExternalLink} & Self-Improving & - & Agentic self-driving lab compresses validation bottleneck \\
{31} & Meta-Reflection & \cite{zhao2026metareflection} & {\small arXiv'26} & \href{https://arxiv.org/abs/2607.01131}{\faExternalLink} & Self-Improving & - & Autonomous discovery via iterative meta-reflection \\
\addlinespace[3pt]
\multicolumn{8}{@{}l}{\cellcolor{tableheader!65}\textbf{\textsf{\textcolor{white}{~~Research Platforms \& Infrastructure}}}} \\
\addlinespace[1pt]
{32} & R\&D-Agent & \cite{chen2025rdagent} & {\small arXiv'25} & \href{https://arxiv.org/abs/2505.14738}{\faExternalLink} & Infrastructure & \githubicon{https://github.com/microsoft/RD-Agent} & Researcher+Developer dual-agent; MLE-Bench top \\
{33} & autoresearch & \cite{karpathy2025autoresearch} & {\small GitHub'25} & \href{https://github.com/karpathy/autoresearch}{\faExternalLink} & Infrastructure & \githubicon{https://github.com/karpathy/autoresearch} & ${\sim}$12 exp/hour overnight \\
{34} & Google Co-Scientist & \cite{gottweis2025aicoscientist} & {\small arXiv'25} & \href{https://arxiv.org/abs/2502.18864}{\faExternalLink} & Platform & - & Multi-agent hypothesis gen + validation \\
{35} & ResearchTown & \cite{researchtown2025} & {\small ICML'25} & \href{https://arxiv.org/abs/2412.17767}{\faExternalLink} & Multi-Agent & \githubicon{https://github.com/ulab-uiuc/research-town} & Simulates research community with LLM agents \\
{36} & AgentRxiv & \cite{schmidgall2025agentrxiv} & {\small arXiv'25} & \href{https://arxiv.org/abs/2503.18102}{\faExternalLink} & Multi-Agent & - & 11.4\% improvement on MATH-500 \\
{37} & LabClaw & \cite{labclaw2026} & {\small Web'26} & \href{https://labclaw-ai.github.io}{\faExternalLink} & Skill Library & \githubicon{https://github.com/wu-yc/LabClaw} & 206 biomedical skills; always-on autonomous lab agent \\
{38} & PiFlow & \cite{piflow2025} & {\small arXiv'25} & \href{https://arxiv.org/abs/2505.15047}{\faExternalLink} & Multi-Agent & - & Principle-aware scientific discovery \\
\addlinespace[3pt]
{39} & AutoSci & \cite{autosci2026} & {\small arXiv'26} & \href{https://arxiv.org/abs/2605.31468}{\faExternalLink} & E2E Pipeline & - & Memory-centric full-lifecycle agent; self-improving proc. \\
{40} & ScientistOne & \cite{scientistone2026} & {\small arXiv'26} & \href{https://arxiv.org/abs/2605.26340}{\faExternalLink} & E2E Pipeline & - & Chain-of-evidence verifiability against fabrication \\
{41} & Arbor & \cite{arbor2026} & {\small arXiv'26} & \href{https://arxiv.org/abs/2606.11926}{\faExternalLink} & Multi-Agent & - & Coordinator + executor agents; hypothesis-tree refinement \\
{42} & EvoMaster & \cite{evomaster2026} & {\small arXiv'26} & \href{https://arxiv.org/abs/2604.17406}{\faExternalLink} & Self-Improving & - & Self-evolving foundational agent for agentic science at scale \\
\addlinespace[3pt]
\multicolumn{8}{@{}l}{\cellcolor{tableheader!65}\textbf{\textsf{\textcolor{white}{~~Lifecycle Benchmarks}}}} \\
\addlinespace[1pt]
{43} & ResearchClawBench & \cite{researchclawbench2026} & {\small arXiv'26} & \href{https://arxiv.org/abs/2606.07591}{\faExternalLink} & Benchmark & - & E2E research tasks grounded in hidden papers + rubrics \\
{44} & Act as Real & \cite{realresearcher2026} & {\small arXiv'26} & \href{https://arxiv.org/abs/2606.07462}{\faExternalLink} & Benchmark & - & Full-lifecycle suite for frontier models + harnesses \\
{45} & ResearchArena & \cite{researcharena2026} & {\small arXiv'26} & \href{https://arxiv.org/abs/2605.19156}{\faExternalLink} & Benchmark & - & Minimal scaffold for running/assessing the complete loop \\
\bottomrule
\end{tabular}}
\vspace{-1cm}
\end{table*}
\endgroup

\clearpage\clearpage
\clearpage
\section{Survey Coverage Comparison \& Taxonomy Analysis}
\label{sec:appendix_surveys}

\Cref{tab:survey_comparison} compares our coverage with five closely related concurrent efforts. The goal is not to rank surveys, but to clarify how our lifecycle framework differs in scope and organization.

Our eight-stage framework subsumes several prior taxonomies while making two distinctions explicit: AI auto-research should be analyzed across the complete lifecycle, and stages should be grouped by epistemological function rather than only by task name or autonomy level.

\begin{wraptable}{r}{0.59\textwidth}
\vspace{-12pt}
\centering
\caption{%
  Comparison of survey coverage across our four-phase research lifecycle framework.
  \cmarkc~= in-depth coverage, \pmarkc~= partial coverage, \xmarkc~= not covered,
  {\textcolor{violet}{\scriptsize$\bigstar$\,new}}~= newly introduced stage.
}
\vspace{-6pt}
\label{tab:survey_comparison}
\renewcommand{\arraystretch}{1.15}
\setlength{\tabcolsep}{3pt}
\resizebox{0.56\textwidth}{!}{
\begin{tabular}{@{}l!{\vrule width 0.8pt}c!{\vrule width 0.8pt}ccccc@{}}
\toprule
\rowcolor{tableheader!10}
\textbf{Stage} &
\rotatebox{75}{\makecell[l]{\textbf{Ours}}} &
\rotatebox{75}{\makecell[l]{LLM4SR\\{\tiny 2501}}} &
\rotatebox{75}{\makecell[l]{PR Survey\\{\tiny 2501}}} &
\rotatebox{75}{\makecell[l]{Auto$\to$Auton\\{\tiny 2505}}} &
\rotatebox{75}{\makecell[l]{AI4Research\\{\tiny 2507}}} &
\rotatebox{75}{\makecell[l]{AI Scientists\\{\tiny 2510}}} \\
\multicolumn{7}{l}{\cellcolor{S1color!8}\textbf{Phase~1: Creation}} \\
\makecell[tl]{\Sone: Idea Generation\\\textcolor{gray}{\scriptsize novelty, feasibility, multi-agent}}
  & \cmarkc & \cmarkc & \xmarkc & \cmarkc & \cmarkc & \cmarkc \\
\makecell[tl]{\Stwo: Literature Review\\\textcolor{gray}{\scriptsize retrieval, survey gen, deep research}}
  & \cmarkc & \xmarkc & \xmarkc & \cmarkc & \cmarkc & \cmarkc \\
\makecell[tl]{\Sthree: Coding \& Experiments\\\textcolor{gray}{\scriptsize paper-to-code, execution, analysis}}
  & \cmarkc & \cmarkc & \xmarkc & \cmarkc & \cmarkc & \cmarkc \\
\makecell[tl]{\Sfour: Figures \& Tables\\\textcolor{gray}{\scriptsize diagrams, plots, formulas}}
  & \cmarkc & \xmarkc & \xmarkc & \pmarkc & \pmarkc & \xmarkc \\
\midrule
\multicolumn{7}{l}{\cellcolor{S5color!8}\textbf{Phase~2: Writing}} \\
\makecell[tl]{\Sfive: Paper Writing\\\textcolor{gray}{\scriptsize semi-auto, full-auto, detection}}
  & \cmarkc & \cmarkc & \xmarkc & \pmarkc & \cmarkc & \cmarkc \\
\midrule
\multicolumn{7}{l}{\cellcolor{S6color!8}\textbf{Phase~3: Validation}} \\
\makecell[tl]{\Ssix: Peer Review\\\textcolor{gray}{\scriptsize auto-review, matching, quality}}
  & \cmarkc & \cmarkc & \cmarkc & \pmarkc & \cmarkc & \xmarkc \\
\makecell[tl]{\Sseven: Rebuttal \& Revision\\\textcolor{gray}{\scriptsize comment analysis, rebuttal gen}}
  & {\textcolor{violet}{\scriptsize$\bigstar$\,new}} & \xmarkc & \xmarkc & \xmarkc & \xmarkc & \xmarkc \\
\midrule
\multicolumn{7}{l}{\cellcolor{S8color!8}\textbf{Phase~4: Dissemination}} \\
\makecell[tl]{\Seight: Dissemination\\\textcolor{gray}{\scriptsize poster, slides, video, social}}
  & {\textcolor{violet}{\scriptsize$\bigstar$\,new}} & \xmarkc & \xmarkc & \xmarkc & \xmarkc & \xmarkc \\
\midrule
\rowcolor{tableheader!6}
\textbf{Stages covered}
  & \textbf{8/8} & \textbf{4} & \textbf{1} & \textbf{5} & \textbf{5} & \textbf{4} \\
\bottomrule
\end{tabular}}
\vspace{-9pt}
\end{wraptable}
\begin{itemize}[leftmargin=*]
  \item \textbf{AI4Research}~\cite{ai4research2025} defines five task categories: Comprehension, Survey, Discovery, Writing, and Review. These overlap with our \Sone--\Sthree, \Sfive, and \Ssix. Our framework newly elevates \Sfour (Tables \& Figures), \Sseven (Rebuttal \& Revision), and \Seight (Dissemination) as independent lifecycle stages.

  \item \textbf{From Automation to Autonomy}~\cite{zheng2025automation} organizes systems by autonomy level, from tool-like assistance to scientist-level automation. This axis is complementary: each of our stages can be instantiated at different autonomy levels, while our framework specifies \emph{where} in the research lifecycle the system operates.

  \item \textbf{LLM4SR}~\cite{luo2025llm4sr} proposes a four-part view centered on hypothesis, experiment, writing, and review. This structure is close to ours, but does not separately model rebuttal and revision as a feedback stage. Our Validation phase separates \Ssix from \Sseven, making the review--response loop explicit.

  \item \textbf{Automated Scholarly Paper Review}~\cite{zhuang2025asprsurvey} provide in-depth coverage of review generation, quality assessment, and reviewer--paper matching. They are complementary to our work: they focus on \Ssix, while our framework places peer review within the broader lifecycle.

  \item \textbf{AI Scientist Survey}~\cite{tie2025survey} focus on autonomous or semi-autonomous scientific discovery, overlapping mainly with \Sone--\Sthree and \Sfive. Our framework extends this view by also covering scientific visualization, peer validation, rebuttal, and dissemination.
\end{itemize}

These comparisons show that prior taxonomies often list research tasks sequentially, while leaving functional distinctions and feedback loops implicit. Our four-phase framework makes these dependencies explicit. For example, \Ssix (\emph{Peer Review}) and \Sseven (\emph{Rebuttal \& Revision}) do not simply follow paper writing as isolated downstream steps; they can redirect the workflow back to \Sthree for additional experiments, \Sfour for revised figures or tables, and \Sfive for manuscript restructuring. Similarly, dissemination artifacts in \Seight may expose ambiguities in the original framing, requiring revisions to claims, explanations, or visual evidence. 

These cross-stage dependencies are central to real research practice and are especially important for AI-assisted workflows, where errors can propagate from generated ideas to experiments, from experiments to claims, and from claims to public-facing summaries. By organizing the field into \emph{Creation}, \emph{Writing}, \emph{Validation}, and \emph{Dissemination}, our framework highlights not only which stages are covered by existing systems, but also where evidence, claims, critique, and communication must remain aligned.

\clearpage\clearpage
\bibliographystyle{abbrvnat}
\bibliography{references}

\end{document}